\documentclass[fleqn]{llncs}

\usepackage{latexsym}
\usepackage{times}
\usepackage{theorem}

\global
\global

\def\OR      { \,\vee\,                   }

\def\IMPL    { \,\supset\,                }

\def\IFF     { \,\equiv\,         }

\newcommand{\seqinf}[3]{(#1)_{#2\geq #3}}

\newcommand{\fneg}{\mbox{$\sim$}}
\newcommand{\tempevo}{{\rm\textsc{TempEvo}}}
\renewcommand{\tempevo}{\textsc{TempEvo}}

\newcommand{\PSPACE}{\mbox{\rm{PSPACE}}}

\newcommand{\EXPSPACE}{\mbox{\rm{EXPSPACE}}}

\newcommand{\TWOEXPSPACE}{\mbox{\rm{$2$-EXPSPACE}}}
\newcommand{\TWOEXPTIME}{\mbox{\rm{$2$-EXPTIME}}}

\renewcommand{\phi}{\varphi}

\newcommand{\dsc}{\mathit{dsc}}
\newcommand{\tree}{\mathcal{T}}
\newcommand{\EKBL}{\mbox{\sf{EKBL}}}
\newcommand{\can}{\mathit{can}}
\newcommand{\EF}{\mathit{EF}}

\newcommand{\AC}{\mbox{$\mathcal{AC}$}}

\newcommand{\cl}[1]{\overline{#1}} 
\renewcommand{\cl}[1]{\neg #1}

\newcommand{\commadots}[0]{,\ldots ,}

\newcommand{\useq}[1]{{\mathbf #1}}
\renewcommand{\useq}[1]{\mbox{\it\bfseries #1}}

\newcommand{\url}[1]{\texttt{#1}}

\newcommand{\wrt}[0]{w.r.t.\ }
\renewcommand{\wrt}[0]{with respect to }

\newcommand{\Cn}[1]{\mathit{Cn}(#1)}
\newcommand{\Cni}[2]{\mathit{Cn}_{#1}(#2)}

\newcommand{\iec}[0]{i.e.,\ }
\newcommand{\egc}[0]{e.g.,\ }

\newcommand{\upd} {\lhd}

\newcommand{\head}[1]{H(#1)}

\newcommand{\body}[1]{B(#1)}
\newcommand{\bodyp}[1]{B^+(#1)}
\newcommand{\bodyn}[1]{B^-(#1)}
\newcommand{\KB}{{\it KB}}
\newcommand{\la}{\leftarrow}
\newcommand{\naf}{{\it not}\,}

\newcommand{\NP}{\mbox{${\rm N\!P}$}}
\renewcommand{\NP}{\mbox{\rm NP}}
\newcommand{\coNP}{\mbox{${\rm coN\!P}$}}

\renewcommand{\coNP}{\mbox{\rm coNP}}

\newcommand{\PiP}[1]{{\Pi}_{#1}^{P}}
\newcommand{\at}{{\mathit At}}
\renewcommand{\at}{{\mathcal A}}

\newcommand{\extat}{\at_{ext}}

\renewcommand{\extat}{{\mathit Lit}_{\at}}

\newcommand{\rs}{Rej}
\renewcommand{\rs}{\mathit{Rej}}

\newcommand{\bel}{Bel}
\newcommand{\lang}{{\cal L}}

\newcommand{\tuple}[1]{\langle#1\rangle}

\newcommand{\SMup}{\mathcal{D}}

\newcommand{\AS}{\mathit{AS}}
\renewcommand{\AS}{\mathcal{AS}}
\newcommand{\ASup}{\mathcal{U}}

\newcommand{\nop}[1]{}

\newcommand{\tr}{{\it tr}}
\newcommand{\rf}{\rho}
\newcommand{\KS}{\mathit{KS}}
\newcommand{\UP}{{\mathcal{U}}}
\newcommand{\up}{\Pi}
\newcommand{\compo}{\mathit{comp}}
\newcommand{\comp}[1]{\compo(#1)}

\newcommand{\EC}{{\mathcal{EC}}}

\newcommand{\UL}{{\sf EPI}}

\newcommand{\A}{\mathsf{A}}
\newcommand{\E}{\mathsf{E}}
\newcommand{\F}{\mathsf{F}}
\newcommand{\G}{\mathsf{G}}
\newcommand{\X}{\mathsf{X}}
\newcommand{\U}{\mathsf{U}}
\newcommand{\B}{\mathsf{B}}

\begin{document}

\title{Reasoning about Evolving \\
Nonmonotonic Knowledge Bases%
\thanks{A preliminary version of this paper 
appeared in: Proc.\ 8th International Conference on Logic 
for Programming, Artificial Intelligence and Reasoning (LPAR 2001),
R. Nieuwenhuis and A. Voronkov (eds), pp.\ 407--421, LNCS 2250,
Springer 2001.}}

\author{{Thomas Eiter},
{Michael Fink}, 
{Giuliana Sabbatini}, and
{Hans Tompits}
}
\institute{
Institut f{\"u}r Informationssysteme,
Technische Universit\"{a}t
Wien  \\ 
Favoritenstra{\ss}e 9-11, A-1040 Vienna, Austria\\
\email{\{eiter,michael,giuliana,tompits\}@kr.tuwien.ac.at}}

\maketitle

\begin{abstract}
Recently, several approaches to updating knowledge bases modeled as
extended logic programs have been introduced, ranging from basic
methods to incorporate (sequences of) sets of rules into a logic
program, to more elaborate methods which use an update policy for
specifying how updates must be incorporated. In this paper, we
introduce a framework for reasoning about evolving knowledge bases,
which are represented as extended logic programs and maintained by an
update policy. We first describe a formal model which captures various
update approaches, and we define a logical language for expressing
properties of evolving knowledge bases. We then investigate semantical
and computational properties of our framework, where we focus on
properties of knowledge states with respect to the canonical reasoning
task of whether a given formula holds on a given evolving knowledge
base. In particular, we present finitary characterizations of the
evolution for certain classes of framework instances, which can be
exploited for obtaining decidability results. In more detail, we
characterize the complexity of reasoning for some meaningful classes
of evolving knowledge bases, ranging from polynomial to double
exponential space complexity.
\end{abstract}

\noindent{\bf Keywords:} logic program updates, nonmonotonic knowledge
bases, knowledge base evolution, temporal reasoning, answer sets,
program equivalence, computational complexity

\newcounter{myenumctr}
\newenvironment{myenumerate}{\begin{list}{{\bf (\arabic{myenumctr})}\ }{\usecounter{myenumctr}
\setlength{\leftmargin}{0pt}
\setlength{\itemindent}{1.35\labelwidth}}}
{\end{list}}
\newenvironment{myitemize}{\begin{list}{$\bullet$}{\setlength{\leftmargin}{0pt}
\setlength{\itemindent}{\labelwidth}}}
{\end{list}}

\section{Introduction} 

Updating knowledge bases is an important issue in the area of data and
knowledge representation. While this issue has been studied
extensively in the context of classical knowledge bases (cf., \egc
\cite{wins-90,hbdru-3}), attention to it in the area of nonmonotonic
knowledge bases, in particular in logic programming, is more
recent. Various approaches to evaluating logic programs in the light
of new information have been presented.  The proposals range from
basic methods to incorporate an update $U$, given by a set of rules,
or a sequence of such updates $U_1,\ldots,U_n$, into a (nonmonotonic)
logic program $P$
\cite{alfe-etal-99a,foo-zhan-98,inou-saka-99,eite-etal-00f}, to more
general methods which use an {\em update policy} to specify, by means
of update actions, how the updates $U_1,\ldots,U_n$ should be
incorporated into the current state of knowledge
\cite{mare-trus-98,alfe-etal-99b,eite-etal-01}. Using these approaches, queries to
the knowledge base, like ``is a fact $f$ true in $P$
after updates $U_1,\ldots,U_n$?'', can then be evaluated.

Notably, the formulation of such queries is treated on an ad hoc
basis, and more involved queries such as ``is a fact $f$ true in $P$
after updates $U_1,\ldots,U_n$ and possibly further updates?'' are not
considered. More generally, reasoning about an evolving knowledge base
$\KB$, maintained using an update policy, is not formally
addressed. However, it is desirable to know about properties of the
contents of the evolving knowledge base, which also can be made part
of a specification for an update policy. For example, it may be
important to know that a fact $a$ is always true in $\KB$, or that a
fact $b$ is never true in $\KB$. Analogous issues, called {\em
maintenance} and {\em avoidance}, have recently been studied in the
agent community \cite{wool-00b}. Other properties may involve more
complex temporal relationships, which relate the truth of facts in
the knowledge base over time. A simple example of this sort is the
property that whenever the fact $\mathit{message\_to(tom)}$, which
intuitively means that a message should be sent to Tom, is true in
$\KB$ at some point, then the fact $\mathit{sent\_message\_to(tom)}$,
representing that a message has been sent to Tom, will be true in the
evolving knowledge base at some point in the future.

\medskip

\noindent{\bf Main problems addressed.} In this paper, we
aim at a framework for expressing reasoning problems over evolving
knowledge bases, which are modeled as extended logic programs
\cite{gelf-lifs-91} and may be maintained by an update policy as
mentioned above. In particular, we are interested in a logical
language for expressing properties of the evolving knowledge base,
whose sentences can be evaluated using a clear-cut formal semantics. The
framework should, on the one hand, be general enough to
capture different approaches to incorporating updates $U_1,\ldots,U_n$
into a logic program $P$ and, on the other hand, pay attention to the
specific nature of the problem. Furthermore, it should be possible to
evaluate a formula, which specifies a desired evolution behavior,
across different realizations of update policies based on different
definitions.

\medskip

\noindent{\bf Main results.} The main contributions and results of this paper
are summarized as follows.

\begin{myenumerate}
\item We introduce a formal model in which various approaches for
updating extended logic programs can be expressed.
In particular, we introduce the concept of an {\em evolution frame},
which is a structure $\EF=\tuple{\at,\EC,\AC,\up,\rf,\bel}$ whose
components serve to describe the evolution of knowledge states of an
agent associated with the knowledge base. This structure comprises
\begin{itemize}
\item a logic programming semantics, $\bel$, for extended logic
programs $P$, resp.\ sequences  $(P_1,P_2,\ldots,P_m)$ of  extended
logic programs $P_i$, over an alphabet $\at$; 

\item a nonempty class of {\em events}, which are sets of rules communicated to
the agent; and 

\item an update frame $\tuple{\AC,\up,\rf}$, consisting of a set of
update actions $\AC$, an update policy $\Pi$, and a realization assignment
 $\rf$, which together 
specify how to incorporate {\em events},
which are sets of rules drawn from a class of possible events $\EC$
and communicated to the agent, into the knowledge base.
\end{itemize}

In our framework, a {\em knowledge state} $s =
\tuple{\KB;E_1,\ldots,E_n}$ of the agent
consists of an initial knowledge base $\KB$, given by an extended
logic program over the alphabet $\at$, and a sequence of events
$E_1,\ldots,E_n$. Associated with the knowledge state $s$ is the {\em
belief set} $\bel(s)$ of the agent, which comprises all formulas which
the agent beliefs to hold given its state of knowledge.  

The agent reacts on an event by adapting its belief
state through the update policy $\up$, which singles out update
actions $A\subseteq \AC$ from a set of possible update actions $\AC$
for application. These update actions are executed, at a physical
level, by compilation, using the realization assignment $\rf$, into a single logic
program $P$, resp.\ a sequence of logic programs
$(P_0,\ldots,P_n)$, denoted $\compo_{\EF}(s)$. The
belief set $\bel(s)$ is then given by the
belief set of the compiled knowledge state, and is obtained by
applying the belief operator $\bel(\cdot)$ for (sequences of) logics
programs to $\compo_{\EF}(s)$. Suitable choices of $\EF$ allow one to
model different settings of logic program updates, such as
\cite{alfe-etal-99a,mare-trus-98,inou-saka-99,eite-etal-00f}.

\medskip
  
\item We define the syntax and, based on evolution frames, the
semantics of a logical language for reasoning about evolving knowledge
bases, 
which employs linear and branching-time operators familiar from
Computational Tree Logic (CTL)~\cite{emer-90}.  Using this language,
properties of an evolving knowledge base can be formally stated and
evaluated in a systematic fashion, rather than ad hoc. For example,
the maintenance problem from above can be expressed by a formula $\A\G\,a$,
and the avoidance problem by a formula $\A\G \fneg b$; the property
about Tom's messages is expressed by 
\[
\A\G(\mathit{message\_to(tom)} \rightarrow
\A\F\mathit{sent\_message\_to(tom)}).
\] 
  
\medskip

\item We investigate semantical properties of knowledge states
for reasoning. 
Since in
principle a knowledge base may evolve forever, we are in particular concerned with
obtaining finitary characterizations of evolution. To this end, we introduce
various notions of equivalence between knowledge states, and show several
filtration results: under certain properties of the components of
$\EF$, evolution of a knowledge state $s$ in an evolution frame $\EF$
can be described by a finite transition graph $G^\star(s,\EF)$, which
is a subgraph bisimilar to the whole natural transition graph
$G(s,\EF)$ over knowledge states that includes an arc from
$s_1=\tuple{\KB,E_1,\ldots,E_n}$ to every immediate successor state $s_2 =
\tuple{\KB,E_1,\ldots,E_n,E_{n+1}}$.
In some cases, $G^\star(s,\EF)$ is constructible by exploiting
locality properties of the belief operator $\bel(\cdot)$ and increasing
compilations $\compo_{\EF}(\cdot)$, while in others it results by
canonization of the knowledge states. 

In a concrete case study, we establish this
for evolution frames which model policies in the $\UL$ framework for
logic program updates using the answer set semantics
\cite{eite-etal-01}, as well as for the LUPS
\cite{alfe-etal-99b,alfe-etal-02} and LUPS$^\ast$ 
policies~\cite{leite-01} under the dynamic stable model 
semantics~\cite{alfe-etal-98,alfe-etal-99a}. Similar results apply to updates 
under other semantics in the literature.
                
\medskip

\item We derive complexity results for reasoning.  
Namely, we analyze the problem of deciding, given an evolution frame $\EF$, a knowledge state $s$, and a
formula $\varphi$, whether $\EF,s \models \varphi$ holds.  While this
problem is undecidable in general, we single out several cases in
which the problem is decidable, adopting some general assumptions
about the underlying evolution frame. In this way, we identify
meaningful
conditions under which the problem ranges from \PSPACE\ up to \TWOEXPSPACE\
complexity. 
We then apply this to the $\UL$ framework under the answer set semantics
\cite{eite-etal-01,eite-etal-02c},
and show that its propositional fragment has \PSPACE-complexity.
Similar results might be derived for the LUPS and
LUPS$^\ast$ frameworks.
We also consider the complexity of sequences of extended logic
programs (ELPs) and generalized logic programs (GLPs), respectively. We show 
that deciding whether two sequences
$\useq{P}=(P_0,\ldots,P_n)$ and $\useq{Q}=(Q_0,\ldots,Q_m)$ of
propositional ELPs are strongly equivalent under the update answer set
semantics, i.e., whether for every sequence $\useq{R} = (R_0,\ldots,R_k)$,
$k\geq 0$, the concatenated sequences $\useq{P}+\useq{R}$ and
$\useq{Q}+\useq{R}$ have the same belief sets, is
$\coNP$-complete. This result is not immediate, since potentially
infinitely many pairs $\useq{P}+\useq{R}$ and $\useq{Q}+\useq{R}$ need
to be checked. Thus, testing strong equivalence between
sequences of ELPs is not more expensive than standard inference of a
literal from all answer sets of an ELP (cf.\ \cite{dant-etal-01}). Analogous 
results hold  for sequences of GLPs.
\end{myenumerate}

To the best of our knowledge, no similar effort to formally express
reasoning about evolving nonmonotonic knowledge bases at a level as
considered here has been put forth so far. 
By expressing various approaches in our framework, we obtain a formal
semantics for reasoning problems in them. Furthermore, results about
properties of these approaches (e.g., complexity results) may be concluded
from the formalism by this embedding, as we illustrate for the $\UL$
framework. Note that J.~Leite, in his recent Ph.D.\ thesis
\cite{leite-02}, considers properties of evolving logic programs in a
language inspired by our $\UL$ language
\cite{eite-etal-01,eite-etal-02c}, and derives some properties for
dynamic logic programs similar to properties for update programs
derived in Section~\ref{sec:complexity}. 

The rest of this paper is structured as follows. In the next section,
we give some basic definitions and fix notation. In
Section~\ref{sec:evolution}, we introduce our notion of an evolution
frame, which is the basic setting for describing update formalisms, and in 
Section~\ref{sec:frameworks}, we show how different approaches to updating 
logic programs can be captured by it. In Section~\ref{sec:reasoning}, we then 
define the syntax and semantics of our logical language for reasoning about
evolving knowledge bases. Section~\ref{sec:equivalence} is devoted to
the study of equivalence relations over knowledge states, which are
useful for filtration of the infinite transition graph that arises
from an evolving knowledge base. In particular, conditions are
investigated under which a restriction to a finite subgraph is
feasible. After that, we address  in Section~\ref{sec:complexity}  the
complexity of reasoning. Related work 
is discussed in Section~\ref{sec:conclusion}, where we also 
draw some conclusions and outline issues for further research.

\section{Preliminaries}\label{prel}

We consider knowledge bases represented as \emph{extended logic
programs} (ELPs) \cite{gelf-lifs-91}, which are finite sets of rules 
built over a first-order alphabet $\at$ using default negation $\naf$
and strong negation $\neg$. A rule has the form
\begin{equation}\label{eq:rule}
r:\quad L_0 \la L_1,\ldots, L_m, \naf
L_{m+1},\ldots,
\naf L_n,
\end{equation}
where each $L_i$ is a literal of form $A$ or $\neg A$, where $A$ is an
atom over $\at$. 
For a literal $L$, the \emph{complementary literal}, $\cl{L}$, is $\neg A$ 
if $L=A$, and $A$ if $L=\neg A$, for some atom $A$. 
For a set $S$ of literals, we define $\cl{S}=\{ \cl{L} \mid L \in S\}$.
We also denote by $\extat$ the set $\at \cup \cl{\at}$ of all literals 
over $\at$. 
The set of all rules is denoted by $\lang_\at$. We
call $L_0$ the \emph{head} of $r$ (denoted by $\head{r}$), and the set
$\{L_1,\ldots, L_m, \naf L_{m+1},\ldots, \naf L_n \}$ the \emph{body}
of $r$ (denoted by $\body{r}$).  We define $\bodyp{r}=\{L_1,\ldots,
L_m\}$ and $\bodyn{r}=\{L_{m+1},\ldots, L_n\}$. We allow the case
where $L_0$ is absent from $r$, providing $\body{r}\neq\emptyset$; such a rule $r$ is called a
\emph{constraint}. If $\head{r}=\{L_0\}$ and $\body{r}=\emptyset$, 
then $r$ is called \emph{fact}.  We often write $L_0$ for a fact $r=L_0\la$. 
Further extensions, \egc $\naf$ in the rule head~\cite{alfe-etal-99a}, might
be added to fit other frameworks.

An \emph{update program}, $\useq{P}$, is a sequence $(P_0,\ldots,P_n)$
of ELPs ($n\geq 0$), representing the evolution of program $P_0$ in
the light of new rules $P_1,\ldots,P_n$.  We sometimes use $\cup\useq{P}$ to 
denote the set of all rules occurring in $\useq{P}$, \iec 
$\cup\useq{P}=\bigcup_{i=1}^n P_i$.
The semantics of update
programs can abstractly be described in terms of a belief operator $\bel(\cdot)$, which
associates with every sequence $\useq{P}$ a set $\bel(\useq{P})
\subseteq \lang_\at$ of rules, intuitively viewed as the consequences
of $\useq{P}$. $\bel(\cdot)$ may be instantiated in terms of various
proposals for update semantics, like, \egc the approaches described in
\cite{alfe-etal-99a,foo-zhan-98,inou-saka-99,eite-etal-00f,mare-trus-98}.

\subsection{Update answer sets} 

For concrete examples, we consider the answer set semantics for
propositional update programs as introduced
in~\cite{eite-etal-00f,eite-etal-00g}, as well as the semantics for dynamic 
logic programs  
as defined in~\cite{alfe-etal-99a,leite-02}. The former semantics defines 
answer sets of a sequence of ELPs, $\useq{P}=(P_0,\ldots,P_n)$, in terms of 
answers sets of a 
single ELP $P$ as follows. An {\em interpretation}, $S$, is a set of classical
literals containing no opposite literals $A$ and $\neg A$.  The
\emph{rejection set}, $\rs(S,\useq{P})$, of $\useq{P}$ with respect to
an interpretation $S$ is
\(
\textstyle\rs(S,\useq{P}) = \bigcup_{i=0}^n \rs_i(S,\useq{P}),
\)
where
$\rs_n(S,\useq{P})=\emptyset$, and, for $n > i\geq 0$,
$ \rs_i(S,\useq{P})$ contains every rule $r\in P_i$ such that
$\head{r'}=\neg\head{r}$ and $S \models
\body{r}\cup\body{r'}$, for some
$r' \in P_j\setminus \rs_j(S,\useq{P})$ with $j>i$.  That is,
$\rs(S,\useq{P})$ contains the rules in $\useq{P}$ which are rejected
by unrejected rules from later updates. Then, an interpretation $S$ is
an \emph{answer set} of $\useq{P}=(P_0\commadots P_n)$ iff $S$ is a
consistent answer set \cite{gelf-lifs-91} of the program $P =
\bigcup_{i}P_i\setminus\rs(S,\useq{P})$, \iec $S$ is a minimal
consistent set of literals closed under the rules of the {\em reduct\/} 
$P^S =
\{ \head{r} \la \bodyp{r} \mid r \in P \ \textrm{and}\ \bodyn{r} \cap S
= \emptyset \}$. 
The
set of all answer sets of $\useq{P}$ is denoted by $\ASup(\useq{P})$. 
This definition properly generalizes consistent answer sets from
single ELPs to sequences of ELPs. Nevertheless, we use $\AS(P)$ to denote the 
set of all answer sets of a single ELP $P$. Moreover, an ELP $P$ is called 
\emph{inconsistent} if it has no consistent answer set, \iec $\AS(P)=\emptyset$.
Update answer sets for arbitrary
(non-ground) update programs $\useq{P}$ are defined in terms of its
ground instance similar as answer sets for ELPs \cite{gelf-lifs-91}.

\begin{example}
Let $P_0= \{ b\la \naf a,$ $a\la\}$, $P_1 = \{ \neg a\la,$
$c\la\}$, and $P_2=\{ \neg c\la\}$. Then, $P_0$ has the single
answer set $S_0=\{ a\}$ with $\rs(S_0,P_0)=\emptyset$; $( P_0,P_1)$
has answer set $S_1=\{\neg a, c, b\}$ with
$\rs(S_1,(P_0,P_1))=\{a\la\,\}$; and $(P_0,$ $P_1,$ $P_2)$ has
the unique answer set $S_2=\{\neg a, \neg c, b\}$ with
$\rs(S_2,(P_0,P_1,P_2))=\{ c\la,a\la\}$.
\end{example}

The belief operator $\bel_E(\cdot)$ in the framework of \cite{eite-etal-00f} is given by $\bel_E(\useq{P}) = \{ r \in
\lang_\at \mid S\models r$ for all $S \in \ASup(\useq{P}) \}$, 
where $S\models r$ means that for each ground instance $r'$ of $r$,
either $\head{r'}\!\in S$, or $L\!\notin\!S$ for some $L
\in \body{r'}$, or $L\!\in S$ for some $\naf L \in \body{r'}$.

\subsection{Dynamic answer sets}

By the term \emph{dynamic answer sets} we refer to the extension of
dynamic stable models, defined for sequences of \emph{generalized
logic programs} (\emph{GLPs}) by Alferes \emph{et
al.}~\cite{alfe-etal-99a}, to the three-valued case. In GLPs, default
negation may appear in the head of rules, but strong negation is
excluded. The definition of dynamic stable models uses a slightly
non-standard concept of stable models, where weakly negated literals
$\naf A$ ($A$ some atom) are treated like ordinary propositional
atoms, and rules $A_0\la A_1\commadots A_m,$ $\naf
A_{m+1}\commadots \naf A_n$ are viewed as
\emph{Horn clauses}.  Accordingly, an interpretation
$I$ is in this context understood as a set of atoms and
weakly negated atoms such that $A\in
I$ iff $\naf A\notin I$ holds for each atom $A$.  To distinguish it from
a usual interpretation, we call it a \emph{generalized
interpretation}. As usual, a set $B$, comprising atoms and weakly
negated atoms, is true in a generalized interpretation $I$,
symbolically $I\models B$, iff $B\subseteq I$.  Towards defining
stable models, the following notation is required:

Let, for a set of atoms  $\at$, $\naf \at$ stand for the set 
$\{\naf A\mid A\in \at\}$. Furthermore, for $M\subseteq\at\cup\naf\at$, we set 
$M^-=\{\naf A\mid\naf A\in M\}$, and, for $Z\in\at\cup\naf\at$, we define 
$\naf Z=\naf A$ if $Z=A$ and $\naf Z=A$ if $Z=\naf A$.
For a program $P$ over $\at$, the deductive closure, $\Cni{\at}{P}$, 
is given by the set 
$
\{L\mid\mbox{$L\in\at\cup\naf\at$ and $P \vdash L$}\},
$
where $P$ is interpreted as a propositional Horn theory and ``$\vdash$'' 
denotes classical derivability. 
A generalized interpretation $S$ is a stable model of a program $P$ iff $S=\Cni{\at}{P\cup S^-}$.

Let $\useq{P}=(P_0\commadots P_n)$ be a sequence of GLPs over $\at$, and let 
$I$ be a generalized interpretation. 
Alferes \emph{et al.}~\cite{alfe-etal-99a} introduce the following concepts:
\begin{tabbing}
XX\=\kill
\> $\mathit{Rejected}(I,\useq{P})$ = \= $\bigcup_{i=0}^n \{ r \in P_i
\mid$ \= $\exists r' \in P_j, \textrm{ for some } j\in \{i+1,\ldots, n\}, \textrm{ such }$\\
\>\>\>  \textrm{that } $\head{r'} = \naf \head{r} \textrm{ and } I\models 
\body{r}\cup\body{r'}\}$; \\
\>\> $\mathit{Defaults}(I,\useq{P})$ =\' $\{ \naf A \mid \not\exists
r$ in $\useq{P}$ such that $\head{r}=A$ and $I\models\body{r} \}$. 
\end{tabbing}

A set $S\subseteq \at\cup\naf\at$ is a {\em dynamic stable model}\/ of 
$\useq{P}$ iff
\[
S=\Cn{(\cup\useq{P} \setminus \mathit{Rejected}(S,\useq{P})) \cup 
\mathit{Defaults}(S,\useq{P})}.
\]
 We remark that Alferes {\em et al.}  
defined dynamic stable models of $\useq{P}$ as projections
$S= S'\cap (\at \cup \naf \at)$ of the stable models of a single GLP, 
$\useq{P}_\oplus$, resulting from a transformation (for a detailed definition 
cf.~\cite{alfe-etal-99a}), and then proved the above characterization as a 
result.

Alferes \emph{et al.}~\cite{alfe-etal-99a} defined also an extension of their semantics to the
three-valued case: Let
$\useq{P}=(P_0,\ldots,P_n)$ be a sequence of ELPs over $\at$. Then, a consistent set $S
\subseteq \extat$ 
is a {\em dynamic answer set} of $\useq{P}$ iff $S \cup \{ \naf L \mid L \in \extat\setminus S\}$ is a
dynamic stable model of the sequence  $\useq{P} =
 (P_0,\ldots,P_n\cup \{ \naf A \la \neg A,\, \naf\neg A\la A \mid A
\in \at\})$ of GLPs.  Here, the rules in $\{ \naf A \la \neg A,\, \naf\neg
A\la A \mid A \in \at\}$ serve for emulating classical negation
through weak negation.

\begin{example}
Let, as in the previous example,  $P_0= \{ b\la \naf a,$ $a\la\}$, $P_1 = \{ \neg a\la,$
$c\la\}$, and $P_2=\{ \neg c\la\}$ over $\at = \{a,b,c\}$. Then, $P_0$ has the single
dynamic answer set $S_0=\{ a, \naf b, \naf c\}$, where $\mathit{Rejected}(S_0,P_0)=\emptyset$ and
$\mathit{Defaults}(S_0,P_0)=\{\naf b, \naf c\}$; the sequence $(P_0,P_1)$
has the dynamic answer set $S_1=\{\neg a, c, b\}$, where
$\mathit{Rejected}(S_1,(P_0,P_1))=\{\,a\la\,\}$ and
$\mathit{Defaults}(S_1,(P_0,P_1))=\{\naf\neg b$, $\naf\neg c\}$; and $(P_0,$ $P_1,$ $P_2)$ has
the single dynamic answer set $S_2=\{\neg a, \neg c, b\}$, where 
$\mathit{Rejected}(S_2,$ $(P_0,P_1,P_2))=\{\,a\la\,,\, c\la\,\}$ and
$\mathit{Defaults}(S_2,(P_0,P_1,P_2))=\{\naf\neg b\}$. Note that in
these simple examples, update answer sets and dynamic answer sets
coincide, which is not the case in general \cite{eite-etal-00g}.
\end{example}
 
Similarly to the belief operator $\bel_E(\cdot)$, we can define a belief 
operator $\bel_{\oplus}(\cdot)$ for dynamic stable models as 
$\bel_{\oplus}(\useq{P}) = \{ r \in \lang_\at \mid S\models r$ for all 
$S \in \SMup(\useq{P}) \}$, where $\SMup(\useq{P})$ denotes the set of all 
dynamic stable models of $\useq{P}$.

Finally, we remark that while we defined answer sets and belief sets for 
sequences of finite programs, they can be defined for 
sequences of possibly infinite programs in an analogous way. 

\section{Knowledge-Base Evolution}\label{sec:evolution}

We assume that the agent has an initial knowledge base, $\KB$, in form of
an extended logic program, and a background \emph{update policy},
$\up$, describing the update behavior of the agent, \iec how it has to
react when it receives new information from the
environment. Information arrives to the agent in form of a sequence of
\emph{events}, each event being a finite set of rules from a given
\emph{event class}. The update policy specifies what rules or facts
have to be incorporated into or retracted from the knowledge base,
depending on the content of the event and on the belief set of the agent. The evolution of the agent's
\emph{knowledge state} is thus completely described when $\KB$ and a
sequence of events $E_1,\ldots,E_n$ are given, provided an update
policy $\up$ is specified.

\subsection{Events and knowledge states}
\label{sec:defs}

We start with the basic formal notions of an \emph{event} and of the
\emph{knowledge state} of an agent maintaining a knowledge base.

\begin{definition}
Let $\at$ be some alphabet. An \emph{event class over $\at$} $($or simply \emph{event class}, if no ambiguity arises\/$)$ is a
collection $\EC \subseteq 2^{\lang_\at}$ of finite sets of rules. The
members $E\in \EC$ are called \emph{events}.
\end{definition}

Informally, $\EC$ describes the possible events (\iec sets of
communicated rules) an agent may experience. In the most general case,
an event is an arbitrary ELP; a plain case is that an event just
consists of a set of facts, which are formed over a subset of the
alphabet.
In a deductive database setting, the latter case corresponds to an
extensional database that is undergoing change while the intensional
part of the database remains fixed.

\begin{definition}
Let $\EC$ an event class over some alphabet $\at$. A \emph{knowledge
state over $\EC$} $($simply, a \emph{knowledge state}$)$ is a tuple
$s=\tuple{\KB;E_1,\ldots , E_n}$, where $\KB\subseteq\lang_\at$ is an
ELP $($called \emph{initial knowledge base)} and each $E_i$ $(1\leq
i\leq n)$ is an event from $\EC$. The \emph{length of} $s$, denoted
$|s|$, is $n$.  The set of all knowledge states over $\at$ given $\EC$
is denoted by $\KS(\EC)$.
\end{definition}

Intuitively, $s=\tuple{\KB;E_1,\ldots,E_{n}}$ captures the agent's
knowledge, starting from its initial knowledge base. When a new event
$E_{n+1}$ occurs, the current knowledge state $s$ changes to
$s'=\tuple{\KB;E_1,\ldots,E_{n},E_{n+1}}$, and the agent should
adapt its belief set in accordance with the new event obeying its
given update policy.

\subsection{Evolution frame}
\label{sec:frame}

The ``universe'' in which the evolution of an agent's knowledge base
takes place is given by the concept of an \emph{evolution frame},
which comprises different components that parameterize the update
mechanism and the semantics used on the evolving knowledge base. 
This structure comprises, together with an alphabet $\at$, 
\begin{itemize}
\item a semantics, $\bel(\cdot)$, for ELPs, resp.\ sequences  of ELPs,
over $\at$; 

\item a nonempty event class $\EC$ over $\at$; and, 

\item an {\em update frame} $\tuple{\AC,\up,\rf}$, consisting of a set of
\emph{update commands} $\AC$, an \emph{update policy} $\Pi$, and a \emph{realization assignment} $\rf$, which together specify how to incorporate events into the knowledge base.
\end{itemize}

In more detail, the components of an update frame are as follows.  

\paragraph{Update commands.} The {update commands} (or
\emph{actions}) in $\AC$ are names for commands which are supposed to
be executed on the knowledge base. Simple, elementary update commands are
$\mathit{insert}(r)$ and $\mathit{delete}(r)$, which add and remove a
rule to a logic program, respectively, without a sophisticated
semantics handling potential inconsistencies (which may be delegated
to the underlying update semantics). More involved update commands
have been proposed in the literature (cf., \egc
\cite{alfe-etal-99b,eite-etal-01}). However, several update frameworks
can be modeled using these simple commands. The semantics (i.e., effects) of update
actions are given by the realization assignment, $\rf$, which is
described below. 

\paragraph{Update policy.} The {update policy} 
$\up$, which is a function mapping every pair $(s,E)$ of a knowledge state $s$
over $\EC$ (\iec $s \in \KS(\EC)$) and 
an event $E\in\EC$ into a set $\up(s,E)\subseteq\AC$ of update
commands, determines {\em which} actions should be executed.
Update policies allow for specifying sensible and flexible ways
to react upon incoming events. A very simple policy is
$\up_{\mathit{ins}}(s,E) = \{ \mathit{insert}(r) \mid r \in E\}$; it models an 
agent which incorporates the new information unconditionally.
More sophisticated policies may define exceptions for the incorporation of 
rules from events, or the insertion of rules may be conditioned on
the belief in other rules.

\paragraph{Realization assignment.} The {realization
assignment} $\rf$ assigns to each pair $(s,A)$ of a knowledge state $s$ over $\EC$
and a set $A\subseteq\AC$ of update commands a sequence
$\rf(s,A)=(P_0\commadots P_n)$ of ELPs $P_i$ over $\at$ $(0\leq i\leq
n)$. It associates in this way a meaning with the set of actions $A$
which must be executed on the knowledge state $s$, in terms of an ELP,
resp.\ a sequence of ELPs, and ``realizes'' the update in this
way. The agent's beliefs from the updated knowledge base may then be
given by the operator $\bel(\cdot)$ applied to the result of
$\rf(s,A)$ as defined in Section~\ref{sec:comp-belief} below.

Different possibilities for concrete realization assignments $\rf$ may
be used.  A simple realization assignment, $\rf_\pm(s,A)$, which works
for sets $A$ of actions of form $\mathit{insert(r)}$ and $\mathit{delete(r)}$,
and assumes that each knowledge state $s$ is assigned with 
an ordinary
ELP $P(s)$, is given by 
$$\rf_{\pm}(s,A) = (P(s) \cup \{ r \mid \mathit{insert(r)} \in A\})
\setminus \{ r \mid \mathit{delete(r)}\in A\},$$ i.e., the insertion
and deletion commands in $A$ are ``physically'' implemented, with no
further enforcement that consistency is preserved, or, as for deletion,
that $r$ is actually logically deleted from the knowledge base.
Its
restriction to insertion commands is the realization assignment
$\rf_{\mathit{ins}} = (s,A) = P(s) \cup \{ r \mid \mathit{insert(r)}
\in A\}$, which may be used in contexts where data are not physically
removed, for whatever reasons.

More sophisticated realization assignments might block, at the
logical level, the applicability of rules in the knowledge base, by
using a sequence $(P_0,\ldots,P_n)$ of ELPs as a representation, and
aim at enforcing consistency of the knowledge base. For instance, in the
dynamic logic programming semantics of sequences of ELPs in \cite{alfe-etal-99a,eite-etal-01},
more recent rules occur later in a
sequence and 
override rules from programs 
which occur earlier in the sequence;
this mechanism is also
used
in the $\UL$ framework for incorporating changes to the knowledge base
at the logical level \cite{eite-etal-01,eite-etal-02c}.

\medskip

In summary, we formally define an evolution frame as follows. Let,
for any alphabet $\at$, denote $\mathit{ELP}^\ast(\at)$ the set of all
sequences $\useq{P} =(P_0,\ldots,P_n)$, $n\geq 0$, of ELPs $P_i$ over~$\at$.  

\begin{definition}
An \emph{evolution frame} 
is a tuple $\EF=\tuple{\at,\EC,\AC,\up,\rf,\bel}$, where
\begin{itemize}
\item $\at$ is a finite $($first-order\/$)$ alphabet;
\item $\EC$ is a nonempty event class over $\at$;
\item $\AC$ is a set of {update commands} $($or {actions}$)$;
\item $\up : \KS(\EC) \times\EC\rightarrow 2^{\mbox{\scriptsize\AC}}$ is an 
{update policy};
\item $\rf : \KS(\EC) \times 2^{\mbox{\scriptsize\AC}}\rightarrow \mathit{ELP}^\ast(\at)$ 
is a {realization assignment}; and
\item $\bel : \mathit{ELP}^\ast(\at)\rightarrow 2^{\lang_\at}$ is a belief 
operator for sequences of ELPs.
\end{itemize}
The set of all knowledge states in $\EF$, denoted by $\mathcal{S}_{\EF}$, is given by $\KS(\EC)$.
\end{definition}

The concept of an evolution frame allows us to model
various update approaches, as we discuss below in Section~\ref{sec:frameworks}.

\subsection{Compilation and belief set}
\label{sec:comp-belief}

While $\up$ determines {\em what} to do, the realization assignment
$\rf$ states {\em how} this should be done. Informally, $\rf(s,A)$
``executes'' actions $A$ on the knowledge state $s$ by producing a
logic program $P$ or, in general, a sequence of logic programs $\useq{P}$. 
We can use $\rf$ to ``compile'' a knowledge state $s$ into a (sequence of) logic
programs, by determining the set of actions $A$ from the last event
in $s$.  We introduce the following notation.

For any knowledge
state $s=\tuple{\KB;E_1,\ldots , E_n}$ over $\EC$, denote by $\pi_i(s)
= \tuple{\KB;E_1,\ldots,E_i}$ its projection to the first $i$ events,
for $0\leq i\leq n$.
In particular, $\pi_0(s)$ is the initial knowledge base $\KB$. 
We call $\pi_i(s)$ a
\emph{previous knowledge state} (or simply an \emph{ancestor}) of
$s$ if $i<n$. Dually, a knowledge state $s'$ over $\EC$ is \emph{a future
knowledge state} (or simply a \emph{descendant}) of $s$ if $s$ is
previous to $s'$. Furthermore, $\pi_{n-1}(s)$ is the
\emph{predecessor} of $s$, and $s'$ is a \emph{successor} of $s$, if
$s$ is predecessor of $s'$. Finally, for events $E_1'\commadots E_m'$,
we write $s+E_1'\commadots E_m'$ to denote the concatenated knowledge
state $\tuple{\KB;E_1,\ldots , E_n,E_1'\commadots E_m'}$; a similar
notation is used for the concatenation of sequences of logic programs.

\begin{definition}\label{comp}
Let $\EF=\tuple{\at,\EC,\AC,\up,\rf,\bel}$ be an evolution frame.
For any knowledge state
$s=\tuple{\KB;E_1\commadots E_{n}}$ over $\EC$, the \emph{compilation associated with $s$}
is
\[
\compo_{\EF}(s)=
\left\{
\begin{array}{ll}
\rf(s,\emptyset), & \textrm{if $|s|=0$, \iec $s=\tuple{\KB}$},  \\
\rf(\pi_{n-1}(s),\up(\pi_{n-1}(s),E_n)), & \textrm{otherwise}. 
\end{array}
\right.
\]
\end{definition}

Note that $\compo_{\EF}(\cdot)$ is a function which is fully determined
by $\EF$; we often write $\comp{\cdot}$
instead of $\compo_{\EF}(\cdot)$ if $\EF$ is understood.

This definition of compilation is fairly general. It first 
computes the actions for the latest event $E_n$, and then requires
that these actions are executed on the predecessor state.
Observe that, in view of $\compo_{\EF}(s)$, we could equally well model update policies as unary functions $\hat{\Pi}(\cdot)$ such that $\hat{\Pi}(s)=\up(\pi_{n-1}(s),E_n)$.
However, we chose binary update policies to stress the importance of
the last event in $s$. Furthermore, $\up$ may be restricted in the
compilation process, \egc such that only the belief set
$\bel(\pi_{n-1}(s))$ of the predecessor state is considered rather
than the whole state itself; this will be considered in Section~\ref{sec:canonical}.

\paragraph{Incremental Compilation.} 
An important class of compilations are those in which, for
a future knowledge state $s'$, $\comp{s'}$  results by appending some further
elements to the sequence $\comp{s}$ of logic programs for the current
knowledge state $s$. This motivates the following notion:

\begin{definition}
Given an evolution frame $\EF=\tuple{\at,\EC,\AC,\up,\rf,\bel}$, 
$\compo_{\EF}(\cdot)$ is 
\emph{incremental} iff, for each 
$s=\tuple{\KB;E_1,\ldots , E_n}$,
$\compo_{\EF}(s)=(P_0\commadots P_n)$ such that
$\rf(\tuple{\KB},\emptyset) = P_0$ and
$\rf(\pi_{i-1}(s),\up(\pi_{i-1}(s),E_i)) = (P_0,\ldots,P_i)$ for $1\leq i\leq n$.
\end{definition}

This definition amounts to the expected behavior:
\begin{proposition}
The mapping $\compo_{\EF}(\cdot)$ is incremental iff, for each
knowledge state $s$,  
$\compo_{\EF}(s) = Q$ if $|s|=0$, 
and
$
\compo_{\EF}(s) = \compo_{\EF}(\pi_{|s|-1}(s))+ Q'
$
otherwise, where $Q, Q'$ are logic programs and ``+'' is the
concatenation of sequences.
\end{proposition}

\begin{proof} The proof proceeds by straightforward induction on $|s|$.
\hfill$\Box$
\end{proof}

\begin{example}
A simple incremental compilation results 
for $\AC_{\mathit{ins}} =
\{ \mathit{insert}(r) \mid r \in \lang_\at\}$, $\up =
\up_{\mathit{ins}}$ as defined in Subsection~\ref{sec:frame}, and
$\rf_{\mathit{ins}}$ such that $\compo_{\EF}(\tuple{\KB}) = \KB$ and 
$\compo_{\EF}(s)=\compo_{\EF}(\pi_{|s|-1}(s)) + (\{r \mid \mathit{insert}(r)\in
A\})$, where $A=\up_{\mathit{ins}}(\pi_{|s|-1}(s),E_n)$,
given that $s=\tuple{\KB;E_1,\ldots , E_n}$. Note that
$\compo_{\EF}(s)$ is in this setting just the sequence $(\KB,E_1,\ldots,E_n)$.
\end{example}

While incremental compilations are natural,
we stress that other compilations are of course also highly relevant. In particular,
the compilation might perform optimizations (cf.\
Section~\ref{sec:canonical}), or output only an ordinary logic program. 

We also point out that our notion of incremental compilation should
not be confused with an {\em iterative compilation}; such a
compilation would, similar in spirit, consider the events $E_i$ in a
knowledge state $s=\tuple{\KB,E_1,\ldots,E_n}$ in their chronological
order one by one and instantaneously incorporate updates
$A_i=\up(\pi_{i-1}(s),E_i)$ into the result $\compo_{EF}(\pi_{i-1}(s))$
for the previous knowledge state and return a single, ordinary logic program as
the result.

The compilation of a knowledge state into a (sequence of) ELPs is
used, via the semantics $\bel(\cdot)$ for sequences of ELPs, to
ascribe a set of beliefs to the agent in the respective knowledge
state. More formally, the belief set emerging from a knowledge state
is as follows.

\begin{definition}
Let $\EF=\tuple{\at,\EC,\AC,\up,\rf,\bel}$ be an evolution frame and
$s$ a knowledge state. The belief set of
$s$, denoted $\bel(s)$, is given by $\bel(\compo_{\EF}(s))$.
\end{definition}

This completes the exposition of evolution frames and their
semantics. Before we consider some examples, let us close this
subsection with some remarks. 

\paragraph{Remarks.} (1) As mentioned earlier, our definition of an
update policy, and similarly of a realization assignment, which
effectively lead to the notion of a compilation, is very general. We
may stipulate additional postulates upon them, like the
incrementability property or an iterativity property (which we omit
here). Likewise, the concept of a semantics $\bel(\useq{P})$ for
sequences $\useq{P}$ of ELPs is very abstract, and further axioms and
conditions could be imposed on it. An example of this is the requirement
that $\bel(\useq{P})$ is characterized by rules of bounded length, and
in particular by rules without repeated literals;
this will be the case in
Section~\ref{sec:complexity}.

(2) Our definition does not capture nondeterministic update policies,
where $\up(s,E)$ may return one out of several possible sets of update
actions.
In order to model this, the notion of a knowledge state can be extended by taking
previous actions into account, \iec a knowledge state $s$ is then of the form
$\tuple{\KB,(E_1,A_1),\ldots,(E_n,A_n)}$, where each $E_i$ is an
event, and $A_i$ is the set of update commands 
executed at step $i$.
In practice, we may assume a suitable
\emph{selection function} $\sigma$, which chooses
one of the possible outcomes of $\up(s,E)$, and we are back to a
deterministic update policy $\up_\sigma$. If the selection function
$\sigma$ is unknown, we may consider all evolution frames
$\EF_\sigma$ arising for each $\sigma$.

\subsection{Examples}
\label{sec:examples}

Let us illustrate our framework on two examples, which serve as 
running examples throughout the remainder of the paper.

\begin{example}[Shopping Agent]
\label{exa:shop-agent} 
Consider a shopping agent selecting Web shops in search for some specific merchandise. Suppose its knowledge base, $\KB$,
contains the rules
$$
\begin{array}{lr@{~}l@{~}l}
r_1 : & \mathit{query}(S) & \la & \mathit{sale}(S), up(S), \naf \neg \mathit{query}(S); \\[0.5ex]
r_2 : & \mathit{site\_queried} & \la & \mathit{query}(S); \\[0.5ex]
r_3 : & \mathit{notify}    & \la & \naf \mathit{site\_queried}; 
\end{array}
$$
\vspace{-.4em}\noindent
and a fact $r_0:\mathit{date}(0)$ as an initial time stamp. 
Here, $r_1$ expresses that a shop $S$, which has a sale and whose Web site is 
up, is queried by default, and $r_2$, $r_3$ serve to detect that no site is 
queried, which causes `$\mathit{notify}$' to be true. 

Assume that an event, $E$, might consist of one or more of the following 
items:
\begin{itemize}
        \item at most one fact  $\mathit{date(t)}$, for some date $t$;
        \item facts $\mathit{up}(s)$ or $\neg\mathit{up}(s)$, 
              stating that a shop $s$ is up or down, respectively;
        \item ground rules of form 
              $\mathit{sale}(s) \la \mathit{date}(t)$, stating that shop $s$ 
              has a sale on date $t$.
\end{itemize}

An update policy $\UP$  may be defined as follows: 
{\small
\begin{tabbing}
XX\=\kill
\> $\up(s,E)$ = \= $\{ \mathit{insert(\alpha) \mid \alpha \in
\{up(S),\neg up(S), date(T)\},\alpha\in E} \}$ $\cup$ \+\+ \\
  $\{ \mathit{insert(sale(S) \la date(T))},\ \mathit{insert(track(S,T))} \mid$  \\
 \qquad $\mathit{sale(S) \la date(T) \in E}, \mathit{date(T')\in \bel(s), T\geq T'} \}$ $\cup$\\
  $\{ \mathit{delete(track(S,T))},\ \mathit{delete(sale(S) \la date(T))} \mid$ \\
 \qquad $\mathit{date(T') \in E, track(S,T) \in \bel(s)}, \mathit{date(T) \in \bel(s), T'\neq T}\}$ $\cup$\\
 $\{ \mathit{delete(date(T)) \mid date(T') \in E, date(T) \in \bel(s), T'\neq T}\}\,\}.$
\end{tabbing}
}

Informally, this update policy incorporates information about future
sales, only. The information of the sale is removed, when the sale
ends (assuming the time stamps increase). To this end, facts
$\mathit{track}(S,T)$ are used to keep track of inserted sale
information. Similarly, the current time stamp $\mathit{date}(t)$ is
maintained by deleting the old values.  The realization assignment
$\rf$ might be $\rf_{\pm}$ from Subsection~\ref{sec:frame}, which
always returns a single ELP, and for $\bel$ we may take any function
which coincides on sequences $\useq{P}=P_0$ of length one with the
standard answer set semantics for ELPs. Or, we might choose
a realization assignment which maps $s$ and a set of
$\mathit{insert(r)}$ and $\mathit{delete(r)}$ commands to a sequence $(P_0,\ldots,P_n)$ of ELPs, using as $\bel$ the answer set semantics for
sequences of ELPs as discussed in Subsection~\ref{sec:frame}.
\end{example}

\begin{example}[Mail Agent]
\label{exa:mail-agent} Consider a more complex mail agent,
which has the following initial knowledge base $\KB$, whose rules are
instantiated over suitable variable domains:
{\small
$$
\begin{array}{rrcl}
r_1: & \mathit{type(M,private)} & \la & \mathit{from(M,tom)} ;\\ 
r_2: & \mathit{type(M,business)} & \la & \mathit{subject(M,project)};\\ 
r_3: &   \mathit{type(M,other)} & \la & \naf \mathit{type(M,private),\naf type(M,business), msg(M)}; \\
r_4: &        \mathit{trash(M)} & \la & \mathit{remove(M), \naf save(M)}; \\
r_5: &       \mathit{remove(M)} &\la & \mathit{date(M,T),today(T'),\naf save(M)}, T' >(T+30); \\
r_6: &        \mathit{found(M)} & \la & \mathit{search(T),type(M,T), \naf trash(M)}; \\
r_7:&         \mathit{success} & \la & \mathit{found(M)}; \\ 
r_8:&         \mathit{failure} & \la & \mathit{search(T), \naf success}.
\end{array}
$$
}
\nop{**** old, short
The knowledge base contains rules about classifying message types
($r_1$--$r_3$), trash and removal of mails ($r_4$, $r_5$), and further rules ($r_6$--$r_8$)
to determine success or failure of a search for messages of a particular
type. 
}
The knowledge base allows to express several attributes of a message
and determine the $\mathit{type}$ of a message based on these attributes
(rules $r_1$ and $r_2$). By means of $r_3$, a default type is assigned
to all messages which are neither $\mathit{private}$ nor $\mathit{business}$. 
Rule $r_4$ implicitly states that a $\mathit{save}$ operation is stronger
than a $\mathit{remove}$ one. Note that in this way, once a message has been
saved, it can never be removed. By means of $r_5$, all those messages
are removed which have not been saved and are older than thirty
days. Rules $r_6$, $r_7$ and $r_8$ are used to look for all messages
of a given type, which have not been sent to the trash yet, and to
signal if at least one such message has been found ($\mathit{success}$) or not
($\mathit{failure}$).

Suppose that an event $E$ may consist in this scenario of one or more of the following items:
\begin{itemize}
        \item at most one fact  $\mathit{today(d)}$, for some date $d$;
        \item a fact $\mathit{empty\_trash}$, which causes  messages in the trash to be eliminated;
        \item facts  $\mathit{save(m)}$ or $\mathit{remove(m)}$, for mail identifiers $m$;
        \item at most one fact  $\mathit{search(t)}$, for some mail type $t\in \{ \mathit{other, business, private}\}$;
        \item zero or more sets of facts $\mathit{from(m,n),subject(m,s)}$, or $\mathit{date(m,d)}$ for mail identifier $m$, name $n$, subject $s$, and date $d$.
\end{itemize}

The update policy $\up$ may be as follows:
{\small
\begin{tabbing}
XX\=\kill
\> $\up(s,E)$ = \= $\{ \mathit{insert(R)} \mid R\in E \} \cup \{ \mathit{insert(msg(M))} \mid
\mathit{from}(M,N) \in E\}$ $\cup$ \+\\
\> $\{ \mathit{delete(today(D)) \mid today(D') \in E, today(D) \in
\bel(s), D'\neq D}\}$ $\cup$\\
\> $\{ \mathit{delete(\alpha)} \mid$ \= $\mathit{\alpha \in \{ trash(M),msg(M), type(M,T)}\}$, \\
\>\> \quad $\mathit{empty\_trash \in  E,  trash(M)\in \bel(s)}\}$ $\cup$ \\
\> $\{ \mathit{delete(\alpha) \mid \alpha \in \{from(M,N),subject(M,S), date(M,D)}\}$,\\
\>\> \quad $\mathit{save(M) \notin \bel(s), msg(M)\in \bel(s), remove(M) \in E}\}$ $\cup$ \\
\> $\{ \mathit{delete(\alpha)} \mid \alpha\in\bel(s)$, \\
\>\> $\alpha \in \{ \mathit{search(T), found(T), success, failure,
empty\_trash}\}\,\}$.
\end{tabbing}
}
This update policy (which does not respect possible conflicts of
$\mathit{save}$ and $\mathit{remove}$) intuitively adds all incoming information, plus
a fact $\mathit{msg}(M)$ for each incoming mail to the knowledge base. The current date is maintained
by deleting the old date. As well, all old information from a previous
event, relative to a search or to the trash, is removed. If
an event contains $\mathit{empty\_trash}$, then all  messages in the trash
are eliminated. Like in the previous example, the realization 
assignment $\rf$ may be given by $\rf_{\pm}$ from Subsection~\ref{sec:frame}, or could map 
$s$ and $A$ incrementally  to a sequence of ELPs using 
as $\bel$ simply the answer set semantics for 
sequences of
ELPs.
\end{example}

\section{Capturing Frameworks for Knowledge Evolution}
\label{sec:frameworks}

To emphasize the generality of our framework, we now discuss how
existing frameworks for updating nonmonotonic knowledge bases can be
captured in terms of evolution frames. This is possible at two
different levels:

\begin{myenumerate}
\item 
At an ``immediate update'' level, frameworks for updating logic programs
can be considered, where each event is  an {\em update program}, and the update policy is
the (implicit) way in which update programs and the current knowledge are
combined, depending on the semantics of updates of each approach. For
example, the formalisms of update programs~\cite{eite-etal-00f,eite-etal-00g}, dynamic logic
programs~\cite{alfe-etal-99a}, revision
programs~\cite{mare-trus-94,mare-trus-98}, abductive theory
updates \cite{inou-saka-99}, and updates through prioritized logic
programs (PLPs)~\cite{foo-zhan-98} fall into this class.

\item 
At a higher level, frameworks can be considered which allow for
specifying an explicit {\em update policy} in some specification
language, and which offer a greater flexibility in the handling of
updates.  Examples of such frameworks are $\UL$~\cite{eite-etal-01},
LUPS and LUPS$^\ast$ \cite{alfe-etal-99b,alfe-etal-02,leite-01},
KABUL~\cite{leite-02}, and, while not directly given in these terms,
${\cal PDL}$~\cite{lobo-etal-99}.
\end{myenumerate}

In what follows, we show how some of the above mentioned frameworks
can be expressed in evolution frames, which shows the generality of
the approach.  We start capturing the formalisms at the update level
introduced in Section~\ref{prel}, \iec the answer set semantics for update programs,
represented by $\bel_E(\cdot)$,  and the dynamic stable model
semantics for generalized logic programs, represented by $\bel_{\oplus}(\cdot)$. For
both semantics, we also show how they are combined with convenient
specification languages to form higher level frameworks:
$\bel_{\oplus}(\cdot)$ is combined with the language
LUPS~\cite{alfe-etal-99b,alfe-etal-02}, which allows for more flexibility 
of the update process, permitting to dynamically specify the contents of a
sequence of updates by means of update commands; and the semantics
$\bel_E(\cdot)$ is employed together with the language
$\UL$~\cite{eite-etal-01}, which is more expressive 
than LUPS. It allows for update statements to depend on other update
statements in the same $\UL$ policy, and more complex conditions
on both the current belief set and the actual event can be specified. Further
frameworks and semantics are also discussed here, albeit more briefly and stressing only the main characterizations.
We repeatedly use the particular set $\AC_{\mathit{ins}}$
of insert commands, the insert policy $\up_{\mathit{ins}}$, and the insert
realizations $\rf_{\mathit{ins}}$ and $\rf_{\pm}$ from Subsection~\ref{sec:frame}.

\subsection{Update Programs and $\UL$}

Update programs  \cite{eite-etal-00f,eite-etal-00g} are captured by the following evolution frame:
$$
\EF_{\upd}=\tuple{\at,\EC_\at,\AC_{\mathit{ins}},\up_{\mathit{ins}},\rf_{\mathit{ins}},\bel_E},
$$
where $\EC_\at$ is the collection of all ELPs
over $\at$, and $\bel_E(\cdot)$ is the belief operator defined in Section~\ref{prel}. The $\UL$ framework \cite{eite-etal-01,eite-etal-02c}
corresponds to the 
evolution frame
$$
\EF_{\UL}=\tuple{\at,\EC,\AC_{\UL},\up_{\UL},\rf_{\UL},\bel_E}, 
$$
where
\begin{itemize}
\item \begin{tabbing}
$\AC_{\UL}= \{$\=${\bf assert}(r),\, {\bf retract}(r),\, 
             {\bf always}(r),\, {\bf cancel}(r),\, {\bf ignore}(r),$\\
          \> ${\bf assert\_event}(r),\, {\bf
              retract\_event}(r),\, {\bf always\_event}(r) \mid r \in \lang_\at\}$,
\end{tabbing}
and the commands have the meaning as in~\cite{eite-etal-01};
\item $\up_{\UL}$ is defined by any set of update statements in the
language $\UL$, which are evaluated through a logic program as
defined in~\cite{eite-etal-01};
\item $\rf_{\UL}$ realizes the translation
$\tr(\KB;U_1,\ldots,U_n)$ from \cite{eite-etal-01}, which compiles the
initial knowledge base $\KB$ and the sets of update commands
$U_1,\ldots,U_n$, in response to the events $E_1,\ldots,E_n$ in
$s=\tuple{\KB,E_1,\ldots,E_n}$, into a sequence $(P_0,\ldots,P_n)$ of
ELPs. The resulting compilation $\compo_{\UL}(\cdot)$ is incremental.
\end{itemize}

Observe that, while $\tr(\cdot)$ as in \cite{eite-etal-01} is involved and
has to keep track of persistent update commands ${\bf
always[\_event]}(r)$ from the past, as shown in
\cite{eite-etal-02c}, it is possible, by encoding persistent
commands in polynomial time in the belief set, to restrict actions, without loss of 
expressiveness, to the commands ${\bf assert}$ and ${\bf retract}$ (whose 
meaning is the intuitive one) and making $\rf$ actually depend only the
belief set $\bel(\pi_{n-1}(s))$ of the predecessor and the event $E_n$.

\subsection{Dynamic Logic Programs, LUPS, and LUPS$^\ast$}

Dynamic logic programming \cite{alfe-etal-98,alfe-etal-99a} can be captured by the following evolution frame:
$$
\EF_{\oplus}=\tuple{\at,\EC_{\mathit{gp}},\AC_{\mathit{ins}},\up_{\mathit{ins}},\rf_{\mathit{ins}},\bel_{\oplus}},
$$
where $\EC_{\mathit{gp}}$ is the collection of all finite sets of
generalized logic program rules, \iec no strong negation is available
and weak negation can occur in the head of rules, and
$\bel_{\oplus}(\cdot)$ is the semantics of dynamic logic programs as
given in Section~\ref{prel}.

The LUPS framework \cite{alfe-etal-99b} for update specifications corresponds to the following evolution frame:
$$
\EF_{L}=\tuple{\at,\EC_{L},\AC_{L},\up_{L},\rf_{L},\bel_{\oplus}},
$$
where 
\begin{itemize}
        \item $\EC_{L}$ is the collection of all finite sets of LUPS statements (cf.~\cite{alfe-etal-99b});
        \item \begin{tabbing}
          $\AC_{L}= \{$\=${\bf assert}(r),\, {\bf retract}(r),\, {\bf
                always}(r),\, {\bf cancel}(r),$\\
          \> ${\bf assert\_event}(r),\, {\bf
              retract\_event}(r),\, {\bf always\_event}(r) \mid r \in \lang_\at\}$,
\end{tabbing}
        where the commands have the meaning explained in~\cite{alfe-etal-99b};
        \item $\up_{L}$ is defined by $\up_{L}(s,E)= \{ \mathit{cmd}(r) \in
                \AC_{L}$ $\mid$  $E$ contains $\mathit{cmd}(r)\ \mathit{when} \ \mathit{cond}$
                and $\mathit{cond} \in \bel_{\oplus}(s)\}$; 
\item $\rf_{L}$ is as described in~\cite{alfe-etal-99b}; that is, $\rf_{L}(s,A)$ adds for
$s=\tuple{\KB,E_1,\ldots,E_n}$ and $A$ a program $P_{n+1}$ to the sequence of programs
$(P_0,\ldots,P_n)$ associated with $s$, returning
$(P_0,\ldots,P_{n+1})$, where $P_{n+1}$ is computed from the persistent commands $PC_n$
valid at state $s$, $\bel(s)$, and the LUPS commands in $A$.
\end{itemize}

In~\cite{leite-01}, the semantics of LUPS has been slightly modified and 
 extended by a permanent retraction command. The resulting 
framework, LUPS$^\ast$, can be captured by the following evolution frame:
$$
\EF_{L^\ast}=\tuple{\at,\EC_{L^\ast},\AC_{L^\ast},\up_{L^\ast},\rf_{L^\ast},\bel_{\oplus}},
$$
where 
\begin{itemize}
        \item $\EC_{L^\ast}$ is the collection of all finite sets of 
LUPS$^\ast$ statements (cf.~\cite{leite-01});
        \item \begin{tabbing}
          $\AC_{L^\ast}= \{$\=${\bf assert}(r),\, {\bf retract}(r),\, 
           {\bf assert\_event}(r),\, {\bf retract\_event}(r),$\\   
          \> ${\bf always\ assert}(r),\, {\bf always\ retract}(r),\, 
              {\bf always\ assert\_event}(r),$\\
          \> ${\bf always\ retract\_event}(r),\, {\bf cancel\ assert}(r),\, 
              {\bf cancel\ retract}(r) \mid$\\
          \> $r \in \lang_\at\}$,
\end{tabbing}
        where the commands have the meaning as described 
        in~\cite{leite-01};
        \item $\up_{L^\ast}$ is defined by $\up_{L^\ast}(s,E) \!=\! \{ \mathit{cmd}(r) \in
                \AC_{L^\ast}$ $\mid$  $E$ contains $\mathit{cmd}(r)\ \mathit{when} \ \mathit{cond}$
                and $\mathit{cond} \in \bel_{\oplus}(s)\}$; 
\item $\rf_{L^\ast}$ is given as in~\cite{leite-01}; like 
before, 
$\rf_{L^\ast}(s,A)$ adds a program $P_{n+1}$ to the sequence of programs
$(P_0,\ldots,P_n)$ associated with $s$, where $P_{n+1}$ is computed from  
persistent commands $PC^\ast_n$ valid at state $s$, $\bel(s)$, and the LUPS$^\ast$ commands in $A$.
\end{itemize}

Like in the case of $\UL$, the compilation functions $\compo_{L}(\cdot)$ and 
$\compo_{L^\ast}(\cdot)$ are incremental, and also persistent commands 
(${\bf always}(r)$ and ${\bf always\_event}(r)$, as well as 
${\bf always\ assert}(r)$, ${\bf always\ assert\_event}(r)$, 
${\bf always\ retract}(r)$, and ${\bf always}$ ${\bf retract\_event}(r)$, 
respectively) can be eliminated through coding into the knowledge base.

\subsection{Revision Programs}

In \cite{mare-trus-94,mare-trus-98}, a language for revision specification of
knowledge bases is presented, which is based on logic programming
under the stable model semantics. A knowledge base is in this context
a set of atomic facts, i.e., a plain relational database. Revision
rules describe which elements are to be present (so-called
\emph{in-}rules) or absent (\emph{out-}rules) from the knowledge base,
possibly under some conditions. A fixed-point operator, which
satisfies some minimality conditions, is introduced to compute the
result of a revision program. As for stable models, there may be
several knowledge bases or no knowledge base satisfying a given revision program.

The framework of revision programs can be captured by the following
evolution frame:
$$
\EF_{\mathit{Rev}}=\tuple{\at,\EC_{\mathit{Rev}},\AC_{\mathit{Rev}},\up_{\mathit{Rev}},\rf_{\mathit{Rev}},\bel_{\mathit{Rev}}},
$$
where 

\begin{itemize}
\item $\EC_{\mathit{Rev}}$ is the collection of finite sets of {\em
revision} rules, \iec negation-free rules whose constituents are of
the form $\mathit{in}(B)$ or $\mathit{out}(B)$, where $B$ is an atom from 
$\at$;
        \item $\AC_{\mathit{Rev}} = \{ \mathit{insert}(B), \mathit{delete}(B) \mid B\in \at\}$;
       \item $\up_{\mathit{Rev}}$ is defined by
$\up_{\mathit{Rev}}(s,E)=\{ \mathit{insert}(B) \mid B \in I\} \cup \{
\mathit{delete}(B) \mid B \in O\}$, where $(I,O)$ is the
\emph{necessary change} (cf.~\cite{mare-trus-94}) for $\comp{s}$ with
respect to $E$;
       \item $\rf_{\mathit{Rev}}$ is defined by 
$\rf_{\mathit{Rev}}(s,\emptyset)=\KB$ if $s=\langle\KB\rangle$, and
$$                
        \rf_{\mathit{Rev}}(s,A)= (\comp{s} \cup \{ B \mid \mathit{insert}(B) \in A\})
\setminus \{ B \mid \mathit{delete}(B) \in A\}
$$
otherwise, \iec $\rf_{\mathit{Rev}}(s,A)$ corresponds to 
$\rf_{\pm}(s,A)$ where $P(s)= \comp{s}$. Notice that
$\rf_{\mathit{Rev}}(s,A)$, and in particular $\comp{s}$, where
$s=\tuple{\KB;E_1,\ldots,E_n}$, is thus a set of facts.
\item $\bel_{\mathit{Rev}}$ is such that, for each ELP $P$, it returns the collection of facts in $P$.

\end{itemize}

\subsection{Abductive Theory Updates}

Inoue and Sakama \cite{inou-saka-99} developed an approach to theory
update which focuses on nonmonotonic theories.  They introduced an
extended form of abduction and a framework for modeling and
characterizing nonmonotonic theory change through abduction.
Intuitively, this is achieved by extending an ordinary abductive
framework by introducing the notions of negative explanation and
\emph{anti-explanation} (which makes an observation invalid by adding
hypotheses), and then defining autoepistemic updates by means of this
framework.

The framework of extended abduction is then used in~\cite{inou-saka-99} to 
model updates of nonmonotonic theories which are represented by ELPs. 
For theory updates, the whole knowledge base is subject to change. New
information in form of an update program has to be added to
the knowledge base and, if conflicts arise, higher priority is
given to the new knowledge. The updated knowledge base is
defined as the union  $Q\cup U$ of the new information $U$ and a maximal
subset $Q\subseteq P$ of the original program that is consistent with the
new information (which is always assumed to be consistent). The
abductive framework is in this context used for specifying
priorities between current and new knowledge, by choosing as
abducibles the difference between the initial and the new
logic program.
The framework for updates by means of abduction can be captured by the following evolution frame:
$$
\EF_{\mathit{Abd}}=\tuple{\at,\EC_\at,\AC_{\mathit{Abd}},\up_{\mathit{Abd}},\rf_{\mathit{Abd}},\bel_{\mathit{Abd}}},
$$
where 

\begin{itemize}
\item $\AC_{\mathit{Abd}} = \AC_{\mathit{ins}} \cup \AC_{\mathit{del}}$, where $\AC_{\mathit{del}} = \{
\mathit{delete}(r) \mid r \in \lang_\at\}$; 

        \item $\up_{\mathit{Abd}}$ is defined by
$$
\up_{\mathit{Abd}}(s,E) = \{ \mathit{insert}(r) \mid r \in E\} \cup \{ \mathit{delete}(r) \mid r \in F \subseteq
                \comp{s}\setminus E\}, 
$$                
where $F$ is, as defined in~\cite{inou-saka-99},  a maximal set of rules
to be removed from the current knowledge base $\comp{s}$, which is a
single logic program. Note that, in general, $F$ may not be unique. Hence, for a deterministic update 
policy, we assume a suitable \emph{selection function} $\sigma$ which 
chooses one of the possible outcomes for $F$.

\item $\rf_{\mathit{Abd}}$ is defined by $\rf_{\mathit{Abd}}(s,\emptyset)=\KB$ if $s=\langle\KB\rangle$, and
$$
\rf_{\mathit{Abd}}(s,A)=(\comp{s} \setminus \{ r \mid \mathit{delete}(r) \in
                A\}) \cup \{ r \mid \mathit{insert}(r) \in A\}
$$                
otherwise, i.e., $\rf_{\mathit{Abd}}$ amounts to $\rf_{\pm}(s,A)$ for
$P(s)=\comp{s}$, provided that $A$ does not contain conflicting
commands $\mathit{delete}(r)$ and $\mathit{insert}(r)$ for any rule $r$.

\item $\bel_{\mathit{Abd}}$ is such that, for any ELP $P$, it is 
the ordinary answer set semantics of ELPs.
\end{itemize}

\subsection{Program Updates by means of PLPs}

In~\cite{foo-zhan-98}, the update of a knowledge base of ground
literals by means of a {\em prioritized logic program} (PLP) is
addressed. The idea in updating the initial program, $P$, with respect
to the new one, $Q$, is to first eliminate contradictory rules from
$P$ with respect to $Q$, and then to solve conflicts between the
remaining rules by means of a suitable PLP. The semantics of the
update is thus given by the semantics of the corresponding PLP, for
which Zhang and Foo use the one they have proposed earlier
in~\cite{zhan-foo-97b}, which extends the answer set semantics.  The method
is to reduce PLPs to ELPs by progressively deleting rules that, due to
the defined priority relation, are to be ignored. The answer sets of
the resulting ELP are the intended answer sets of the initial PLP.
Formulated initially for static priorities, the method was extended to
dynamic priorities, which are handled by a transformation into a
corresponding (static) PLP.

The framework for updates by means of PLPs is defined only for single
step updates, and a generalization to multiple steps is not
immediate. We can model a single-step update by the following
evolution frame:
$$
\EF_{\mathit{PLP}}=\tuple{\at,\EC_\at,\AC_{\mathit{ins}}\cup \AC_{\mathit{del}},\up_{\mathit{PLP}},\rf_{\mathit{PLP}},\bel_{\mathit{PLP}}},
$$
where 
\begin{itemize}
\item $\up_{\mathit{PLP}}$ is defined by $\up_{\mathit{PLP}}(s,E)=\{ \mathit{insert}(r) \mid r \in
E \} \cup \{ \mathit{delete}(r) \mid r\in R(s,E)\}$ 
where $R(s,E)$ is computed, along the
procedure in \cite{foo-zhan-98}, as a set of rules to be retracted from
the current knowledge base; 
\item $\rf_{\mathit{PLP}}$ is defined by 
$\rf_{\mathit{PLP}}(s,\emptyset)=\KB$ if $s=\langle\KB\rangle$, and
$\rf_{\mathit{PLP}}(s,A) = (P_1,P_2)$ otherwise, where $P_1 = \comp{s} \setminus \{ r
                \mid \mathit{delete}(r) \in A \}$ and $P_2=\{ r \mid \mathit{insert}(r) \in A\}$;
\item $\bel_{\mathit{PLP}}$ is the semantics for prioritized logic programs \cite{zhan-foo-97b},
viewing $(P_1,P_2)$ as a program where the rules of
$P_2$ have higher priority than the ones in $P_1$. 
\end{itemize}

\nop{**** 
Furthermore, the following formalisms can be expressed in a similar fashion:
dynamic logic programming \cite{alfe-etal-99a} (by allowing
$\naf$ in rule heads), LUPS
\cite{alfe-etal-99b}, abductive theory updates \cite{inou-saka-99},
and program updates by means of PLPs~\cite{foo-zhan-98}.
****}
Thus, several well-known approaches to
updating logic programs can be modeled by evolution frames.

\subsection{Further approaches}

We
remark that further approaches, though not concerned with logic
programs, might be similarly captured. For example, to some extent,
Brewka's declarative revision strategies \cite{brew-00} can be
captured.  Brewka introduced a nonmonotonic framework for belief
revision which allows reasoning about the reliability of information,
based on meta-knowledge expressed in the object language itself. In
this language, revision strategies can be declaratively specified as
well. The idea is to revise nonmonotonic theories by adding new
information to the current theory, and to use an appropriate
nonmonotonic inference relation to compute the accepted conclusions of
the new theory.

The desired result is achieved in two steps. The first step consists in an 
extension of default systems in order to express preference information in 
the object language, together with an appropriate new definition of theory 
extensions. In a second step, a notion of prioritized inference is introduced, 
formalized as the least fixed-point of a monotone operator, thus identifying 
epistemic states with preferential default theories under this ad hoc 
semantics. 
The approach can be captured by a suitable evolution frame
$$
\EF_{\cal T}=\tuple{\at,\EC_{\cal T},\AC_{\cal T},\up_{\cal T},\rf_{\cal T},\bel_{\cal T}},
$$
which naturally models the insertion of formulas into a
preference default theory, \iec
\begin{itemize}
        \item $\EC_{\cal T}$ is the set of all propositional formulas of the language;
        \item $\AC_{\cal T} = \{ \mathit{insert}(f) \mid f \in \EC_{\cal T}\}$;
        \item $\up_{\cal T}$ is implicitly encoded in the current
                knowledge state (\iec the current preference default
                theory, cf.~\cite{brew-00}), and is such that
                $\up_{\cal T}(s,E)= \up_{\mathit{ins}}(s,E)= \{ \mathit{insert}(f) \mid f \in E\}$;
        \item $\rf_{\cal T}$ produces the new preference default
                theory by simply executing the insertion of the new
                formula(s) into it, i.e., $\rf_{\cal
        T}(s,A)=\rf_{\mathit{ins}}(s,A) = {\cal T}(s) \cup \{ f \mid
        \mathit{insert}(f) \in A\}$, where ${\cal T}(s) = \comp{s}$ and $\rf_{\cal T}(s,\emptyset)=\KB$ if $s=\langle\KB\rangle$, as usual; and
        \item $\bel_{\cal T}$ is the function assigning to each
                preference default theory its set of accepted
                conclusions, as defined in~\cite{brew-00}.
\end{itemize}
However, there also exist frameworks which take a different point of view and 
cannot be captured by our definition of an evolution frame. This is the case 
if the state of an agent and the environment, and thus the action taken 
by an agent, are dependent on the whole history of events and actions
taken 
(which is also known as a {\em run}), not only on the current state. The approach of 
Wooldridge~\cite{wool-00b} is an example of such a framework.

\section{Reasoning About Knowledge-Base Evolution} 
\label{sec:reasoning}

We now introduce our logical language for expressing properties of
evolving knowledge bases, \EKBL\ (Evolving Knowledge Base
Logic),
which we define as a branching-time temporal logic akin to CTL
\cite{emer-90}, which has become popular for expressing temporal
behavior of concurrent processes and modules in finite state systems.

\paragraph{Syntax.} The primitive logical operators of the language \EKBL\ 
are:
\begin{itemize}
\item
the Boolean connectives $\land$ (``and'') and $\fneg$ (``not''); 

\item
the evolution quantifiers $\A$ (``for all futures'') and $\E$ (``for some future''); and

\item
the linear temporal operators  $\mathsf{X}$ (``next time'') and $\U$ (``until'').
\end{itemize}

Atomic formulas of \EKBL\ are identified with the rules in the language
$\lang_\at$, given an alphabet $\at$;
composite formulas are \emph{state formulas}, as defined---by means of atomic 
formulas and \emph{evolution formulas}---below. Note that we use the symbol 
$\fneg$ for negation in composite formulas, in order to distinguish it
from the negation symbols used in atomic formulas occurring in rules.
\begin{enumerate}
        \item Each atomic formula is a state formula. 
        \item If $\varphi$ and $\psi$ are state formulas, then $\varphi \wedge \psi$ and $\fneg \varphi$ are state formulas.
        \item If $\varphi$ is an evolution formula, then $\E\varphi$ and $\A\varphi$ are state formulas.
        \item If $\varphi,\psi$ are state formulas, then $\mathsf{X}\varphi$ and $\varphi \U\psi$ are evolution formulas.
\end{enumerate}

Intuitively, evolution formulas describe properties of the evolving
knowledge base, since they use the linear-time operators ``next time''
and ``until,'' which apply to a given infinite evolution path
consisting of knowledge states which are reached by successive events.
The operator $\mathsf{X}$ refers to the next state of the path and
states that the formula $\varphi$ is true, while the operator $\U$ refers to a (possibly empty) initial segment of the path, and
asserts that $\varphi$ holds true in each state of this segment and
that immediately after it $\psi$ holds true.

We may extend our language by defining further Boolean connectives
$\OR$ (``or''), $\IMPL$ (``implies''), and $\IFF$ (``equivalence'') in
terms of other connectives, as well as important linear-time operators
such as $\F\varphi$ (``finally $\varphi$'') and $\G\varphi$
(``globally $\varphi$''), which intuitively evaluate to true in path
$p$ if $\varphi$ is true at some resp.\ every stage $p_i$.

The following examples illustrate the use of the logical language
\EKBL\ for expressing certain properties
a of given evolution frame. 

\begin{example} Even for our rather simple shopping agent of 
Example~\ref{exa:shop-agent} some interesting 
properties can be formulated. For convenience, we allow in formulas non-ground
rules as atoms, which is a shorthand for the conjunction of all ground
instances which is assumed to be finite.  Recall that we identify
facts with literals.
\begin{itemize}
\item There can never be two current dates:
\begin{equation}  \label{exa-unique-date}
\varphi_1 \,=\, \A\G((\mathit{date}(T)\wedge \mathit{date}(T'))\supset T=T').
\end{equation}

\item If there is a shop on sale which is up, then a query is always performed:
\begin{equation}\label{exa-try-query}
\varphi_2 \,=\, \A\mathsf{G}((\mathit{up}(S)\ \wedge\ (\mathit{sale}(S)) \supset \mathit{site\_queried}).
\end{equation}
\end{itemize}
\end{example}

\begin{example}
In order to see whether the mail agent in
Example~\ref{exa:mail-agent} works properly, the first property of the 
previous example (formula (\ref{exa-unique-date})), stating that there can 
never be two different current dates, applies with slight syntactic 
modifications, \iec  
\begin{itemize}
\item[]\begin{equation}  \label{exa-unique-date-mail}
\varphi_3 \,=\, \A\G((\mathit{today}(D)\wedge \mathit{today}(D'))\supset D=D').
\end{equation}
\end{itemize}
In addition, we may consider the following properties.
\begin{itemize}
\item The type of a message cannot change: 
\begin{equation}\label{exa-type-change}
\varphi_4 \,=\, \A\G(\mathit{type}(M,T) \supset \fneg \E\F(\mathit{type}(M,T')\ \wedge \ T \neq T')).
\end{equation}

\item 
If a message is removed or saved (at least once), then the message is never 
trashed until it is either deleted or saved:
\begin{eqnarray}\label{exa-trashed} 
\varphi_5 \,=\, \A\G\big(&&(\mathit{msg}(m) \wedge \A\F(\mathit{remove}(m)\vee\mathit{save}(m))) \supset \nonumber \\
&&\A(\fneg \mathit{trash}(m)\U(\mathit{remove}(m)\vee\mathit{save}(m))\ \big).
\end{eqnarray}
\end{itemize}
\end{example}

\paragraph{Semantics.} We now define formally the semantics of formulas in our language with
respect to a given evolution frame. To this end, we introduce the
following notation. 
\begin{definition}\label{def-path}
Given an event class $\EC$, a \emph{path} is an {\rm (}infinite\/{\rm )} sequence $p=\seqinf{s_i}{i}{0}$ of
knowledge states $s_i \in \KS(\EC)$ such that 
$s_i$ is a successor of
$s_{i-1}$, for every $i>0$. By $p_i$ we denote the knowledge state at stage $i$ in $p$,
i.e., $p_i = s_i$, for every $i>0$.
\end{definition}

\begin{definition}\label{def-model-relation}
Let $\EF=\tuple{\at,\EC,\AC,\up,\rf,\bel}$ be an evolution frame,
let $s$ be a knowledge state over $\EC$, 
and let $p$ be a path.
The satisfaction relation $\EF, s \models \varphi$, resp.\ $\EF, p \models
\varphi$, where $\varphi$ is an $\EKBL$ formula, is recursively defined as follows:

\begin{enumerate}
        \item $\EF, s \models r$ iff  $r \in \bel(s)$, for any atomic
        \EKBL\ formula $r$;

        \item $\EF, s \models \varphi_1 \wedge \varphi_2$ iff $\EF, s \models  
              \varphi_1$  and $\EF, s \models \varphi_2$;

        \item $\EF, s \models \fneg \varphi$ iff $\EF, s \not \models \varphi$;

        \item $\EF, s \models \E\varphi$ iff $\EF, p' \models \varphi$, for some 
              path $p'$ starting at $s$;

        \item\label{def-model-relation-label:1}
 $\EF, s \models \A\varphi$ iff $\EF, p' \models \varphi$, for each 
              path $p'$ starting at $s$;

        \item\label{def-model-relation-label:n}
              $\EF, p \models \mathsf{X} \varphi$ iff  $\EF, p_1 \models \varphi$;
        \item  $\EF, p \models \varphi_1 \U \varphi_2$ iff
              $\EF,p_i\models \varphi_2$ for some $i\geq 0$ and
               $\EF,p_j\models\varphi_1$ for all $j<i$. 
\end{enumerate}

\end{definition}

If $\EF, s \models \varphi$ (resp., $\EF, p \models \varphi$) holds,
then knowledge state $s$ (resp., path $p$) is said to
\emph{satisfy} formula $\varphi$ \emph{in} the evolution frame $\EF$,
or $\varphi$ is \emph{true} at state $s$ (resp., path $p$) \emph{in} the evolution 
 frame $\EF$.

Notice that any evolution frame $\EF$ induces an infinite transition
graph which amounts to a standard Kripke structure
$K_{\EF}=\tuple{S,R,L}$, where $S=\mathcal{S}_\EF$ is the set of knowledge states, $R$
is the successor relation between knowledge states, and $L$ labels
each state $s$ with $\bel(S)$, such that $s$ satisfies $\varphi$ in
$\EF$ iff $K_{\EF},s\models \varphi$ (where $\models$ is defined in the usual way).

As easily seen, the operators $\F$ and $\G$ are expressed by
$\F\varphi = \top \U\varphi$ and $\G\varphi = \fneg(\top
\U\fneg\varphi)$, respectively, where $\top$ is any tautology;
thus, $\A\G\varphi = \fneg \E (\top\U \fneg\varphi)$ and $\E\G\varphi =
\fneg\A(\top\U\fneg\varphi)$.  Other common linear-time operators can be
similarly expressed, e.g., $\varphi \B\psi = \fneg ((\fneg
\varphi)\U\psi)$ \textrm{(``$\varphi$ before $\psi$''), or} $\varphi
\mathsf{V} \psi = \fneg (\fneg \varphi \U(\fneg\psi))$
\textrm{(``$\varphi$ releases $\psi$'')}.

Let us reconsider our running examples. 

\begin{example} It is easily verified that the initial
knowledge base $\KB$ of the shopping agent satisfies both formulas
(\ref{exa-unique-date}) and (\ref{exa-try-query}) in
the respective $\UL$ evolution frame $\EF_{\UL}$, \iec $\EF_{\UL},\KB
\models \varphi_1$ and $\EF_{\UL},\KB
\models \varphi_2$. 

As for $\KB$ as in Example~\ref{exa:mail-agent} for the mail agent, this set satisfies formulas
(\ref{exa-unique-date-mail}) and (\ref{exa-trashed}) in
the respective $\UL$ evolution frame $\EF_{\UL}$, \iec $\EF_{\UL},\KB
\models \varphi_3$ and $\EF_{\UL},\KB
\models \varphi_4$, while it is easily seen
that it does not satisfy formula (\ref{exa-type-change}), \iec $\EF_{\UL},\KB
\not\models \varphi_5$ . 
\end{example}

In what follows, we are mainly interested in relations of the form
$\EF, \KB \models \varphi$, \iec whether some formula $\varphi$ is
satisfied by some initial knowledge base $\KB$ with respect to some
given evolution frame $\EF$. In particular, we analyze in
Section~\ref{sec:complexity} the computational complexity
of this problem.

\section{Knowledge-State Equivalence}\label{sec:equivalence}

While syntactically different, it may happen that knowledge states $s$
and $s'$ are semantically equivalent in an evolution frame, \iec $s$ and $s'$ may have
the same set of consequences for the current and all future events. 
We now consider how such equivalences
can be exploited to filtrate a given evolution frame $\EF$ such that, under suitable
conditions, we can decide $\EF,s\models \varphi$ in a finite structure
extracted from the associated Kripke structure $K_{\EF}$. 

We start with the following notions of equivalence.

 \begin{definition}
Let $\EF=\tuple{\at,\EC,\AC,\up,\rf,\bel}$ be an evolution frame and
$k\geq 0$ some integer. Furthermore, let $s,s'$ be knowledge states over $\EC$. Then, 
\begin{enumerate}
\item  $s$ and $s'$ are \emph{$k$-equivalent in $\EF$}, denoted $s\equiv^k_{\EF} s'$, if
$\bel(s+E_1\commadots E_{k'})=\bel(s'+E_1\commadots E_{k'})$, for all
events $E_1\commadots E_{k'}$ from $\EC$, where $k'\in \{0,\ldots,k\}$;

\item $s$ and $s'$ are  \emph{strongly equivalent in $\EF$},
denoted $s\equiv_{\EF} s'$, iff $s\equiv^k_{\EF} s'$ for every $k\geq
0$.
\end{enumerate}

\end{definition}

We call 0-equivalent states also \emph{weakly equivalent}. The
following result is obtained easily.

\begin{theorem}\label{thm:strong-equivalence}
Let $\EF=\tuple{\at,\EC,\AC,\up,\rf,\bel}$ be an evolution frame and $s,s'$ 
knowledge states over $\EC$. Then,

\begin{enumerate}
\item\label{thm:seq-1} $s \equiv_{\EF} s'$ implies that $\EF, s \models \varphi$ is equivalent to $\EF, s' \models \varphi$, for any
formula $\varphi$;

\item\label{thm:seq-2} $s \equiv^k_{\EF}s'$ implies that $\EF, s
\models \varphi$ is equivalent to $\EF, s' \models \varphi$, for any
state formula $\varphi$ in which $\U$ does not occur and the nesting
depth with respect to $\E$ and $\A$ is at most~$k$.
\end{enumerate}
\end{theorem}

\begin{proof} We prove Part~\ref{thm:seq-1} of the theorem by 
induction on the formula structure of the state formula $\varphi$.

\medskip
\noindent
{\sc Induction Base.} Let $\varphi$ be an atomic formula and assume 
$s \equiv_{\EF} s'$, for knowledge states $s, s'$. Obviously, it holds that 
$\bel(s) = \bel(s^\prime)$. Thus, it follows that 
$\EF, s \models \varphi$ iff $\bel(s^\prime) \models \varphi$ and, hence, 
$\EF, s \models \varphi$ iff $\EF, s^\prime \models \varphi$.

\medskip
\noindent
{\sc Induction Step.} Assume that Part~\ref{thm:seq-1} of
Theorem~\ref{thm:strong-equivalence} holds for formulas 
$\psi$ of depth at most $n-1$, \iec $s \equiv_{\EF} s'$ implies 
$\EF, s \models \psi$ iff $\EF, s^\prime \models \psi$. Let $\varphi$
be a formula of depth $n$, and consider the following cases. 

\begin{itemize}
\item $\varphi = \psi_1 \wedge \psi_2$ or $\varphi = \fneg
\psi_1$. \quad  Then, $\psi_1$ and 
$\psi_2$ are of depth $n-1$ and, by the induction hypothesis, it holds that $\EF, s \models \varphi$ iff 
$\EF,s^\prime \models \psi_1$ and $\EF, s^\prime \models \psi_2$, respectively 
$\EF, s \models \varphi$ iff $\EF, s^\prime \not\models \psi_1$. Thus, again 
$\EF, s \models \varphi$ iff $\EF, s^\prime \models \varphi$ follows.

\item  $\varphi = \E\psi$ or $\varphi = \A\psi$.\quad  Then, $\psi$ is
an evolution formula of depth $n-1$ of the form  $\X \psi_1$ or
$\psi_1 \U \psi_2$, where $\psi_1$ and $\psi_2$ have depth
$n-2$. Since $s \equiv_{\EF} s'$, we have that $\EF, p \models \psi$ for a path 
$p=(s_i)_{i\geq0}$ such
that $s_0=s$ iff $\EF, p^\prime \models 
\psi$, where $p^\prime=(s_i^\prime)_{i\geq0}$ results from $p$ by
replacing $s_i$ with any $s^\prime_i$ such that $s_i \equiv_{\EF} s^\prime_i$,
for all $i\geq 0$. Since $s_0\equiv_{\EF} s^\prime_0$,  for each $i \geq 0$, such an $s^\prime_i$
exists. Thus, the induction hypothesis implies that $\EF, p \models \psi$ iff 
$\EF, p^\prime \models \psi$.

Hence, $\EF, s \models \E\psi$ iff $\EF, s^\prime \models \E\psi$
follows. By symmetry of $\equiv_{\EF}$, we similarly conclude that $\EF, s \models \A\psi$ iff $\EF, s^\prime \models \A\psi$.
\end{itemize}

Thus, for every state formula of depth $n$, the statement in Part~\ref{thm:seq-1}
of Theorem~\ref{thm:strong-equivalence} holds. This concludes the
induction  and proves Part~\ref{thm:seq-1} of our result.

Concerning Part~\ref{thm:seq-2} of the theorem, observe that in order 
to prove a formula $\varphi$ in which $\U$ does not occur and the
evolution quantifier depth 
is at most $k\geq 0$, initial path segments of length at most $k+1$
need to be considered. This follows from the fact that evolution
subformulas of $\varphi$ can only  be of form $\X\psi$. Moreover, since
every evolution formula must be preceded by a quantifier $\E$ or
$\A$, at most $k$ nested evolution formulas can occur in $\varphi$ and
every evolution formula of the above form, \iec $\X\psi$, can be
verified by considering the truth value of $\psi$ in successor states
of the current state. Hence, initial path segments of length at most
$k+1$ do suffice. Since for two knowledge states $s$ and $s'$ such that
$s \equiv^k_{\EF}s'$, all knowledge states reachable in $k$ steps are
equivalent, $\EF, s \models \varphi$ iff $\EF, s^\prime \models
\varphi$ holds by the same inductive argument as in the proof of
Part~\ref{thm:seq-1} above.  Thus, Part~\ref{thm:seq-2} of
Theorem~\ref{thm:strong-equivalence} follows.  \hfill$\Box$
\end{proof}

\nop{*** old, no longer needed ***
\begin{theorem}\label{thm:k-equiv}
Let $\EF$ be a regular evolution frame.  If knowledge states $s$ and
$s'$ are $k$-equivalent with respect to $\EF$, for some $k\geq 0$,
then $s$ and $s'$ are also $k'$-equivalent with respect to $\EF$, for
any $k'\leq k$.
\end{theorem}

\begin{corollary}
Under the circumstances of Theorem~\ref{thm:k-equiv}, if $s$ and $s'$ are strongly equivalent with respect to $\EF$, then they are also weakly equivalent.
\end{corollary}

The converse of Theorem~\ref{thm:k-equiv} does not hold. 
For a counterexample, consider the two knowledge states $s=\tuple{\emptyset;\{a\la\}}$ and $s'=\tuple{\{a\la\};\emptyset}$, and the following update policy: if $a\la$ is in the initial knowledge base, then add $\neg a\la$, otherwise add noting. Using, \egc the semantics described in \cite{xxx}, for which the belief operator is regular, we have that $s$ and $s'$ are 1-equivalent but not 2-equivalent.
***} 

By Part~1 of Theorem~\ref{thm:strong-equivalence}, strong
equivalence can be used to filtrate an evolution frame $\EF$ in the following 
way. 
For an equivalence relation $E$ over some set $X$, and any $x\in X$,
let $[x]_E=\{y\mid x\,E\,y \}$ be the equivalence class of $x$ 
and let $X/E=\{[x]_E\mid x\in X\}$ be the set of all equivalence
classes. Furthermore, $E$ is said to have a \emph{finite index {\rm (}with 
respect to $X$}), if $X/E$ is finite. 

Then, any equivalence relation $E$ over some set
$S\subseteq\mathcal{S}_\EF$ of knowledge states of $\EF$ compatible
with $\equiv_\EF$ (i.e., such that $s\,E\,s'$ implies $s\equiv_\EF
s'$, for all $s,s'\in S$) induces a Kripke structure
$K_{\EF}^{E,S}=\tuple{S/E,R_E,L_E}$, where $[s]_E\,R_E\,[s']_E$ iff
$s\,R\,s'$ and $L_E([s]_E)=L(s)$, which is bisimilar to the Kripke
structure $K_{\EF}$ restricted to the knowledge states in $S$.  Thus,
for every knowledge state $s$ and formula $\varphi$, it holds that
$\EF,s\models \varphi$ iff $K_{\EF}^{E,S},[s]_E\models \varphi$, for
any $S\subseteq\mathcal{S}_\EF$ such that $S$ contains all descendants
of $s$.

In the following, we consider two cases in which $S/E$ has finite
index. Prior to this, we introduce some convenient terminology and notation. 

For any state $s \in \mathcal{S}_\EF$, we denote by $\dsc(s)$ the set
of knowledge states containing $s$ and all its descendants (with
respect to $\EC$ in $\EF$, which will be clear from the context), and
by $\tree(s)$ the ordered tree with root $s$ where the children of
each node $s'$ are its successor states according to $\EF$, and $s'$
is labeled with $\bel(s')$. Furthermore, for any
$S\subseteq\mathcal{S}_\EF$, we define $\dsc(S)=\bigcup_{s\in
S}\dsc(s)$, and call $S$ {\em successor closed}, if $S = \dsc(S)$,
i.e., each successor of a knowledge state in $S$ belongs to $S$.  Note
that for any $s \in \mathcal{S}_\EF$, $\tree(s)$ has node set
$\dsc(s)$, which is successor closed.

\subsection{Local belief operators} \label{sec:loc-bel}

In the first case, we consider $\equiv_{\EF}$ itself as a relation
compatible with strong equivalence. We obtain a finite index if,
intuitively, the belief set $\bel(s)$ associated with $s$ evolves
differently only in a bounded context.  This is made precise in the following result.

\begin{theorem}\label{thm:finite-index}
Let $\EF=\tuple{\at,\EC,\AC,\up,\rf,\bel}$ be an evolution frame such
that $\EC$ is finite, and let $S\subseteq\mathcal{S}_\EF$ be a 
successor-closed set of knowledge states over $\EC$. Then, the
following two conditions are equivalent:
\begin{enumerate}
\item \label{thm:f-ind-1} $\equiv_{\EF}$ has a finite index with respect to $S$.
\item \label{thm:f-ind-2} $\equiv_{\EF}^0$ has a finite index with respect to $S$ and there is some $k\geq 0$ such that $s \equiv_{\EF}^k s'$ implies $s
\equiv_{\EF} s'$, for all $s,s'\in S$. 
\end{enumerate}
\end{theorem}

\begin{proof} 
\hspace{1ex}

\noindent 
(2 $\Rightarrow$ 1). Consider, for any $s\in S$, the
tree $\tree(s)$. At depth $i\geq 0$, there 
are $|\EC|^i$ different nodes, and thus up to depth $k$ in total $\sum_{i=0}^k|\EC|^i = $
$\frac{|\EC|^{k+1}-1}{|\EC|-1} < 2|\EC|^k$ many different nodes if $|\EC|
>1$, and $k+1$ many if $|\EC|=1$. Thus, if $d = |S/\equiv_{\EF}^0|$ is the number of different equivalence classes of the
relation $\equiv_{\EF}^0$ with respect to $S$, then there are less than $c=d^{\max(\,2|\EC|^k,k+1)}$ many trees
$\tree(s)$, where $s \in S$, which are different up to depth $k$.
Thus, there are at most $c$ knowledge
states
$s_1,\ldots,s_c$, $s_i \in S$, $1\leq i\leq c$, which are pairwise not 
strongly equivalent. Consequently, $\equiv_\EF$ has at most $c$ different 
equivalence classes with respect to $S$, and thus $\equiv_\EF$ has a finite index with respect to~$S$. 
\medskip
\noindent
(1 $\Rightarrow$ 2). Suppose the relation $\equiv_\EF$ has at most $n$ 
different equivalence classes \wrt $S$. Then, there are at most $n$ knowledge 
states $s_1,\ldots,s_n \in S$  which are pairwise not strongly
equivalent, i.e., $s_i\not\equiv_{\EF} s_j$ for all $1\leq i < j\leq n$. 
Since strongly equivalent states are also weakly equivalent, $n$ is
thus also a finite 
upper bound for the equivalence classes of the relation $\equiv_{\EF}^0$ \wrt 
$S$.

Now, for $i,j\in \{1,\ldots,n\}$ such that $i\neq j$,  let $l=l_{i,j}$  be
the smallest integer $l$ for $s_i$ and $s_j$, $1\leq
i<j\leq n$, such that $\bel(s_i+E_1\commadots
E_l)\neq\bel(s_j+E_1\commadots E_l)$, but $\bel(s_i+E'_1\commadots
E'_m)=\bel(s_j+E'_1\commadots$ $E'_m)$, for all sequences of events
$E'_1\commadots E'_m$, $0\leq m<l$. Furthermore, let
$k=\max_{i,j}(l_{i,j})$ be the largest such $l$ over all $s_i$ and
$s_j$. Note that $k$ is well defined and finite because of the finite
index of $\equiv_\EF$ \wrt S. It follows that if any two knowledge
states $s, s' \in S$ are $k$-equivalent, then they are also strongly
equivalent. Indeed, suppose the contrary, \iec suppose $s \equiv_{\EF}^k s'$,
but $s \not\equiv_{\EF} s'$. Then there exists a sequence of $l$
events, $l>k$, such that $\bel(s+E_1\commadots
E_l)\neq\bel(s'+E_1\commadots E_l)$, but $\bel(s+E'_1\commadots
E'_m)=\bel(s'+E'_1\commadots E'_m)$ holds for all sequences of events
$E'_1\commadots E'_m$, $0\leq m\leq k<l$. From the assumption that $s
\not \equiv_{\EF} s'$, it follows that $s \equiv_\EF s_i$ and $s'
\equiv_\EF s_j$, for some $i,j\in \{1,\ldots,n\}$ such that $i\neq
j$. This implies that $l_{i,j} > k$,
which contradicts the maximality of
$k$.  Thus, $s \equiv_{\EF}^k s'$ implies $s \equiv_{\EF} s'$, for all
$s,s'\in S$.  \hfill$\Box$
\end{proof}

The condition that $\equiv^0_\EF$ has a finite index, i.e., that only
finitely many knowledge states $s$ have different belief sets, is
satisfied by common belief operators if, e.g., every knowledge state
$s$ is compiled to a sequence $\compo_{\EF}(s)$ of ELPs or a single
ELP over a finite set of function-free atoms (in particular, if $\at$
is a finite propositional alphabet).

We remark that, as can be seen from the proof of
Theorem~\ref{thm:finite-index}, Condition 1 implies Condition 2 also
for arbitrary $S$, while the converse does not hold  in general for an $S$ which is
not successor closed.

By taking natural properties of $\bel(\cdot)$ and $\compo_\EF(\cdot)$
into account, we can derive an alternative version of
Theorem~\ref{thm:finite-index}. To this end, we introduce the
following concepts.

\begin{definition}
Given a belief operator $\bel(\cdot)$, we call update programs $\useq{P}$
and $\useq{P}^\prime$ \emph{$k$-equivalent}, if $\bel(\useq{P}+(Q_1\commadots
Q_k))=\bel(\useq{P}^\prime+(Q_1\commadots Q_k))$, for all programs $Q_1,\ldots,Q_i$
$(0\leq i\leq k)$. Likewise, $\useq{P}$ and $\useq{P}^\prime$ are 
\emph{strongly equivalent}, if they are $k$-equivalent for all $k\geq
0$. We say that $\bel(\cdot)$ is \emph{$k$-local}, if 
$k$-equivalence of $\useq{P}$ and $\useq{P}^\prime$
implies strong equivalence of $\useq{P}$ and $\useq{P}^\prime$, for any
update programs $\useq{P}$ and $\useq{P}^\prime$. Furthermore, $\bel(\cdot)$
is \emph{local}, if  $\bel(\cdot)$ is $k$-local for some $k\geq 0$.
\end{definition}

We obtain the following result. 

\begin{theorem}\label{thm:finite-index-sequences}
Let $\EF=\tuple{\at,\EC,\AC,\up,\rf,\bel}$ be an evolution frame such
that $\EC$ is finite and $\equiv_{\EF}^0$ has a finite index with
respect to some successor closed $S\subseteq\mathcal{S}_\EF$.  If
$\bel(\cdot)$ is local and $\compo_\EF(\cdot)$ is incremental, then
$\equiv_\EF$ has a finite index with respect to $S$.
\end{theorem}

\begin{proof}
Similar to the proof of Theorem~\ref{thm:finite-index}, consider, for any 
knowledge state $s\in S$, the tree $\tree(s)$. Each node in
$s'$ has label
$\bel(s')=\bel(\useq{P}+(Q_1, \ldots, Q_n))$, where $\useq{P} = \compo_\EF(s)$ and 
$Q_i$, $i\geq 1$, are the increments of $\compo_\EF(\cdot)$ corresponding to 
the successive events $E_i$ in $s'=s+E_1,\ldots,E_n$. Note that
incrementality of $\compo_{\EF}(\cdot)$ guarantees that the length of 
$\compo_{\EF}(s')$ is at most $n$ plus the length of $\compo_{\EF}(s)$.
Since $\bel(\cdot)$ is $k$-local,  up to depth 
$k$, there are at most $c=d^{\max(2|\EC|^k,k+1)}$ many different
trees, where $d = |S/\equiv_{\EF}^0|$.
Thus, there are at most $c$ update programs $\useq{P}_1\commadots\useq{P}_c$, 
and, hence, knowledge states $s_1,\ldots,s_c$, which are pairwise not strongly
equivalent. Consequently, $\equiv_\EF$ has at most $c$ different equivalence 
classes, from which the result follows. 
\hfill$\Box$
\end{proof}

As an application of this result, we show that certain $\UL$ evolution frames
have a finite index. To this end, we use the following lemmata. 

We say that a semantics $\bel(\cdot)$ for sequences of propositional
ELPs satisfies {\em strong noninterference}, if it satisfies, for every
propositional update sequence $\useq{P}$, the following condition: If, for every
ELP $P_1,P_2$, and $Q$ such that $Q \subseteq P_2$ and no pair of
rules $r,r'$ exists with $\head{r}=\neg \head{r'}$, where $r\in Q$ and $r' \in (P_2\setminus Q) \cup P_1$, then
$\bel(\useq{P},P_1,P_2) = \bel(\useq{P},P_1\cup Q, P_2 \setminus Q)$, i.e., the rules from $Q$ can be moved from the last component
to the penultimate one.

Recall that $\bel_E(\cdot)$ is the belief operator of the answer set
semantics of update programs \cite{eite-etal-00g}, as described in
Section~\ref{prel}.

\begin{lemma}\label{lemm:nonInt}
$\bel_E(\cdot)$ satisfies \emph{strong noninterference}.
\end{lemma}

\begin{proof}
The proof appeals to the rejection mechanism of the semantics.  Let
$P_1,P_2$, and $Q$ be ELPs, such that the following condition ($\ast$)
holds: $Q \subseteq P_2$ and no pair of rules $r,r'$ exists with
$\head{r}=\neg \head{r'}$, where $r\in Q$
and $r' \in (P_2\setminus Q) \cup P_1$.
 
If $Q = \emptyset$, the lemma holds trivially. So let $r \in Q$, but no
rule $r' \in Q$ exists such that $\head{r}=\neg \head{r'}$.  Then, there must
not be such a rule $r'$ in $P_1$ or $P_2$, otherwise condition 
($\ast$) is not fulfilled. Hence, no rule of $P_1$ is rejected by
$r$. Moreover, adding $r$ to $P_1$ can neither cause an inconsistency
of $P_1$, nor can $r$ be rejected by a rule from $P_2\setminus
Q$. Thus, $\bel_E(\useq{P}+(P_1,P_2)) = \bel_E(\useq{P}+(P_1\cup Q, P_2
\setminus Q))$ holds in this case.

Now let $Q$ also contain some rule $r'$ such that $\head{r}=\neg \head{r'}$ 
($P_1$ cannot contain such rules without violating the condition 
($\ast$)). Then, $Q$ must contain all rules with heads $\head{r}$ and 
$\neg \head{r}$ of $P_2$, and no such rule may exist in $P_1$, in order to 
fulfill ($\ast$). Again, no rule of $P_1$ can be rejected 
by any rule of $Q$, and no rule of $Q$ can be rejected by any rule from 
$P_2\setminus Q$. Additionally, adding $Q$ to $P_1$ makes $P_1$ inconsistent 
iff $P_2$ is inconsistent. As a consequence, also in this case,
$\bel_E(\useq{P}+(P_1,P_2)) = \bel_E(\useq{P}+(P_1\cup Q, P_2 \setminus Q))$. 
Since there are no other possibilities left, the lemma is shown.
\hfill$\Box$
\end{proof}

In our next result, we require Part~\ref{lemm:belE-1} of the following 
lemma, which in turn will be relevant in Section~\ref{sec:comp-equi}.

\begin{lemma}\label{lemm:belE}
Let $\useq{P}$ and $\useq{Q}$ be sequences of ELPs. Then,  
\begin{enumerate}
\item \label{lemm:belE-1} $\bel_E(\useq{P})=\bel_E(\useq{Q})$ if $\ASup(\useq{P})=\ASup(\useq{Q})$, and
\item \label{lemm:belE-2} given that $\useq{P}$ and $\useq{Q}$ are propositional 
sequences over possibly infinite alphabets, $\ASup(\useq{P})=\ASup(\useq{Q})$ if $\bel_E(\useq{P})=\bel_E(\useq{Q})$.
\end{enumerate}
\end{lemma}

\begin{proof} As for Part~\ref{lemm:belE-1},
if $\ASup(\useq{P}) = \ASup(\useq{Q})$ then 
$\bel_{E}(\useq{P}) = \bel_{E}(\useq{Q})$ is 
immediate from the definition of $\bel_E(\cdot)$.

\medskip\noindent
To show Part~\ref{lemm:belE-2}, it suffices to prove that, 
given ELPs $P_1$ and $P_2$ over a set $\at$ of atoms,
$$
\AS(P_1) = \AS(P_2)\ \mbox{ if } \ \bel_{E}(P_1) = \bel_{E}(P_2).
$$
\noindent
Suppose $\AS(P_1)\neq \AS(P_2)$, and assume first that $\at$ is finite.
Without loss of generality, suppose that $S=\{ L_1,\ldots,L_k\} \in
\AS(P_1)$ but $S \notin \AS(P_2)$. This means that the constraint 
$$
c:\quad \la L_1,\ldots,L_k,\naf L_{k+1},\ldots,\naf L_m \ ,
$$
where $L_{k+1},\ldots,L_m$ are all the atoms from $\at$ missing in $S$, is 
in $\bel_E(P_2)$ but not 
in $\bel_E(P_1)$. However, this 
contradicts the hypothesis that $\bel_E(P_1)=\bel_E(P_2)$.

This proves the result for finite $\at$. For infinite $\at$, it is 
possible to focus
on the finite set of atoms occurring in 
$\useq{P} \cup \useq{Q}$, since, as well-known for the answer set semantics,  $A, \neg A \notin S$ for each $A \in
\at\setminus\at'$ and $S \in \AS(P)$ if $P$ is an ELP on $\at'\subseteq \at$. 
\hfill$\Box$
\end{proof}

Now we can show the following result.

\begin{theorem}\label{theo:belE}
$\bel_E(\cdot)$ is local. In particular, $1$-equivalence
of update programs $\useq{P}$ and $\useq{P}^\prime$ implies
$k$-equivalence of $\useq{P}$ and $\useq{P}^\prime$, for all $k\geq 1$.
\end{theorem}

\begin{proof}
We show the result for propositional update sequences $\useq{P}$ and 
$\useq{P}^\prime$ by induction on $k\geq 1$. Since the evaluation of 
$\bel_E(\cdot)$ for non-ground update sequences amounts to the evaluation 
of propositional sequences, the result for the non-ground case follows easily. 

\medskip
\noindent
{\sc Induction Base.} The base case $k=1$ is trivial.

\medskip
\noindent
{\sc Induction Step.} Assume that $1$-equivalence of $\useq{P}$ and
$\useq{P}^\prime$ implies that they are $(k-1)$-equivalent, $k>1$.
Suppose further, that $\useq{P}$ 
and $\useq{P}^\prime$
are $1$-equivalent, but not $k$-equivalent. Then, there exist programs
$Q_1\commadots Q_{k'}$, where $k'\in \{2,\ldots, k\}$, such that
$\bel_E(\useq{P}+Q_1\commadots Q_{k'}) \neq
\bel_E(\useq{P}^\prime+Q_1\commadots Q_{k'})$, \iec according to
Part~\ref{lemm:belE-1} of Lemma~\ref{lemm:belE}, there exists a
(consistent) answer set $S \in \ASup(\useq{P}+Q_1\commadots Q_{k'})$,
such that $S \notin \ASup(\useq{P}^\prime+Q_1\commadots Q_{k'})$.  We
can remove every rule $r$ from $Q_{{k'}-1}$ and $Q_{k'}$ such that
either $S \not\models \body{r}$, or $r$ is a member of
\begin{eqnarray*}
\lefteqn{
\rs_{{k'}-1}(S,\useq{P}+Q_1\commadots
Q_{k'})\cup\rs_{k'}(S,\useq{P}+Q_1\commadots Q_{k'})=} ~~~~~~~~\\
 &  &
\rs_{{k'}-1}(S,\useq{P}+Q_1\commadots Q_{k'}) = \rs_{{k'}-1}(S,\useq{P}^\prime+Q_1\commadots Q_{k'}).
\end{eqnarray*}
  Let the resulting
programs be denoted by $Q_{{k'}-1}^\prime$ and $Q_{k'}^\prime$,
respectively. Note that still $S \in \ASup(\useq{P}+Q_1\commadots
Q_{{k'}-2},Q_{{k'}-1}^\prime,Q_{k'}^\prime)$ and $S \notin
\ASup(\useq{P}^\prime+Q_1 \commadots
Q_{{k'}-2},Q_{{k'}-1}^\prime,Q_{k'}^\prime)$ must hold, since these
rules can neither be generating for $S$, \iec fire \wrt $S$, nor
reject other rules. Observe also that $Q_{{k'}-1}^\prime \cup\,
Q_{k'}^\prime$ cannot contain a pair of rules with conflicting
heads. Otherwise, contradicting our assumption, $S$ would be
inconsistent since both rules were generating for $S$.

Now we construct the program 
$Q_{{k'}-1}^\ast=Q_{{k'}-1}^\prime\cup\, Q_{k'}^\prime$. 
 From the strong noninterference property (Lemma~\ref{lemm:nonInt}), it follows 
that $\bel_E(\useq{P}+Q_1\commadots Q_{{k'}-2},$
$Q_{{k'}-1}^\ast,\emptyset) \neq
\bel_E(\useq{P}^\prime+Q_1\commadots Q_{{k'}-2},Q_{{k'}-1}^\ast,\emptyset)$. Since, for 
every update sequence $\useq{Q}$, 
$\bel_E(\useq{Q}+\emptyset)=\bel_E(\useq{Q})$,  it follows  
$\bel_E(\useq{P}+Q_1\commadots Q_{{k'}-2},$
$Q_{{k'}-1}^\ast) \neq
\bel_E(\useq{P}^\prime+Q_1\commadots$ $Q_{{k'}-2},$ $Q_{{k'}-1}^\ast)$.
This means that  $\useq{P}$ and 
$\useq{P}^\prime$ are not $k'-1$-equivalent; however, this
contradicts the induction hypothesis that $\useq{P}$ and 
$\useq{P}^\prime$ are $k-1$-equivalent. Hence,  $\useq{P}$ and 
$\useq{P}^\prime$ are $k$-equivalent.
\hfill$\Box$
\end{proof}

Furthermore, in any $\UL$ evolution frame
$\EF=\tuple{\at,\EC,\AC_{\UL},\up_{\UL},\rf_{\UL},\bel_E}$, the update
policy $\up_{\UL}$ is, informally, given by a logic program such that
$\up_{\UL}$ returns a set of update actions from a finite set $A_0$ of
update actions, which are compiled to rules from a finite set $R_0$ of
rules, provided $\EC$ is finite. Consequently, $\equiv^0_\EF$ has
finite index with respect to any set $S$ of knowledge states $s$
which coincide on $\pi_0(s)$, i.e.\ the initial knowledge base $\KB$. 
Furthermore, $\compo_{\UL}(\cdot)$ is incremental. Thus, from the
proof of Theorem~\ref{thm:finite-index-sequences}, we obtain: 

\begin{corollary}
Let $\EF=\tuple{\at,\EC,\AC_{\UL},\up_{\UL},\rf_{\UL},\bel_E}$ be an
$\UL$ evolution frame such that $\EC$ is finite, and let
$S\subseteq\mathcal{S}_\EF$ be a successor-closed set of knowledge states such that $\{
\pi_0(s) \mid s \in S\}$ is finite. Then, $\equiv_{\EF}$ has a finite
index with respect to $S$. Moreover, $|S/\equiv_{\EF}\!\!| \leq
d^{2|\EC|}$, where $d = |S/\equiv^0_{\EF}\!\!|$.
\end{corollary}

In~\cite{leite-02}, an analogous result has been shown for 
$\bel_{\oplus}(\cdot)$, \iec $1$-equivalence of dynamic update programs 
$\useq{P}$ and $\useq{P}^\prime$ implies their strong equivalence, and, thus, 
$\bel_{\oplus}(\cdot)$ is local. Since for update policies over the LUPS or
LUPS$^\ast$ language and their respective compilations, 
the same as for their $\UL$ counterparts holds, we also get the following result:

\begin{corollary}
Let $\EF$ be a LUPS evolution frame $\tuple{\at,\EC,\AC_{L},\up_{L},\rf_{L},\bel_{\oplus}}$ or a LUPS$^\ast$ evolution frame  $\EF=\tuple{\at,\EC,\AC_{L^\ast},\up_{L^\ast},\rf_{L^\ast},\bel_{\oplus}}$
such that $\EC$ is finite, and let
$S\subseteq\mathcal{S}_\EF$ be a successor-closed set of knowledge
states such that $\{ \pi_0(s) \mid s \in S\}$ is finite. Then,
$\equiv_{\EF}$ has a finite index with respect to $S$. Moreover,
$|S/\equiv_{\EF}\!\!| \leq d^{2|\EC|}$, where $d =
|S/\equiv^0_{\EF}\!\!|$.
\end{corollary}

\subsection{Contracting belief operators}\label{sec:canonical}

Next, we discuss a refinement of strong equivalence, called \emph{canonical equivalence}, which 
also yields a
finite
index, provided that the evolution frame possesses, in some sense, only a ``bounded history''.
In contradistinction to the previous case, canonical equivalence uses
semantical properties which allow for a syntactic simplification of
update programs. We need the following notions.

\begin{definition}\label{def:contraction}
\label{property-remove} Let $\bel(\cdot)$ be a belief operator. Then, 
$\bel(\cdot)$ is called \emph{contracting} iff the following conditions hold: {\rm (}i\/{\rm )} $\bel(\useq{P}+\emptyset+\useq{P}^\prime)=\bel(\useq{P}+\useq{P}^\prime)$, for all
update programs $\useq{P}$ and $\useq{P}^\prime$; and {\rm (}ii\/{\rm )}
$\bel(\useq{P})=\bel(P_0, \ldots, P_{i-1}, P_i \setminus \{
r\} , P_{i+1}, \ldots, P_n)$, for any 
sequence $\useq{P}=(P_0,\ldots,P_n)$ and any rule $r\in P_i\cap P_j$ such that $i<j$.
An evolution frame
$\EF=\tuple{\at,\EC,\AC,\up,\rf,\bel}$ is 
contracting iff $\bel(\cdot)$ is contracting.
\end{definition}

Examples of contracting belief operators are $\bel_E(\cdot)$ and the
operator  $\bel_{\oplus}(\cdot)$ (see Section~\ref{prel}). 

By repeatedly removing duplicate rules 
and empty programs 
from any sequence $\useq{P}=(P_0,\ldots,P_n)$ of ELPs, we eventually obtain a
non-reducible sequence $\useq{P}^*=(P_0^*,\ldots,P_m^*)$, which
is called the \emph{canonical form} of $\useq{P}$. Observe that $m\leq
n$ always holds, and that $\useq{P}^*$ is uniquely determined, i.e., the reduction
process is Church-Rosser. We get the following property:

\begin{theorem}\label{thm:regular-bel}
For any contracting belief operator $\bel(\cdot)$ and any update sequence $\useq{P}$, we have that $\useq{P}$ and $\useq{P}^*$ are strongly equivalent.
\end{theorem}

\begin{proof} 
We must show that $\useq{P}$ and $\useq{P}^*$ are $k$-equivalent, for every 
$k\geq 0$. The proof is by induction on $k\geq 0$.

\medskip
\noindent
{\sc Induction Base.} We show that $\useq{P}$ and $\useq{P}^\ast$ are $0$-equivalent. The proof is by induction on the reduction process, \iec on the 
number of required removals of rules or empty programs from $\useq{P}$ in 
order to obtain $\useq{P}^\ast$. For the induction base, suppose 
$\useq{P}=\useq{P}^\ast$.
Then, $\useq{P}$ and $\useq{P}^\ast$ are trivially $0$-equivalent.
For the induction step, assume that 
$\bel(\useq{Q}) = \bel(\useq{Q}^\ast)$, for all sequences of programs 
$\useq{Q}$ such that the canonical form $\useq{Q}^\ast$ can be constructed 
using $n-1$ removals of rules and empty programs. Let $\useq{P}$ be a sequence 
of programs such that the construction of $\useq{P}^\ast$ requires $n$ 
removing steps, and let $\useq{P}^\prime$ denote any sequence of programs 
obtained from $\useq{P}$ after $n-1$ removals. Then, 
$\bel(\useq{P}) = \bel(\useq{P}^\prime)$, by induction hypothesis. 
Furthermore, $\bel(\useq{P}^\prime) = \bel(\useq{P}^\ast)$ follows trivially 
from $\bel$ being contracting. Thus, $\bel(\useq{P}) = \bel(\useq{P}^\ast)$. We have shown that for any sequence $\useq{P}$ of programs,  if $\bel(\cdot)$ is contracting, then $\useq{P}$ and 
$\useq{P}^\ast$ are $0$-equivalent.

\medskip
\noindent
{\sc Induction Step.} 
Suppose $k>0$, and let $\useq{Q}=(\useq{P}+Q_1\commadots Q_k)$ and
$\useq{R}=(\useq{P}^*+Q_1\commadots Q_k)$. Furthermore, let
$\useq{Q}^\ast$ and $\useq{R}^\ast$ denote the canonical forms of $\useq{Q}$ and $\useq{R}$,
respectively.  We show that $\useq{Q}^\ast=\useq{R}^\ast$.

 Suppose
$\useq{P}^\ast$ is obtained from $\useq{P}$ using $n$ reduction steps
and $\useq{R}^\ast$ is obtained reducing $\useq{R}$ in $m$ steps. We
construct $\useq{Q}^\ast$ as follows.  We first perform $n$ reduction
steps on the subsequence $\useq{P}$ of $\useq{Q}$, resulting in the
sequence $\useq{R}$. Then we apply $m$ reduction steps on
$\useq{R}$. Since the reduction process is Church-Rosser, no further
reductions can be applied, which proves $\useq{Q}^\ast=\useq{R}^\ast$.
From the induction base, it follows that $\useq{Q}$ and
$\useq{Q}^\ast$ are weakly equivalent, which proves $k$-equivalence of
$\useq{P}$ and $\useq{P}^\ast$.
\hfill$\Box$
\end{proof}

\subsection{Canonical evolution frames}

In this section, we study the relationship between an evolution frame and its 
\emph{canonized form}:

\begin{definition}
Given an evolution frame $\EF$, we call knowledge states 
$s, s' \in \mathcal{S}_\EF$ \emph{canonically equivalent}, 
denoted $s \equiv^{can}_{\EF}s'$, iff
they are strongly equivalent in the canonized evolution frame
$\EF^*$, which results from $\EF$ by replacing $\compo_{\EF}(s)$ with its canonical form
$\compo_{\EF}(s)^*$ {\rm (\/}\iec $\compo_{\EF^*}(s)=\compo_{\EF}(s)^*${\rm )\/}. 
\end{definition}

Immediately, we note the following properties. 

\begin{theorem}\label{thm:entailment-equiv}
Let $\EF$
be a contracting  evolution frame.   
Then,  
\begin{enumerate}
\item\label{thm:ent-eq-1} $\EF, s\models \varphi$ iff 
$\EF^*, s\models \varphi$, for any knowledge state $s$ and any 
formula~$\varphi$.
\item\label{thm:ent-eq-2} $\equiv_{\EF}^{\can}$ is compatible 
with $\equiv_\EF$, for any $S\subseteq\mathcal{S}_\EF$, \iec 
$s\equiv_{\EF}^{\can} s'$ implies $s\equiv_{\EF} s'$, for every $s,s'\in S$.
\end{enumerate}
\end{theorem}

\begin{proof} In order to show Part~\ref{thm:ent-eq-1}, we 
consider the Kripke structures $K_{\EF}$ and $K_{\EF^*}$,
corresponding to a contracting evolution frame $\EF$ and its canonized
evolution frame, respectively.  Since, for every $s \in
S$, it holds that $\bel(\compo_\EF(s))$ = $\bel(\compo_\EF(s)^*)$ =
$\bel(\compo_{\EF^*}(s))$, equal states have equal labels in $K_{\EF}$
and $K_{\EF^*}$. Hence, $K_{\EF}$ and $K_{\EF^*}$ coincide. As a
consequence, $K_{\EF},s\models\varphi$ iff
$K_{\EF^*},s\models\varphi$, for every $s \in S$, and hence
$\EF,s\models\varphi$ iff $\EF^*,s\models\varphi$, for every $s \in
\mathcal{S}_\EF$.

As for the proof of Part~\ref{thm:ent-eq-2}, assume 
$s\equiv_{\EF}^{\can} s'$, for $s, s' \in S$ and some 
$S\subseteq\mathcal{S}_\EF$. Then, 
$\bel(\compo_{\EF^*}(s))= \bel(\compo_{\EF^*}(s'))$. Moreover, 
$\bel(\compo_\EF(s)) = \bel(\compo_{\EF^*}(s))$ holds, as well as 
$\bel(\compo_\EF(s')) = \bel(\compo_{\EF^*}(s'))$, which implies 
$\bel(\compo_{\EF}(s))= \bel(\compo_{\EF}(s'))$, and the same is true for all 
corresponding successor states of $s$ and $s'$ due to the fact that they are 
canonically equivalent. Thus, $s\equiv_{\EF}^{\can} s'$ implies 
$s\equiv_{\EF} s'$, for every $s,s'\in S$.
\hfill$\Box$
\end{proof}

\nop{***
\begin{definition} 
Let $\EF$
be an evolution frame.
We say that knowledge states $s$ and $s'$ are \emph{canonically equivalent $($relative to $\EF)$}, symbolically $s\equiv_{\EF}^{\can} s'$, iff  $s\equiv_{\EF^*}s'$. 
\end{definition}

Observe that canonical equivalence relative to the evolution frame
$\EF$ is defined in terms of the associated evolution frame $\EF^*$,
and obviously it holds that $\equiv_{\EF}^{\can}$ is compatible with
strong equivalence relative to $\EF^*$.

\begin{proof}
\hfill$\Box$
\end{proof}

Assuming that an evolution frame $\EF$ is both
regular and contracting, Theorem~\ref{thm:regular-bel} yields the
following stronger property:

\begin{theorem}
Let $\EF$
be an evolution frame which is both
regular and contracting. Then, $\equiv_{\EF}^{\can}$ is compatible with strong equivalence relative to $\EF$, \iec $s\equiv_{\EF}^{\can} s'$ implies $s\equiv_{\EF} s'$, for any knowledge state $s,s'$.
\end{theorem}

This result implies that canonical equivalence is applicable to the
inference operator $\models$, \iec the relation
$\EF,[s]_{\equiv_{\EF}^{\can}}\models \varphi$ is a well-defined
expression, and, according to
Definition~\ref{def:equivalence-inference}, we have
$\EF,[s]_{\equiv_{\EF}^{\can}}\models \varphi$ iff $\EF,s\models
\varphi$.  ***}

As a result, we may use $\equiv^{can}_{\EF}$ for filtration of $\EF$,
based on the following concept.

\begin{definition}
Let $\EF=\tuple{\at,\EC,\AC,\up,\rf,\bel}$ be an evolution frame 
and $c\geq 0$ an integer. We say that $\EF$ is \emph{$c$-bounded} if
there are functions $\alpha$, $f$, and $g$ such that
\begin{enumerate}
\item $\alpha$ is a function mapping knowledge states into sets of
events such that, for each $s=\tuple{\KB;E_1\commadots E_n}$,
$\alpha(s)=\{E_{n-c'+1}\commadots E_n\}$, where $c' = \min(n,c)$, 
and

\item $\up(s,E) = f(\bel(s),\alpha(s),E)$ and $\rf(s,A) = g(\bel(s),\alpha(s),A)$, for each knowledge state $s\in\mathcal{S}_\EF$, each event $E\in\EC$, and each  $A\subseteq\AC$.
\end{enumerate}
\end{definition}

This means that in a $c$-bounded evolution frame, the compilation $\compo_\EF(s)$  only
depends on the belief set of the predecessor $s'$ of $s$ and the
last $c+1$ events in $s$ (including the latest event). In
particular, $c=0$ means that only the 
latest event needs to be considered. 

\begin{theorem}\label{thm:c-bound}
Let $\EF=\tuple{\at,\EC,\AC,\up,\rf,\bel}$ be an evolution frame where
$\EC$ is finite, and let $S\subseteq \mathcal{S}_\EF$ be
successor closed and such that $\{
\pi_0(s) \mid s \in S \}$ is finite.
If {\rm (}i\/{\rm )} $\EF$ is contracting, 
{\rm (}ii\/{\rm )} there is some finite set $R_0\subseteq\lang_\at$ such that $\compo_\EF(s)$ contains only rules from $R_0$, for every $s\in S$,  and {\rm (}iii\/{\rm )} $\EF$ is $c$-bounded, for some $c\geq 0$, then $\equiv_{\EF}^{\can}$ has a finite
index with respect to $S$.
\end{theorem}

\begin{proof} We prove that $\equiv_{\EF}^{\can}$ has finite index \wrt $S$ by 
means of Theorem~\ref{thm:finite-index}. That is, we must show that 
$\equiv_{\EF^\ast}^0$ has finite index \wrt $S$ and that there exists a 
$k\geq 0$, such that for any two knowledge states $s,s' \in S$, 
$s \equiv_{\EF^\ast}^k s^\prime$ implies $s \equiv_{\EF}^{\can} s^\prime$.

We first show that the relation $\equiv_{\EF^\ast}^0$, \iec weak 
canonical equivalence, has finite index \wrt $S$.

For any knowledge state $s \in S$, $\compo_{\EF^\ast}(s)$ yields an update 
sequence $\useq{P}$ of at most $|R_0|$ programs, \iec 
$\useq{P}=(P_0\commadots P_n)$ and $n \leq |R_0|$ holds. To see this, suppose 
otherwise that $n > |R_0|$. Since $\compo_{\EF^\ast}(s)$ is 
canonical (and thus \emph{contracting under empty updates}), none of the 
programs $P_i$, $0\leq i \leq n$, is empty. Furthermore, since  
$\compo_{\EF}(s)$ only contains rules from $R_0$, this also holds for 
$\compo_{\EF^\ast}(s)$. Hence, there must be at least one rule $r \in R_0$, 
which occurs in at least two programs $P_i, P_j$, $0\leq i,j \leq n$, and 
$i \neq j$. This, however, contradicts the fact that $\compo_{\EF^\ast}(s)$ is 
canonical (and thus \emph{contracting under rule repetition}). Hence, 
our assumption does not hold, which proves $n \leq |R_0|$. Moreover, 
$|\bigcup_{i=0}^n P_i| \leq |R_0|$ holds for the canonical compilation 
$\useq{P}$ by the same argument: If 
$|\bigcup_{i=0}^n P_i| > |R_0|$, then there must be at least one rule 
$r \in R_0$, such that $r \in P_i \cap P_j$ for at least two programs 
$P_i, P_j$, $0\leq i < j \leq n$; this contradicts
that $\compo_{\EF^\ast}(s)$ is contracting.

As a consequence, we can roughly estimate the number of different 
canonical compilations $\compo_{\EF^\ast}(s)$ by 
\[
d=2^{|R_0|-1}(|R_0|+1)!=
\mathcal{O}(2^{|R_0|(\log|R_0|+1)})
\]
(note that 
\(
(|R_0|+1)! \leq
2^{1+(|R_0|-1)\log(|R_0|+1)} \leq 2^{1+|R_0|\log|R_0|}
\)
for $|R_0|>0$). 
This upper bound can be explained as follows. A canonical compilation need not 
contain all rules of $R_0$, hence we add a special fact for signaling that, 
given an ordered sequence of rules from $R_0$ and the special fact, 
a canonical compilation consists of all rules in the sequence up to the 
special fact. There are $(|R_0|+1)!$ permutations of such sequences, which is 
an over-estimate of the number of canonical compilations consisting of 
different ordered sequences of rules. For each such sequence, there are 
$2^{|R_0|-1}$ possibilities for the rules to be grouped into sequences of at 
most $|R_0|+1$ programs, respecting their order. To see this, observe that if 
we fixed a grouping into a sequence of programs for all but the last rule, 
then for the last rule there are two possibilities: It can either be added to 
the last program of the sequence, or we add a new program, consisting of the 
last rule only, to the sequence. Applying this argument recursively and 
observing that for the first rule there is only one possibility---it has to 
go into the first program of the sequence---the given bound follows. Hence, at 
most $d$ different canonical compilations $\compo_{\EF^\ast}(s)$ can be 
built for all $s \in S$. Thus, at most $d$ different belief sets $\bel(s)$ 
exist among all $s \in S$, proving that $\equiv_{\EF^\ast}^0$ has finite index 
\wrt $S$.

Secondly, we show by induction on $k\geq c$ that for any two 
knowledge states $s, s^\prime \in S$ canonical $c$-equivalence 
$s \equiv_{\EF^\ast}^c s^\prime$ implies strong canonical equivalence 
$s \equiv_{\EF}^{\can} s^\prime$, which proves our result in virtue of 
Theorem~\ref{thm:finite-index}. More precisely, we show for all $k\geq c$, 
that in the canonized evolution frame $\EF^*$, $c$-equivalence of knowledge 
states $s, s^\prime \in S$  implies their $k$-equivalence in $\EF^*$.

\medskip
\noindent
{\sc Induction Base} ($k=c$).  Canonical $c$-equivalence of knowledge states 
$s, s^\prime \in S$ trivially implies $s \equiv_{\EF^\ast}^c s^\prime$.

\medskip
\noindent
{\sc Induction Step} ($k>c$). Assume that, for any two knowledge states 
$s, s^\prime \in S$ and some $k > c$,
$s \equiv_{\EF^\ast}^c s^\prime$ implies
$s \equiv_{\EF^\ast}^{k-1} s^\prime$. We show that under this assumption 
$s \equiv_{\EF^\ast}^{k} s^\prime$
follows. 

Let $s^{ }_k=s+E_1\commadots E_k$, $s^{ }_{k-1}=s+E_1\commadots E_{k-1}$, 
$s^\prime_k=s^\prime+E_1\commadots E_k$, and 
$s^\prime_{k-1}=s^\prime+E_1\commadots E_{k-1}$.
Since $k > c$, the sets $\alpha(s^{ }_{k-1})$ and 
$\alpha(s^\prime_{k-1})$ are equal and 
$\bel(s^{ }_{k-1}) = \bel(s^\prime_{k-1})$ holds by induction hypothesis. 
Hence, the following equations hold:  
\begin{eqnarray*}
A = f(\bel(s^{ }_{k-1}), \alpha(s^{ }_{k-1}),E_k) & = &
f(\bel(s^\prime_{k-1}), \alpha(s^\prime_{k-1}),E_k), \quad \mbox{ and} \\
g(\bel(s^{ }_{k-1}), \alpha(s^{ }_{k-1}),A) & = &
g(\bel(s^\prime_{k-1}), \alpha(s^\prime_{k-1}), A).
\end{eqnarray*}
Consequently, the equality 
$\compo_{\EF}(s^{ }_k)=\compo_{\EF}(s^\prime_k)$ holds, which implies that 
$\compo_{\EF^\ast}(s^{ }_k)=\compo_{\EF^\ast}(s^\prime_k)$, 
and thus $\bel(s^{ }_k)=\bel(s^\prime_k)$. This proves canonical 
$k$-equivalence. 

This proves that $s \equiv_{\EF^\ast}^c s^\prime$ implies 
$s \equiv_{\EF}^{\can} s^\prime$, for any two knowledge states 
$s, s^\prime \in S$. Since we have also shown that $\equiv_{\EF^\ast}^0$ has 
finite index \wrt $S$, it follows  from Theorem~\ref{thm:finite-index} that 
$\equiv_{\EF}^{\can}$ has finite index \wrt $S$.
\hfill$\Box$
\end{proof}

We remark that the existence of $R_0$ is trivial if we have a
function-free (finite) alphabet, and, as common in many logic programming
semantics, repetition of literals in rule bodies has no effect, and
thus the set of nonequivalent rules is finite. A similar remark
applies to the initial knowledge bases $\pi_0(s)$. 
\section{Complexity} 
\label{sec:complexity}

In this section, we investigate the computational complexity of  
our evolution framework. To this end, in what follows we assume that
the alphabet 
$\at$ of the evolution frames under consideration is finite and 
propositional. Thus, we only deal with finite propositional (sequences of) 
programs which are the result of the state compilation $\comp{s}$.

First, we study the computational complexity of the following reasoning task: 
\begin{description}
\item[{\rm \tempevo:}] Given an evolution frame $\EF=\tuple{\at,\EC,\AC,\up,\rf,\bel}$, a knowledge state $s$ over $\EC$, and some formula $\varphi$, does $\EF,s\models\varphi$ hold?
\end{description}

In order to obtain decidability results, we assume that the
constituents of the evolution frame $\EF$ in {\tempevo} are all
computable. More specifically, we assume that
 \begin{itemize}
\item[(i)] $\EC$, $\AC$, and $\bel$ are given as computable functions
deciding $E\in\EC$,  $a\in\AC$, and $r\in\bel(\useq{P})$, and
\item[(ii)] $\up$ and $\rf$ are given as computable functions.
\end{itemize}
Nonetheless, even under these stipulations, it is easy to see that 
\tempevo\ is undecidable. Indeed, the compilation function may efficiently 
simulate Turing machine computations, such that the classical Halting Problem can be 
encoded easily in the above reasoning problem.

The results of Section~\ref{sec:equivalence} provide a basis for
characterizing some decidable cases. We consider here the following
class of propositional evolution frames.

\begin{definition}\label{def:regular}
Let $\EF=\tuple{\at,\EC,\AC,$ $\up,\rf,\bel}$ be a propositional evolution 
frame {\rm (\/}\iec $\at$ is propositional{\rm )\/}. Then, $\EF$ is called 
\emph{regular} if the following three conditions hold:
\begin{enumerate}
\item \label{def:regular-1} The membership tests $E\in \EC$ and $r\in
\bel(\useq{P})$ are feasible in \PSPACE\ {\rm (\/}\egc located in the
polynomial hierarchy{\rm )\/}, and the functions $\up$ and $\rf$ are
computable in polynomial space {\rm (\/}the latter with polynomial
size output{\rm )\/}.

\item \label{def:regular-2} Rules in compilations $\compo_{\EF}(s)$ and events 
$E$ have size polynomial in the
representation size of $\EF$, denoted by $\|\EF\|$ {\rm (\/}i.e., repetition
of the same literal in a rule is bounded{\rm )\/}, and events have size at
most polynomial in $\|\EF\|$. 

\item \label{def:regular-3} $\bel(\cdot)$ is fully characterized by rules of
length polynomial in $\|\EF\|$, \iec there is some constant $c$ such
that $r \in \bel(\useq{P})$ iff $r \in \bel(\useq{P}^\prime)$
for all rules $r$ of length $\leq \|\EF\|^c$ implies
$\bel(\useq{P})=\bel(\useq{P}^\prime)$, for all update sequences
$\useq{P}$ and $\useq{P}^\prime$.
\end{enumerate} 
\end{definition}

Conditions~\ref{def:regular-1} and~\ref{def:regular-3} apply to the
approaches
in~\cite{alfe-etal-99a,eite-etal-00f,eite-etal-01,mare-trus-94,mare-trus-98,inou-saka-99,foo-zhan-98},
and Condition~\ref{def:regular-2} is reasonable to impose; note that
none of these semantics is sensible to repetitions of literals in rule
bodies. However, we could imagine semantics where, similar as in
linear logic, literals are ``consumed'' in the inference process, and that
repetition of literals alludes to available resources.

Before we state our first complexity result, let us briefly recall
some well-known complexity results for the above-mentioned
approaches. Deciding whether a literal $L \in \bel(P)$, for a literal
$L$ and a finite, propositional ELP $P$ is \coNP-complete. The
complexity does not increase for the update approaches
in~\cite{alfe-etal-99a,eite-etal-00f,eite-etal-01,mare-trus-94,mare-trus-98,foo-zhan-98},
\iec deciding $L \in \bel_S(\useq{P})$ is \coNP-complete for $S \in
\{E, \oplus, \mathit{Rev}, \mathit{PLP} \}$, where $\useq{P}$ is a
finite, propositional sequence of (at most two in case of PLP)
ELPs. However, the complexity for abductive theory
updates~\cite{inou-saka-99}, when considering all possible selection
functions, increases one level in the polynomial hierarchy: Deciding
$L \in \bel_{\mathit{Abd}}(\useq{P})$ is $\PiP{2}$-complete.

The following lemma will be used several times in the sequel.

\begin{lemma}\label{lemm:model-check}
Given a regular evolution frame $\EF=\tuple{\at,\EC,\AC,$
$\up,\rf,\bel}$, a knowledge state $s \in \mathcal{S}_{\EF}$, and a
formula $\varphi$, suppose that $\equiv_{\EF}$ has finite index, $c$,
\wrt $S=\dsc(s)$.  Then, there exists a deterministic Turing machine $M$ which
checks $\EF,s\models\varphi$ in space polynomial in $(q+1)\cdot(m_s +
\log c + \|\EF\|+\|\varphi\|)$, where $q$ is the nesting depth of
evolution quantifiers in $\varphi$, $m_s$ is the maximum space required
to store $s' \in S$ representing a class of $S/\equiv_\EF$, and
$\|\varphi\|$ denotes the size of formula $\varphi$.
\end{lemma}

\begin{proof} We first show that for evaluating evolution quantifiers
$\E\psi$ or $\A\psi$, we may consider finite paths of length $c$ in
$\EF$. Note that every path of length greater than $c$ in $\EF$ must
contain at least one pair of strongly equivalent knowledge states.
While this is trivial if $\psi$ is of form $\X\psi_1$, consider the
case where $\psi$ is of form $\psi_1\U\psi_2$.

If a path $p$ starting at $s$ of arbitrary length exists such that $\EF,p \models \psi_1\U\psi_2$,
then there exists also a path $p'$ starting at $s$ and of length at
most $c$, such that $\EF,p'\models \psi_1\U\psi_2$. To see this, note
that $\EF,p \models \psi_1\U\psi_2$ implies that $\psi_1\U\psi_2$ is
satisfied in a finite path $p''$ which is an initial segment of $p$. We can repeatedly
shorten $p''$ to
obtain $p'$ as follows. For any pair of strongly equivalent
knowledge states $s=p''_i$ and $s'=p''_j$ such that $i<j$, consider the
sequence $p''_{i},\ldots,p''_{j}$ of knowledge states between
them. If $\psi_2$ is satisfied by none of them, we can cut
$p''_{i+1},\ldots,p''_{j}$ and replace each state $p''_{i+1}$,  $p''_{i+2}$,
\ldots\ by an equivalent successor of $p''_{j}$, $p''_{j+1}$ such that
we obtain a (finite) path in $\EF$. 
Otherwise, \iec if $\EF,p''_l\models \psi_2$ for some
$l\in\{i,\ldots,j\}$, we can cut $p''$ immediately
after the first such $p''_l$. It is easily verified that the
resulting path $p'$ has length at most $c$ and still satisfies
$\psi_1\U\psi_2$.

Now consider the case $\A(\psi_1\U\psi_2)$. Obviously, if
$\psi_1\U\psi_2$ is satisfied by all finite paths of length $c$
starting at $s$, it will also be
satisfied by all paths of arbitrary length. To see the converse 
direction, assume there is an infinite path $p$
starting at $s$ such  that $\EF, p\not\models \psi_1\U\psi_2$. We
show that then a finite path $p'$ of 
length at most $c$ starting at $s$ exists such that  $\EF,
p'\not\models \psi_1\U\psi_2$. Observe that either (i) 
$\psi_2$ is false in every $p_i$, $i\geq 0$, or (ii) there exists some $i
\geq 0$ such that $\EF,p_i \models \fneg\psi_1 \land \fneg \psi_2$, 
and $\EF,p_j \models \psi_1\land \fneg \psi_2$, for every $j \in \{0,\ldots,i-1\}$. 
In case (i), we can, as above, transform the initial segment
$p''=p_0,p_1,\ldots,p_{c-1}$ of $p$
by repeatedly removing sequences between pairs of strongly equivalent
knowledge states and eventually obtain a path $p'$ as claimed. 
In case (ii), we start with $p''=p_0,p_1,\ldots,p_{i}$ and again repeatedly remove sequences between pairs of 
strongly equivalent knowledge states to obtain a path $p'$ of length at most $c$ starting at $s$
such that $\EF, p'\not\models \psi_1\U\psi_2$.
Hence, if all infinite paths $p$ starting at $s$ satisfy $\psi_1\U\psi_2$, then so do all 
paths of length $c$ starting at $s$. 

This proves that if there are at most $c$ strongly inequivalent 
descendants of $s$, it suffices to consider paths of length $c$ to 
prove whether $\EF,s\models \varphi$.

\medskip
\noindent
Now an algorithm for deciding $\EF,S\models \varphi$ is as follows. Starting at $s$, it
recursively checks the satisfiability of $\varphi$ by checking the
satisfiability of its subformulas and evaluating Boolean
connectives. For any subformula $\varphi'$ of form $\E\psi$
(resp., $\A\psi$),  guess nondeterministically, step by
step, a path $p$ in $\EF$ starting at $s$ in order to witness
$\EF,p\models \psi$ (resp., refute $\EF,p\models \psi$ and
exploit $\A\psi \equiv \fneg\E\fneg\psi$) and check this by iterating
through $p$ for (at most) $c$ steps, using a counter. The counter
occupies space $\log c$ in a standard binary coding.  Per nesting
level, the algorithm requires space for one counter and for one descendant of
$s$, which is bounded by $m_s$. Furthermore, due to the fact that $\EF$
is regular, $\varphi\in \bel(s)$ can be checked, for all $s \in
\mathcal{S}_{\EF}$ and atomic $\varphi$, in space polynomial in $s$,
$\EF$, and $\|\varphi\|$. Hence at each level, the algorithm runs in space $\Delta$
which is polynomial in $m_s + \log c + \|\EF\|+\|\varphi\|$. 

By applying Savitch's Theorem in the formulation for Turing machines
with oracle access (cf.\ Theorem 2.27 in \cite{balc-etal-88}), we can
show by induction on the evolution quantifier depth $q\geq 0$ of
a formula $\varphi$, that deciding $\EF,s\models \varphi$ is feasible on
a deterministic Turing machine $M$ using space at most
$(q+1)\Delta^2$. Savitch's Theorem states that if language $A$ can be
decided by a nondeterministic Turing machine with oracle set $B$ in
space $f(n)$, then it can be decided by a deterministic Turing machine
with oracle set $B$ in space $f(n)^2$, providing $f(n) \geq \log n$.
Furthermore, $f(n)$ must be space constructible (which is the case in
our application of the lemma).

\medskip 
\noindent
{\sc Induction Base} ($q=0$). Since the above algorithm operates deterministically in space
$\Delta \leq \Delta^2$, the existence of $M$ is obvious.

\medskip
\noindent
{\sc Induction Step} ($q>0$). Assume that formulas of evolution
quantifier depth $\leq q-1$ can be decided in deterministic space
$q\cdot\Delta^2$ on some Turing machine $M'$, and let $\varphi$ have
evolution quantifier depth $q$. If $\varphi$ is of form $\E\psi$
(resp., $\A\psi$), then the above algorithm amounts to a
nondeterministic oracle Turing machine $M'$ using work space bounded
by $\Delta$ and calling an oracle for deciding subformulas of form
$\E\psi'$ (resp., $\A\psi'$). By Savitch's Theorem, there is a
deterministic Turing machine $M''$ using work space at most $\Delta^2$
which is equivalent to $M'$ and uses the same oracle set. By the
induction hypothesis, the oracle queries can be deterministically
decided in space $q\cdot\Delta^2$. Hence, from $M''$ we can construct a
deterministic Turing machine $M$ deciding $\EF,s\models \varphi$ which
operates in work space $\Delta^2 + q\cdot \Delta^2 = (q+1)\Delta^2$.
This $M$ is easily extended to decide all $\varphi$ of evolution
quantifier depth $q$ within the same space bound. This concludes the
induction and the proof of the lemma.
\hfill$\Box$ 
\end{proof}

We then obtain the following complexity results. 

\begin{theorem}\label{thm:complexity}
Deciding $\EF,s\models\varphi$, given a
regular propositional evolution frame
$\EF\!=\tuple{\at,\EC,\AC,\up,\rf,\bel}$, a knowledge state $s$, and a
formula $\varphi$ is
\begin{enumerate}
\item\label{thm:complexity:1} \TWOEXPSPACE-complete, if $\bel(\cdot)$
is $k$-local for some $k$ which is polynomial in $\|\EF\|$, and
$\compo_\EF(\cdot)$ is incremental;

\item\label{thm:complexity:2} \EXPSPACE-complete, if $\EF$ is
$c$-bounded, where $c$ is polynomial in $\|\EF\|$, contracting, and functions 
$\up$ and $\rf$ are computable in space polynomial in the size of 
$\compo_{\EF}(\cdot)$.

\item\label{thm:complexity:3} \PSPACE-complete, if $\EF$ is as in \ref{thm:complexity:2}
and, moreover, all rules in the compilations $\compo_{\EF}(s')$ of 
descendants $s'$ of $s$ are from a set $R_0$ of size polynomial in $\|\EF\|$.
\end{enumerate}
\end{theorem}

\begin{proof} We first prove the upper bounds of these results. Recall that we 
assume a finite propositional alphabet $\at$. Hence, by
Condition~\ref{def:regular-3} of a regular evolution frame $\EF$, there are only finitely 
many different belief sets $\bel(s)$. Indeed, the number of rules of
length $L$ is bounded by $(4|\at|)^L$, and hence there are
$\mathcal{O}(2^{\|\EF\|^{l_1}})$ (single exponential in $\EF$) many rules, where $l_1$ is some
constant, which are relevant for characterizing belief sets, and 
there are $\mathcal{O}(2^{2^{\|\EF\|^{l_1}}})$, \iec double exponentially many, different belief sets $\bel(\useq{P})$.
This implies that $|S/\equiv_\EF^0\!\!| \leq d$ where $d = \mathcal{O}(2^{2^{\|\EF\|^{l_1}}})$ 
for any set $S \subseteq\mathcal{S}_\EF$. Observe also 
that $\EC$ is finite and $|\EC|= \mathcal{O}(2^{\|\EF\|^{l_2}})$,
for some constant $l_2$. This follows from the finiteness of $\at$ and the 
fact that rules in events, as well as events themselves, have size at most 
polynomial in $\|\EF\|$. In particular, there exist
$\mathcal{O}(2^{\|\EF\|^{l_{2,1}}})$ 
many different rules in events, for some constant
$l_{2,1}$, and thus there are
$\mathcal{O}(2^{\|\EF\|^{l_{2,1}}})^{\|\EF\|^{l_{2,2}}} = \mathcal{O}(2^{\|\EF\|^{l_2}})$ 
many different events in $\EC$, for some constants $l_{2,2}$ and $l_2$. 
In the following, let $S = \dsc(s)$.

\medskip
\noindent
\emph{Membership, Part~\ref{thm:complexity:1}}. In order to prove an upper bound for Part~\ref{thm:complexity:1} of the 
theorem, since $\bel(\cdot)$ is $k$-local, $\compo_{\EF}(\cdot)$ is 
incremental, and $\equiv_\EF^0$ has finite index \wrt $S$, which is successor closed,
Theorem~\ref{thm:finite-index-sequences} can be applied, establishing that 
$\equiv_\EF$ has finite index \wrt $S$. 
Recall from the proof of Theorem~\ref{thm:finite-index-sequences} that an 
upper bound for $|S/\equiv_\EF\!\!|$ is given by $d^{\,2|\EC|^k}$, where $k$ 
is polynomial in $\|\EF\|$, \iec $k=\|\EF\|^{l_3}$ for some constant 
$l_3$. Furthermore, $|\EC| = \mathcal{O}(2^{\|\EF\|^{l_2}})$, for 
constant $l_2$. 
Hence, we obtain that 
there are at most  
$$
d^{(2\mathcal{O}(2^{\|\EF\|^{l_2}})^{\|\EF\|^{l_3}})} = 
d^{\mathcal{O}(2^{\|\EF\|^{l_2+l_3}})} =
{\mathcal{O}(2^{2^{\|\EF\|^{l_1}}})}^{\mathcal{O}(2^{\|\EF\|^{l_2+l_3}})} = 
\mathcal{O}(2^{2^{\|\EF\|^{l}}}),$$ 
\iec double exponentially many knowledge states $s' \!\in \! S$ which 
are pairwise not strongly equivalent, for some constant $l$; in other words, $|S/\equiv_\EF| = \mathcal{O}(2^{2^{\|\EF\|^{l}}})$. 
Furthermore, we can store a representative, $s'$, of every class in 
$S/\equiv_\EF$ by storing $\KB$ and at most double exponentially many events. 
Since every event can be 
stored in polynomial space, overall double exponential space is 
sufficient to store $s'$.
By application of Lemma~\ref{lemm:model-check}, 
$\EF,s\models\varphi$ can be verified in space polynomial in 
$(q+1)\cdot(m_s+\log c + \|\EF\| +\|\varphi\|)$. We have shown above
that $m_s$ satisfies $m_s=\mathcal{O}(2^{2^{\|\EF\|^{l_0}}})$, for some constant $l_0$. Furthermore, we 
have shown that the index of $\equiv_{\EF}$ \wrt $S$, $c$, satisfies $c=\mathcal{O}(2^{2^{\|\EF\|^{l}}})$.
Hence, $\log c = \mathcal{O}(2^{\|\EF\|^{l'}})$, for a constant $l'$.
Consequently, $\EF,s\models\varphi$ can be verified in \TWOEXPSPACE.

\medskip
\noindent
\emph{Membership, Part~\ref{thm:complexity:2}}. An upper bound for Part~\ref{thm:complexity:2} of the theorem can be obtained 
as follows. The fact that $\equiv_\EF^0$ has finite index \wrt $S$ 
implies that $\equiv_{\EF^\ast}^0$ has finite index \wrt $S$, too. And, as we 
have shown in the proof of Theorem~\ref{thm:c-bound}, canonical $c$-equivalence 
implies strong canonical equivalence in a $c$-bounded, contracting evolution 
frame. Thus, $\equiv_{\EF}^{\can}$ has finite index \wrt $S$ by 
Theorem~\ref{thm:finite-index}. 
Furthermore, according to Theorem~\ref{thm:entailment-equiv}, 
$\equiv_{\EF}^{\can}$ is compatible with $\equiv_{\EF}$. 
Hence, we may represent $\bel(s')$, $s' \in S$, by the canonical 
compilation $\compo_{\EF^\ast}(s')$, together with the last $c$ events
in $s'$, where  $c$ is polynomial in $\|\EF\|$. The polynomial size bound for 
rules in $\compo_{\EF}(s')$ also holds for $\compo_{\EF^\ast}(s')$ and thus, 
since $\EF$ is contracting, $\|\compo_{\EF^\ast}(s')\|$ is bounded by the number of 
different rules, which is $\mathcal{O}(2^{\|\EF\|^{l}})$, for some constant $l$. 
Recalling the bound of $\mathcal{O}(2^{|R_0|(\log|R_0|+1)})$  for the number of different canonical compilations 
from the proof of Theorem~\ref{thm:c-bound}, we obtain that there are 
$$ 
\mathcal{O}(2^{2^{\|\EF\|^{l'}}\cdot(\log 2^{\|\EF\|^{l'}}+1)}) = 
\mathcal{O}(2^{2^{\|\EF\|^{l''}}}),
$$ 
\iec double exponential many different canonical compilations, where $l'$ and $l''$ are suitable constants. Multiplied with the number of possibilities for the last $c$ events, $|\EC|^c$, which is 
single exponential, as  
$$
|\EC|^c =  \mathcal{O}(2^{\|\EF\|^{l_2}})^c =  
\mathcal{O}(2^{\|\EF\|^{h}}),
$$
for some constant $h$, we obtain again that there are at most double exponentially many knowledge 
states $s' \in S$ which are pairwise not strongly equivalent. However, 
since $\|\compo_{\EF^\ast}(s')\|\leq|\at|^{\|\EF\|^{l_1}}=
\mathcal{O}(2^{\|\EF\|^{h'}})$, for some constant $h'$, we can represent every 
strongly inequivalent descendant $s'\in S$ of $s$, using 
$\compo_{\EF^\ast}(s')$ together with the last $c$ events in 
single exponential space. Thus, \EXPSPACE~membership follows from 
Lemma~\ref{lemm:model-check}. 

\medskip
\noindent
\emph{Membership, Part~\ref{thm:complexity:3}}. Next, we prove \PSPACE~membership for Part~\ref{thm:complexity:3} of the 
theorem.
The additional condition on $\compo_\EF(s')$ that all rules are from a set 
$R_0$ of size polynomial in $\|\EF\|$ guarantees that 
$\|\compo_{\EF^\ast}(s')\|$ is polynomial in the size of $\EF$, for any 
$s' \in S$. Using the same estimate as above, we thus obtain at most single 
exponentially many different canonical compilations, for states $s'$. 
Multiplied with the exponential number of possibilities for the last $c$ 
events, we now obtain at most single exponentially many strongly inequivalent 
descendants of $s$. For storing them, we use again $\compo_{\EF^\ast}(s')$ 
together with the last $c$ events, requiring the space of 
$\|\compo_{\EF^\ast}(s')\|$ plus $c$ times the space of an event. Since 
$\|\compo_{\EF^\ast}(s')\|$, $c$, and the size of an event are all  
polynomial in the size of $\EF$, overall polynomial space is needed for 
representation, establishing \PSPACE~membership in virtue of 
Lemma~\ref{lemm:model-check}.

\medskip
\noindent
\emph{Hardness}. 
We show the lower bounds by encoding suitable Turing machine
computations, using padding techniques, into particular evolution
frames. In order to obtain a lower bound for
Part~\ref{thm:complexity:1}, consider a regular
evolution frame $\EF$, where $\at=\{A_i\,|\,1\leq i \leq
n\}\cup\{\mathit{accept}\}$, hence, $|\at\,| = n+1$, and
$\bel(\useq{P})$, defined below, is semantically given by a set of classical
interpretations, where $\naf$ is classical negation and repetition of
literals in rule bodies is immaterial. Then, there exist
$2^{n+1}$ classical interpretations yielding $2^{2^{n+1}}$ different belief
sets $B_0\commadots B_{2^{2^{n+1}}-1}$. We assume an enumeration of 
interpretations $I_0\commadots I_{2^{n+1}-1}$, such that $I_0$ does not contain 
$\mathit{accept}$. Moreover, we consider a single event $E = \emptyset$. Let 
$l=2^{2^n}$. The number of events, $i$, encountered for reaching a 
successor  state $s'$ of $s_0$ in $i<l$ steps serves as an index of its 
belief set, \iec $\bel(s')= B_i$. For $i<l-1$, $B_i$ is obtained using 
interpretations $I_0$ and $I_j$ as models, such that the $j$-th bit, 
$1\leq j\leq\log l = 2^n$, of index $i$ is $1$. Thus, the belief sets 
$B_0\commadots B_{l-2}$ are pairwise distinct and under classical model-based 
semantics, $\mathit{accept} \notin B_i$ holds for $0\leq i \leq l-2$.

In state $s_{l-1}$, we simulate in polynomial time the behavior of an 
\TWOEXPSPACE\ Turing machine $M$ on some fixed input $I$. To this end, we 
use an action $a$ and an update policy $\up$, such that  
$\up(s,E)={a}$ for $|s| = m\cdot l+l-1$, $m\geq 0$, if $M$ accepts $I$, and 
$\up(s,E)=\emptyset$ otherwise. For all other knowledge states, 
\iec if $|s|\bmod l\neq l-1$, $\up(s,E)=\emptyset$. The realization assignment 
$\rho(s,A)$ is incremental and adds an empty program, $\emptyset$, if $A=\emptyset$, 
and the program $\{\mathit{accept}\la\ \}$ in case of $A=\{a\}$. 
The semantics $\bel(P_1\commadots P_k)$ is as follows. If $k\bmod l\neq l-1$, 
then $\bel(P_1\commadots P_k)=B_j$, where $j=k\bmod l$. Otherwise, if 
$P_k=\{\mathit{accept}\la\ \}$, then $\bel(P_1\commadots P_k)=B$, where $B$ is 
a fixed belief set containing $\mathit{accept}$, and, if $P_k\neq \{\mathit{accept}\la\ \}$, then
$\bel(P_1\commadots P_k)=B_{l-1}$, where $B_{l-1}$ is defined as $B_i$ for 
$i<l-1$, \iec $\mathit{accept}\notin B_{l-1}$.
As easy to see,
there are at most $l+1$ states $s_0\commadots s_l$ which are not
$0$-equivalent, and $1$-equivalence of two states $s$ and $s'$ implies
strong equivalence of $s$ and $s'$. To see the latter, observe that
$s\equiv_{\EF}^1 s'$ 
iff $|s|\bmod (l+1) =|s'|\bmod (l+1)$. Thus, $\bel(\cdot)$ is
local. Furthermore, it is easily verified that the functions $\up$ and
$\rho$ can be computed in polynomial time. The same is true for
deciding $r\in\bel(\useq{P})$, $\useq{P} = (P_1,\ldots,P_k)$, where we
proceed as follows. We first compute $j = k \bmod l$. If $j \neq l-1$,
then we scan the bits $b_1,b_2,\ldots,b_{\log j}$ of $j$, and for
every bit $b_{j'}$ such that $b_{j'}=1$, we compute its index, $j'$,
in binary (which occupies at most $n$ bits), and extend its
representation to length $n$ by adding leading zeros if necessary.
The resulting binary string is regarded as representation of the
interpretation $I_{j'}$, where the bits encode the truth values of the
atoms $A_1\commadots A_n$ and $\mathit{accept}$ is false. Hence, each
model $J$ of $\useq{P}$ can be computed in polynomial time; checking whether $J\models r$ is easy. Thus, deciding $r \in
\bel(\useq{P})$ is polynomial if $j\neq l-1$. Otherwise, \iec if $j =
l-1$, depending on $P_k$, $r \in B$ (resp., $r \in B_{l-1}$) can be
similarly decided in polynomial time.

Summarizing, $B_{l-1}$ contains $\mathit{accept}$ iff $M$ accepts 
$I$ iff $\EF,s_0 \models \E\F\mathit{accept}$. 
Note that the dual formula $\varphi = \A\G\mathit{accept}$ can be used if 
every $B_0\commadots B_{l-2}$ contains $\mathit{accept}$ (and $B_{l-1}$ 
contains $\mathit{accept}$ iff $M$ accepts $I$). Furthermore, the membership 
tests $E \in \EC$ and $r \in \bel(\useq{P})$, as well as the functions $\up$ and 
$\rf$, are computable in $\PSPACE$ (in fact, even in polynomial time), 
thus deciding \tempevo\ in $\EF$ is \TWOEXPSPACE-hard.

Let us now prove a lower bound for Part~\ref{thm:complexity:2} of the theorem.
Again, we consider a regular evolution frame $\EF$ over a finite alphabet 
$\at=\{A_i\,|\,1\leq i \leq n\}\cup\{\mathit{accept}\}$. 
Moreover, we consider the single event $E=\emptyset$.
Let $\compo_{\EF}(\cdot)$ be an incremental compilation 
function that compiles a knowledge state $s$, $|s|<2^n$, into a sequence of 
programs, $\useq{P}=(\{r_0\},\commadots \{r_{|s|-1}\})$, consisting of
$|s|$ programs each consisting of a single positive, non-tautological rule, 
such that all rules are pairwise distinct and do not contain $\mathit{accept}$.
Furthermore, let the semantics $\bel(\cdot)$ be given by 
$\bel(P_0,\commadots P_n)$  containing all rules which are true in the 
classical models of $P_n$. Note that under these assumptions, all states of 
length less than $2^n$ have mutually different belief sets, and 
$\compo_{\EF}(\cdot)= \compo_{\EF}^{\can}(\cdot)$.

In state $s$, $|s|=2^n$, we simulate in polynomial time the behavior of an 
\EXPSPACE\ Turing machine $M$ on input $I$.
To this end, $\up(\pi_{2^n-1}(s),E)$ returns $A=\{a\}$, 
where $a$ is an action which causes $\mathit{accept}$ to be included in the 
belief set $B_{2^n}$ iff $M$ accepts $I$, otherwise 
$\up(\pi_{2^n-1}(s),E) = \emptyset$. For all 
knowledge states $s'$, such that  $|s'|\bmod 2^n \neq 0$, 
$\up(\pi_{|s'|-1}(s'),E) = \emptyset$. Thus, $M$ accepts $I$ 
iff $\EF,s_0 \models \E\F\mathit{accept}$. Since $\EF$ is contracting and 
$0$-bounded, and since the membership tests $E \in \EC$ and 
$r \in \bel(\cdot)$, as well as the functions $\up$ and $\rf$ are computable 
in $\PSPACE$, it follows that deciding \tempevo\ in $\EF$ is 
\EXPSPACE-hard.
 
Finally, we give a proof for the lower bound of Part~\ref{thm:complexity:3} of 
the theorem, by encoding the problem of evaluating a quantified Boolean formula 
(QBF), which is well known to be \PSPACE-hard, in the $\UL$ framework: 
Let $\psi = Q_1x_1\ldots Q_nx_n\alpha$ be a QBF and let 
$\varphi = PQ_1x_1,\ldots PQ_nx_n\alpha$, where $PQ_i = \A$ if $Q_i = \forall$ 
and $PQ_i = \E$ if $Q_i = \exists$, $1\leq i \leq n$, be its corresponding 
state formula. Consider the following evolution frame $\EF_{\UL}$, where 
$\at = \{x_i, c_i \,|\, 1\leq i \leq n\}\cup\{0,1\}$, the initial knowledge 
base $\KB = \{x_i\,|\, 1\leq i \leq n\}\cup\{c_1\}$, $\EC=\{\{0\},\{1\}\}$, 
and the update policy $\up_{\UL}$ is given by the following actions:
$$
\begin{array}{lr@{~}l@{~}l}
\up_{\UL}(s,E) & = \{ & \mathbf{assert}(c_{i+1}) & \mid c_i \in \bel(s),
1\leq i\leq n-1 \} \cup \\
& \{ & \mathbf{retract}(c_i) & \mid c_i \in \bel(s), 1\leq i\leq n \}
\cup \\
& \{ & \mathbf{assert}(\neg x_i) & \mid c_i \in \bel(s), 0 \in E, 1\leq
i\leq n \} \cup \\
& \{ & \mathbf{retract}(x_i) & \mid c_i \in \bel(s),0 \in E, 1\leq i\leq n \}.
\end{array}
$$
Intuitively, a counter for events is implemented using atoms 
$c_i$, $1\leq i \leq n$, and each event, which may be $0$ or $1$, assigns a 
truth value to the variable encoded by literals over atoms 
$x_i$, $1\leq i \leq n$. Hence, $\up_{\UL}$ creates a truth assignment in $n$ 
steps. Thus, it is easily verified that $\EF_{\UL},\KB\models \varphi$ 
iff $\psi$ is true.
Note that after $n$ steps, \iec for all knowledge states $|s|\geq n$, 
$\up_{\UL}(s,E)$ is always empty. This implies 
that $\EF_{\UL}$ is $n$-bounded. Moreover, $\up$ is factual, \iec it  
consists only of facts (of update commands), 
yielding a contracting compilation function $\compo_{\UL}(\cdot)$ which  
uses only facts over $\at$. Thus, and since the membership tests $E \in \EC$ and 
$r \in \bel_E(\cdot)$, as well as the functions  $\up$ and $\rf$ are 
computable in $\PSPACE$, it follows that \tempevo\ in 
Part~\ref{thm:complexity:3}
is 
\PSPACE-hard. 
\hfill$\Box$
\end{proof}

While, for the propositional \UL\ framework, $\bel(s)$ 
depends  in general on all events in $s$, it is possible to restrict $\AC_{\UL}$ to the 
commands ${\bf assert}$ and ${\bf retract}$, by efficient coding techniques 
which store relevant history information in $\bel(s)$, such that the
compilation in $\compo_{\UL}(s)$ depends only on $\bel(\pi_{n-1}(s))$
and the last event $E_n$ in $s$, as shown in \cite{eite-etal-02c}. Furthermore, the 
policy $\up_{\UL}$ is sensible only to polynomially many rules in events, and
$\compo_{\UL}(s)$ contains only rules from a fixed set $R_0$ of rules,
whose size is polynomial in the representation size of $\EF$. Thus, by 
Part~\ref{thm:complexity:3} of Theorem~\ref{thm:complexity}, we get the 
following result.

\begin{corollary}\label{thm:epi:complexity}
Let $\EF=\tuple{\at,\EC,\AC_{\UL},\up_{\UL},\rf_{\UL},\bel_E}$ be a
propositional $\UL$ evolution frame, let $s$
be a knowledge state, and let $\varphi$ be a
formula. Then, deciding $\EF,s\models\varphi$ is \PSPACE-complete.
\end{corollary}

The encoding of the QBF evaluation problem in the proof of 
Part~\ref{thm:complexity:3} of Theorem~\ref{thm:complexity} has further interesting 
properties. The initial knowledge base used, $\KB$, is stratified and the 
resulting update policy is also \emph{stratified} and \emph{factual}
as defined in \cite{eite-etal-01,eite-etal-02c}. This means that 
$r\in \bel_E(\useq{P})$ can be decided in polynomial time for the given 
evolution frame. Since, moreover, the membership test $E\in \EC$, as well as the functions $\up$ and $\rf$ are computable 
in polynomial time, we get another corollary. To this end, we introduce the 
following notion.

\begin{definition}\label{def:strong-reg}
Let $\EF=\tuple{\at,\EC,\AC,$ $\up,\rf,\bel}$ be a propositional
evolution frame. $\EF$ is called \emph{strongly regular} if the
membership tests $E\in \EC$ and $r\in \bel(\useq{P})$ are feasible in
polynomial time, as well as $\up$ and $\rf$ are computable in
polynomial time.
\end{definition}

Now we can state the following result.

\begin{corollary}\label{thm:complexity2:3}
Deciding $\EF,s\models\varphi$, given a strongly regular propositional 
evolution frame $\EF\!=\tuple{\at,\EC,\AC,\up,\rf,\bel}$, a knowledge state 
$s$, and a formula $\varphi$, is \PSPACE-complete, if $\EF$ is
$c$-bounded, where $c$ is polynomial in $\|\EF\|$, contracting, and all rules 
in the compilations $\compo_{\EF}(s')$ of descendants $s'$ of $s$ are from a 
set $R_0$ of size polynomial in $\|\EF\|$.
\end{corollary}

Thus, concerning evolution frames according to Part~\ref{thm:complexity:3} of 
Theorem~\ref{thm:complexity}, we stay within the same complexity class if we 
suppose strong regularity. For strongly regular evolution frames according to 
Parts~\ref{thm:complexity:1} and~\ref{thm:complexity:2} of the theorem, we can 
establish the following result.

\begin{theorem}\label{thm:complexity2}
Given a strongly regular propositional evolution frame, 
$\EF\!=\tuple{\at,\EC,$ $\AC,$ $\up,\rf,\bel}$, a knowledge state $s$, and a
formula $\varphi$, deciding $\EF,s\models\varphi$ is
\begin{enumerate}
\item\label{thm:complexity2:1} \TWOEXPTIME-complete, if $\bel(\cdot)$
is $k$-local for some $k$ which is polynomial in $\|\EF\|$, and
$\compo_\EF(\cdot)$ is incremental;

\item\label{thm:complexity2:2} \EXPSPACE-complete, if $\EF$ is
$c$-bounded, where $c$ is polynomial in $\|\EF\|$, and contracting.
\end{enumerate}
\end{theorem}

\begin{proof} We first prove \TWOEXPTIME\ membership for 
Part~\ref{thm:complexity2:1} of the theorem. 

We do so by constructing a Kripke structure $K'=\tuple{S',R',L'}$ in 
double exponential time in $\|\EF\|$, such that $K', s \models \varphi$ iff $\EF,s \models \varphi$, 
and $S', R'$ are of size at most double exponential in the size of $\EF$. 
This proves \TWOEXPTIME\ membership by a well known result from model 
checking~\cite{clar-etal-99}, stating that there is an algorithm for 
determining whether $\varphi$ is true in state $s$ of 
$K'=\tuple{S',R',L'}$, running in time 
$\mathcal{O}(|\varphi|\cdot(|S'|+|R'|))$, where $|\varphi|$ denotes the 
evolution quantifier nesting depth of $\varphi$.

The Kripke structure $K'=\tuple{S',R',L'}$ results from the Kripke 
structure $K^{E,S}_{\EF}=\tuple{S,R,L}$, where $E=S/\equiv_{\EF}$, by 
restricting the labeling $L$ to atomic subformulas of $\varphi$. Let 
$\at_\varphi$ denote the set of all atomic subformulas in $\varphi$. Then, 
$S'=S$, $R'=R$, and $L'$ is the labeling function assigning to every $s\in S'$ 
a label $L'(s)= \bel(s)\cap \at_\varphi $. It is well known that 
$K',s\models\varphi$ iff $K^{E,S}_{\EF},s\models\varphi$, which in turn holds 
iff $\EF,s \models \varphi$.
Recall from the proof of Lemma~\ref{lemm:model-check} that in order to 
prove $\EF,s\models\varphi$, paths need to be considered only up to length $c$, 
where $c=|E|$ is the maximum number of strongly inequivalent descendants of 
$s$. Moreover, we can use one knowledge state as a representative for every 
equivalence class in $E$, thus $c$ strongly inequivalent knowledge states are 
sufficient. Recall also from the proof of Part~\ref{thm:complexity:1} of 
Theorem~\ref{thm:complexity} that for the given evolution frame $\EF$, $c$ is 
double exponential in $\|\EF\|$, and that there are at most 
single exponentially many different events, \iec $|\EC|=\mathcal{O}(2^{\|\EF\|^{l}})$. 
We construct $K'$ using a branch and bound algorithm that proceeds as follows.

The algorithm maintains a set $O$ of open knowledge states, as well as the 
sets $S'$, $R'$, and $L'$ of $K'$.
Initially, $O=\{s\}$, $S'=\{s\}$, $R'=\emptyset$, and 
$L'(s)=\bel(s)\cap \at_\varphi$. 
For every knowledge state $s\in O$, the algorithm removes $s$ from $O$ and 
generates all possible (immediate) successor states $s'$ of $s$.
For every such $s'$, it is checked whether it is strongly inequivalent 
to every $s\in S$. If so, $s'$ is added to $O$ and $S$, the tuple 
$\tuple{s,s'}$ is added to $R$, and $L'(s') = \bel(s')\cap \at_\varphi$ is computed. 
Otherwise, if $s'$ is strongly equivalent to a knowledge sate $s''\in S'$, 
then the tuple $\tuple{s',s''}$ is added to $R'$. The algorithm proceeds 
until $O$ is empty.

Since there are at most $c$ strongly inequivalent descendants of $s$,
the algorithm puts into $O$ at most $c$, \iec double exponentially
many knowledge states, each of which has size at most double
exponential in $\|\EF\|$.  Furthermore, since there exist at most
single exponentially many different events, in every expansion of a
node in $O$, at most exponentially many successors are generated, each
in polynomial time. Since $\bel(\cdot)$ is polynomial and $k$-local,
we can detect $s\equiv_{\EF}s'$ in single exponential time by
comparing the trees $\tree(s)$ and $\tree(s')$ up to depth $k$, 
respectively. On levels 0, 1, \ldots, $k$, $\tree(s)$ and $\tree(s')$
contain
$|\EC|^{2k}=\mathcal{O}(2^{\|\EF\|^{l}})^{2k}=\mathcal{O}(2^{\|\EF\|^{l'}})$
many nodes each, where $l'$ is some constant. For each pair $s_1$ and
$s'_1$ of corresponding nodes in $\tree(s)$ and $\tree(s')$, we must
check whether $\bel(s_1) = \bel(s'_1)$
holds. Condition~\ref{def:regular-3} of a regular evolution frame $\EF$ implies that
single exponentially many tests $r \in \bel(\comp{s_1})$ iff $r \in \bel(\comp{s'_1})$ (for all rules $r$ of length
polynomial in $\|\EF\|$) are sufficient. Strong regularity implies
that deciding $r \in \bel(s_1) = \bel(\comp{s_1})$ and $r \in
\bel(s'_1) = \bel(\comp{s'_1})$ are polynomial. Hence, deciding
$\bel(s_1) = \bel(s'_1)$ is feasible in single 
exponential time in
$\|\EF\|$.

Summing up, testing for (at most) double exponentially many
knowledge states $s$ times single exponentially many successor states
$s'$ whether $s \equiv_{\EF} s'$ can be done in
\[
\mathcal{O}(2^{2^{\|\EF\|^{l_1}}} \cdot 2^{\|\EF\|^{l_2}} \cdot
2^{\|\EF\|^{l_3}}) = \mathcal{O}(2^{2^{\|\EF\|^{l}}})
\]
time, for
constants $l_1, l_2, l_3$ and $l$. Thus, the overall
algorithm proceeds in double exponential time, \iec $K'$ can be computed in
in double exponential time. This proves \TWOEXPTIME\ membership of $\EF,s \models
\varphi$.

Hardness follows from a suitable encoding of \TWOEXPTIME\ Turing machines $M$. 
To this end, a similar construction as in the hardness proof of 
Part~\ref{thm:complexity:1} of Theorem~\ref{thm:complexity} can be used, where 
the update policy $\up(s,E)$ simulates a \TWOEXPTIME\ Turing machine rather 
than a \TWOEXPSPACE\ Turing machine; note that the components of $\EF$
there have polynomial time complexity.

\medskip
\noindent
We prove Part~\ref{thm:complexity2:2} of the theorem by showing that the 
lower bound does not decrease when demanding strong regularity. To this end, 
we encode the computations of an \EXPSPACE\ Turing machine, $M$, into a strongly 
regular evolution frame $\EF$, such that $\EF$ is $c$-bounded, where $c$ is 
polynomial in $\|\EF\|$, and contracting.

Assume that $M$ has binary tape alphabet $\{0,1\}$ and runs in space
$2^l$, where $l$ is polynomial in $\|\EF\|$. Let us consider the
following strongly regular evolution frame $\EF$, where
$\at=\{A_i\,|\,1\leq i \leq l\}\cup\{P_i\,|\,1\leq i \leq
l\}\cup\{Q_i\,|\,1\leq i \leq m\}\cup\{\mathit{Q_{accept}}\}$, hence,
$|\at\,| = 2l+m+1 = k$.  Then, there exist $2^k$ classical models,
which we use to represent the configuration of $M$ as follows. Atoms $Q_i$, 
$1\leq i \leq m$, and $\mathit{Q_{accept}}$ encode the state of $M$. 
Literals over atoms $A_i$ and $P_i$, $1\leq i \leq l$, are 
used to represent an index of $M$'s tape in binary format. We use a 
``disjunctive'' semantics as follows. Observe that we 
could use conjunctions of literals over atoms $Q_i$ and $P_i$ to encode the 
current state of $M$ and the position of $M$'s head, and conjunctions of 
literals over atoms $A_i$ to encode the fact that the tape cell at 
the encoded index contains $1$. By building the 
disjunction of a set of such conjunctions, we get a formula in disjunctive 
normal form (DNF) describing by its models the current configuration of $M$. 
We can make use of this observation by stipulating that we use rules to 
describe anti-models, \iec interpretations which 
are not models of the current knowledge base $\KB$. 
By defining $\bel(\cdot)$, taking the negation of the conjunction of all rules, 
we get a DNF, describing the models of $\KB$, as intended, and 
$r \in \bel(\KB)$ can be computed in polynomial time 
(by checking whether $r$ is entailed by every disjunct), as required. 
Furthermore, the semantics for sequences of programs $\bel(P_1\commadots P_n)$ 
is defined by the semantics of their union $\bel(P_1\cup\cdots\cup P_n)$. 

We simulate in polynomial time the behavior of the \EXPSPACE\ Turing machine 
$M$ on input $I$ as follows. Without loss of generality, we assume that the 
leftmost cell of $M$'s tape (the cell at index $0$) is always $1$, that $M$ 
initially is in state $Q_0$ and its head is in Position~$0$, and that $M$ uses 
the first steps to write $I$ to the tape (without accepting). Hence, the 
initial configuration can be represented by the single disjunct:
$$
Q_1\wedge\neg Q_2\wedge\ldots\wedge\neg Q_m\wedge
\neg \mathit{Q_{accept}}\wedge\neg P_1\wedge\ldots\wedge\neg P_l\wedge
\neg A_1\wedge\ldots\wedge\neg A_l.
$$ 
Thus, the initial knowledge base $\KB$ consists of the single constraint:
$$
\la Q_1,\neg Q_2\commadots\neg Q_m, \neg \mathit{Q_{accept}},
\neg P_1\commadots\neg P_l,\neg A_1\commadots\neg A_l.
$$
We use a single event $E=\emptyset$ as the tick of the clock and let 
$\up(s,E)$ implement $M$'s transition function. That is, $\up(s,E)$ is a set  
$A$ of actions $\mathit{insert}(r)$ and $\mathit{delete}(r')$, where $r$ and 
$r'$are constraints over  $\at$. Furthermore, we use $\rf_\pm$ 
(see Section~\ref{sec:frame}) for adding (resp., removing) the 
constraints to (resp., from) the knowledge base $\KB$, which amounts to 
the addition (resp., removal) of corresponding disjuncts to 
(resp., from) the DNF representing the current configuration of $M$. 
Since for every transition of $M$ at most $|\KB|+1$ rules need to be inserted 
and at most $|\KB|$ rules need to be removed, $\up$ and $\rf_\pm$ are 
polynomial in the representation size of the belief set.
Moreover, $M$ accepts $I$ iff $\EF,KB \models \E\F\mathit{Q_{accept}}$. 
Since $\EF$ is contracting and $0$-bounded, and since the membership tests 
$E \in \EC$ and $r \in \bel(\cdot)$, as well as the functions $\up$ and $\rf$ 
are computable in time polynomial in the size of $\EF$, deciding \tempevo\ in 
$\EF$ is \EXPSPACE-hard.
\hfill$\Box$
\end{proof}

\begin{table}[t]
\caption{Complexity results for regular and strongly regular evolution frames.}
\label{table:complexity}
\begin{center}
\renewcommand{\arraystretch}{1.5}
 \begin{tabular}{|l||r@{}l|r@{}l|}
 \hline 
 ~ evolution frame $\EF$ & \multicolumn{2}{c|}{regular} & \multicolumn{2}{c|}{strongly regular} \\
 \hline\hline
~(1):~ $k$-local \& incremental & ~ \TWOEXPSPACE&-complete ~ & ~ \TWOEXPTIME&-complete ~ \\
~(2):~ $c$-bounded \& contracting ~ & \EXPSPACE&-complete & \EXPSPACE&-complete \\
~(3):~ (2) \& $|R_0|$ polynomial & \PSPACE&-complete & \PSPACE&-complete
 \\ \hline 
 \end{tabular}
 \end{center}
\end{table}

The complexity results obtained so far are summarized in
Table~\ref{table:complexity}.  Further results can be derived by
imposing additional meaningful constraints on the problem instances.  We
remark that if we restrict the semantics for $\bel(\cdot)$
to be defined in terms of a unique model (\egc the extended well-founded
semantics for ELPs \cite{brew-96,pere-alfe-92}), then in case of a $c$-bounded and contracting
regular evolution frame $\EF$, the complexity of deciding \tempevo\
drops from \EXPSPACE\ to
\PSPACE. This can be argued by the observation that, in case of a unique
model semantics, we have only single exponentially many different
belief sets, and a knowledge state $s$ can be represented by storing
the (unique) model of $\comp{s}$ and the last $c$ events, which is
possible in polynomial space. On the other hand, already for
$0$-bounded, contracting, strongly regular evolution frames with
polynomial-size rule set $R_0$, the problem \tempevo\ is \PSPACE-hard, as
can be shown by adapting the construction in the proof of
Part~\ref{thm:complexity:3} in Theorem~\ref{thm:complexity:3} to, e.g.,
evolution frames based on stratified or well-founded semantics
for ELPs \cite{brew-96,pere-alfe-92}. 

\subsection{Complexity of state equivalence} \label{sec:comp-equi} 

We conclude our complexity analysis with results concerning
weak, strong, and $k$-equivalence of two finite propositional update programs 
under $\bel_E(\cdot)$ and $\bel_{\oplus}(\cdot)$, respectively. 

We can state our first result, concerning the complexity of deciding 
weak equivalence under $\bel_E(\cdot)$, as a consequence of 
Lemma~\ref{lemm:belE} (cf.~Section~\ref{sec:loc-bel}).

\begin{theorem}\label{theo:weak}
Deciding whether two given finite propositional update programs $\useq{P}$
and $\useq{Q}$ are weakly equivalent under $\bel_E$, i.e., 
satisfying $\bel_E(\useq{P})=\bel_E(\useq{Q})$, is \coNP-complete.
\end{theorem} 

\begin{proof}
Membership follows from Lemma~\ref{lemm:belE}: The problem of checking 
$\bel_E(\useq{P})=\bel_E(\useq{Q})$ for finite propositional update programs 
$\useq{P}$ and $\useq{Q}$ is equivalent to the task of checking whether they 
yield the same answer sets, \iec whether $\ASup(\useq{P})=\ASup(\useq{Q})$, which 
is in \coNP.

For the lower bound, suppose that, without loss of generality, $\useq{P}$ has no 
answer set. Then checking whether $\bel_E(\useq{P})=\bel_E(\useq{Q})$,  
for an update sequence $\useq{Q}$, amounts to the task of testing whether 
$\useq{Q}$ has no answer set, which is \coNP-complete 
(cf.\ \cite{dant-etal-01}).  
\hfill$\Box$
\end{proof}

For deciding $1$-equivalence, the following lemma is useful:

\begin{lemma}\label{lemma:one-equiv}
Let $\useq{P}$ and $\useq{Q}$ be finite propositional update programs over 
possibly infinite alphabets.  Then,
$\useq{P}$ and $\useq{Q}$ are not 1-equivalent under $\bel_E(\cdot)$ iff
there is an ELP $P$ and a set $S$ such that
{\rm (}i\/{\rm )} $S \in \ASup(\useq{P}+P)$ but $S\notin \ASup(\useq{Q}+P)$, or vice versa, 
{\rm (}ii\/{\rm )}~$|S|$ is at most the number of different literals in $\useq{P}+\useq{Q}$ plus $1$, and 
{\rm (}iii\/{\rm )}~$|P|\leq |S|+1$. 
{\rm (}Note that $P$ has polynomial size in the size of $\useq{P}$ and~$\useq{Q}$.{\rm )}
\end{lemma}

\begin{proof}
Intuitively, this holds since any answer set $S$ of $\useq{P}+P$ can
be generated by at most $|S|$ many rules. Furthermore, if $S$ is not
an answer set of $\useq{Q}+P$, by unfolding rules in $P$ we may
disregard for an $S$ all but at most one literal which does not occur in
$\useq{P}$ or $\useq{Q}$. To generate a violation of $S$ in
$\useq{Q}+P$, an extra rule might be needed; this means that a $P$ with $|P|\leq
|S|+1$ is sufficient.

\medskip
\noindent
\emph{If part.} Let $\useq{P}$ and $\useq{Q}$ be finite propositional update 
programs and $S$ a set such that Conditions (i), (ii), and (iii) hold. 
Then, $\useq{P}$ and 
$\useq{Q}$ are not $1$-equivalent, since, by Lemma~\ref{lemm:belE}, 
$\bel_E(\useq{P}+P) \neq \bel_E(\useq{Q}+P)$ follows from (i). 

\medskip
\noindent
\emph{Only-if part.} Let $\useq{P}$ and $\useq{Q}$ be finite propositional 
update programs which are not $1$-equivalent, \iec there exists an ELP $P$ 
such that $\bel_E(\useq{P}+P) \neq \bel_E(\useq{Q}+P)$. Moreover, again by 
application of Lemma~\ref{lemm:belE}, there exists a set $S$ such that, 
without loss of generality, $S \in \ASup(\useq{P}+P)$ but 
$S\notin \ASup(\useq{Q}+P)$, \iec Condition (i) holds. 
 
By means of $P$ and $S$, we construct a program $P^\prime$ and a set 
$S^\prime$ such that Conditions~(i), (ii), and (iii) hold for $P^\prime$ and 
$S^\prime$: Consider the program 
$\Pi_1 = ((\useq{P}\cup P)\setminus \rs(S,\useq{P}+P))^S$. Then, according to 
the update answer set semantics, $S$ can be generated from the rules in $\Pi_1$ 
by means of constructing its least fixed-point. 
Moreover, this still holds for the following simplification $P_0^S$ of $P^S$. 
First, all rules which are not applied when constructing $S$ can be removed. 
Second, among the remaining rules, we delete all rules with equal heads, 
except one of them, namely the rule which is applied first in the least 
fixed-point construction of $S$. (If several rules with equal head are applied at this level of the fixed-point 
construction, then we keep an arbitrary of them.)
Thus, $P_0^S$ consists of $k$ positive rules, $r_1 \commadots r_k$, with $k$ 
different heads, $L_1, \commadots L_k$, which are exactly those literals 
derived by $P_0^S$. Hence, $|P_0^S| \leq |S|$.  

We will create the program $P^\prime$ from $P_0^S$ by  employing unfolding. This 
means that some of the literals $L_i$, $1\leq i\leq k$, will be eliminated 
by replacing every rule $r \in P_0^S$ such that $L_i \in \body{r}$
by a rule $r'$ such that $\head{r'}=\head{r}$ and 
$\body{r'}=(\body{r}\setminus\{L_i\})\cup\body{r_i}$, where $r_i$ is the 
(single) rule having $\head{r_i}=L_i$. Consider the program 
$\Pi_2 = (\useq{Q}\cup P\setminus \rs(S,\useq{Q}+P))^S$. Since $S$ is not 
in $\ASup(\useq{Q}+P)$, it has a least fixed-point different from $S$. There are 
two scenarios: 
\begin{enumerate} 
\item Some literal $L_i \in S$ cannot be derived in $\Pi_2$. Let 
$S^\prime = \{L \in S\,|\,L$ occurs in $\useq{P}+\useq{Q}\} \cup \{ L_i\}$, 
and construct the program $P^\prime$ from $P_0^S$ by unfolding, eliminating 
all literals $L \notin S^\prime$.  
\item All literals $L_i \in S$ can be derived, as well as some literal 
$L_{k+1} \notin S$ is derived by a rule $r \in P^S
\setminus P_0^S$. Let 
$S^\prime = \{L \in S\,|\,L \mbox{ occurs in } \useq{P}+\useq{Q}\} \cup \{ L_{k+1}\}$, and 
build $P^\prime$ from $P_0^S$ by adding the rule 
$r_{k+1}\,:\, L_{k+1} \la \body{r}$ and eliminating all 
literals $L\notin S^\prime$ from the resulting program by unfolding.
\end{enumerate}

Then, in both cases, $S^\prime$ is not a least fixed-point of 
$((\useq{Q}\cup {P^\prime}) \setminus 
\rs(S^\prime,$ $\useq{Q}+P^\prime))^{S^\prime}$, while it is a least 
fixed-point of $((\useq{P}\cup {P^\prime}) 
\setminus \rs(S^\prime,\useq{P}+P^\prime))^{S^\prime}$. This proves Condition~(i). 
Furthermore, $|S^\prime|$ is at most the number of different literals in 
$\useq{P}+\useq{Q}$ plus $1$, and $|P_0^S| \leq |S|$ implies 
$|P^\prime| \leq |S^\prime|+1$. Hence, Conditions (ii) and (iii) hold.
\hfill$\Box$
\end{proof}

\begin{theorem}
Deciding strong equivalence {\rm (}or $k$-equivalence, for a given
$k\geq 0${\rm )} of two given finite propositional update programs 
$\useq{P}$ and $\useq{Q}$ over possibly infinite alphabets,  
is $\coNP$-complete under $\bel_E(\cdot)$.
\end{theorem}

\begin{proof} 
For $k=0$, the result is given by Theorem~\ref{theo:weak}. 
Since, according to Theorem~\ref{theo:belE}, $1$-equivalence implies 
$k$-equivalence for all $k \geq 1$ under $\bel_E(\cdot)$, it remains to 
show $\coNP$-completeness for $k=1$. 

Membership follows from Lemma~\ref{lemma:one-equiv}: For deciding
whether $\useq{P}$ and $\useq{Q}$ are not $1$-equivalent, we
guess a set $S$ and a program $P$ according to
Conditions (ii) and (iii) of Lemma~\ref{lemma:one-equiv}. Then we
check in time polynomial in the size of $\useq{P}+\useq{Q}$ whether
$\useq{P}$ and $\useq{Q}$ are not $1$-equivalent. Hence, this problem
is in $\NP$.  Consequently, 
checking whether $\useq{P}$ and $\useq{Q}$ are $1$-equivalent is in $\coNP$.

For showing $\coNP$-hardness, for $k=1$, we give a reduction from 
the problem of tautology checking. Consider a formula 
$F= \bigvee_{i=1}^m (L_{i1}\wedge L_{i1} \wedge L_{i3})$ over atoms 
$A_1,\ldots,A_n$, and two programs $P$ and $Q$ over an alphabet 
$\at \supseteq \{A_1,\ldots,A_n,T\}$ as follows:
\begin{eqnarray*}
P &=& \{\neg A_i \la \naf A_i, \ A_i \la \naf \neg A_i \mid
  i=1,\ldots,n\} \cup \\
  && \{T \la L_{j1},L_{j2},L_{j3}\mid j=1,\ldots,m\}; \\[0.5ex]
Q &=&\{\neg A_i \la \naf A_i, \ A_i \la \naf \neg A_i \mid i=1,\ldots,n\} \cup \\
&&  \{T \la \ \}.
\end{eqnarray*}
Clearly, $P$ and $Q$ can be constructed in polynomial time. We show
that $F$ is a tautology if and only if $P$ and $Q$ are $1$-step
equivalent.

\medskip
\noindent
\emph{If part.} Suppose $F$ is not a tautology. Then, there is a truth
assignment $\sigma$ to $A_1\commadots$ $A_n$ such that $F$ is false,
\iec $L_{i1}\wedge L_{i1} \wedge L_{i3}$ is false for $1 \leq i \leq
m$. Let $R$ be the program consisting of facts $A_i \la$, for every
atom $A_i$, $1 \leq i \leq n$, which is true in $\sigma$. It is easily
verified that the set $S = \{A_i \mid A_i \la\, \in R\} \cup \{\neg
A_j \mid A_j \la\, \notin R \}$ is the only update answer set of $P+R$,
while $S^\prime = S \cup \{T\}$ is the only update answer set of
$Q+R$. Thus, $P$ and $Q$ are not $1$-equivalent.

\medskip
\noindent
\emph{Only-if part.} Suppose $F$ is a tautology. Towards a
contradiction, assume that $P$  and $Q$ are not $1$-equivalent.
Then, by virtue of Lemma~\ref{lemm:belE}, 
there is a (consistent) program $R$ and some set $S$ such that
either $S \in \ASup(P+R)$ and $S \notin \ASup(Q+R)$, or $S \notin \ASup(P+R)$ and $S \in \ASup(Q+R)$ holds. Observe that, for any 
set $S$ and every program $R$, the sets $\rs(S,P+R)$ and $\rs(S,Q+R)$ do 
not differ \wrt rules in $P\cap Q$. 
Furthermore, $P^S$ and $Q^S$ do not differ \wrt to rules in 
$(P\cap Q)^S$.

We first show that $|S\cap\{A_i,\neg A_i\}|=1$ holds, for $1\leq
i\leq n$. Indeed, since $S$ is consistent, $\{A_i, \neg A_i\}
\subseteq S$ cannot hold for any $1\leq i \leq n$.  On the other hand,
suppose that neither $A_i\in S$, nor $\neg A_i\in S$ holds for some
$1\leq i \leq n$. Then, $S$ entails the rules $\neg A_i \la \naf A_i$
and $A_i \la \naf \neg A_i$ of $P\cap Q$, which also cannot be
rejected (since neither $A_i\in S$ nor $\neg A_i\in S$). Thus, both
$\neg A_i \la\,$ and $A_i \la\,$ are in $((P\cup R)\setminus
\rs(S,P+R))^S$ as well as in $((Q\cup R)\setminus
\rs(S,Q+R))^S$. However, this contradicts the assumption that $S$ is a
consistent answer set of either $P+R$ or $Q+R$. This proves $|S\cap\{A_i,\neg
A_i\}|=1$, for $1\leq i\leq n$.
 
Assume first that $\neg T \in S$. Then, every rule $r$ of $P$ such that
$\head{r}=T$ and $S\models \body{r}$ are in $\rs(S,P+R)$, and $T
\la\,$ is in $\rs(S,Q+R)$. Since $S$ is an answer set of either $P+R$
or $Q+R$, it is either the least set of literals closed
under the rules of $P_1 = ((P\cup R)\setminus \rs(S,P+R))^S$ or under the
rules of $Q_1 = ((Q\cup R)\setminus \rs(S,Q+R))^S$. Since $P$ and $Q$
coincide on all rules with head different from $T$, it follows that
$S$ must be the least set of literals closed under the rules of $P_1$
as well as of $Q_1$. Thus, $S$ is an answer set of both $P+R$ and
$Q+R$, which is a contradiction. Hence, $\neg T \notin S$ holds.

It is now easy to show that $T \in S$ must hold. Indeed, if $S \in
\ASup(P+R)$, then, since $F$ is a tautology, 
$S\models\body{r}$ for some rule $r\in P$ such that $\head{r}=T$. 
Moreover, $r \notin \rs(S,P+R)$ since $\neg T \notin S$, which in turn means 
$T \in S$. If, on the other hand, $S$ is an answer set of $Q+R$, then $T\la \ \notin \rs(S,Q+R)$ holds, and thus $T
\in S$ must hold. 

Now suppose that $S \in \ASup(P+R)$. Since $F$ is a tautology, 
$S\models\body{r}$ for some rule $r\in P$ such that
$\head{r}=T$. Since  $\neg T \notin S$, it follows that $T \in
S$. Since $T\la\;$ is in $Q$, and $P$ and $Q$ coincide on all rules 
except those with head $T$, it follows that $S$ is the least set of
literals closed under the rules of $((Q\cup R)\setminus
\rs(S,Q+R))^S)$. Thus, $S \in \ASup(Q+R)$, which is a contradiction. 
On the other hand, suppose $S \notin \ASup(Q+R)$ first. Since $\neg
T\notin S$, we have $T\in S$, and thus clearly $S$ must be the least set of
literals closed under the rules of $((P\cup R)\setminus
\rs(S,P+R))^S)$. Hence, $S \in \ASup(P+R)$, which is again a
contradiction. 

Hence, a program $R$ and a set $S$ as hypothesized cannot exist. This
shows  that $P$ and $Q$ are 1-equivalent. 

We have shown that for every $k\geq0$, deciding $k$-equivalence of finite  
propositional update sequences $\useq{P}$ and $\useq{Q}$ is $\coNP$-complete 
under the $\bel_E(\cdot)$ semantics, which proves our result. 
\hfill$\Box$
\end{proof}

In~\cite{leite-02} it is shown that two dynamic logic programs, $\useq{P}$ and 
$\useq{Q}$, are not $k$-equivalent, for $k > 0$, iff there exists a GLP $P$ 
such that $\SMup(\useq{P}+P)\neq \SMup(\useq{Q}+P)$. This result, together 
with a corresponding version of Lemma~\ref{lemm:belE} and complexity results 
for dynamic logic programming from~\cite{leite-02}, can be used to obtain the 
following analogous result.

\begin{proposition}
Deciding weak, strong, or $k$-equivalence, for a given $k\geq 0$, of two 
given finite propositional dynamic logic programs $\useq{P}$ and $\useq{Q}$, 
is $\coNP$-complete under $\bel_{\oplus}(\cdot)$.
\end{proposition}

\section{Related Work and Conclusion} 
\label{sec:conclusion}

Our work on evolving nonmonotonic knowledge bases is related to
several works in the literature on different issues.

Clearly, our formalization of reasoning from evolution frames is closely related to model
checking of CTL formulas~\cite{clar-etal-99}, and so are our
complexity results. The major difference is, however, that in Kripke 
structures the models are given implicitly by its labels.  Nevertheless, 
since the semantics of
evolution frames can be captured by Kripke structures, it is
suggestive to transform reasoning problems on them into model checking
problems. However, in current model checking systems (\egc the 
\emph{Symbolic Model Verifier} (SMV)~\cite{mcmillan-93}, or its new version 
NuSMV~\cite{cima-etal-00}), state
transitions must be specified in a polynomial-time language, but
descriptions of these Kripke structures would require exponential
space even for evolution frames with \PSPACE~complexity (e.g., $\UL$
evolution frames).  Thus, extensions of model checking systems would
be needed for fruitful usability.

Our filtration results for identifying finitary characterizations,
which are based on various notions of equivalence between knowledge
states, are somewhat related to results in~\cite{pearce-01,lin-02},
obtained independently of our work and of each other. While we were
concerned with the equivalence of sequences of ELPs (including the
case of single ELPs), one can define two logic programs, $P_1$ and
$P_2$, to be equivalent (\iec weakly equivalent in our terminology), if
they yield the same answer sets. They are
called strongly equivalent, similar in spirit to $1$-equivalence in our
terminology, iff, for any logic program $P$, programs $P_1\cup P$ and
$P_2\cup P$ have the same answer sets.
Pearce  \emph{et al.}~\cite{pearce-01} investigated efficient (\iec
linear-time computable) encodings of
\emph{nested logic programs}, a proper generalization of disjunctive
logic programs, into QBFs. In accordance with our results, they found
that deciding whether two propositional nested logic programs are
strongly equivalent is \coNP-complete. In~\cite{lin-02},
the same result has been shown, but merely for disjunctive
logic programs.

Lobo  \emph{et al.}\ introduced the ${\cal PDL}$ \cite{lobo-etal-99} language
for policies, which contain \emph{event-condition-action rules},
serving for modeling reactive behavior on observations
from an environment. While similar in spirit, their model is
different, and \cite{lobo-etal-99} focuses on detecting action
conflicts (which, in our framework, is not an issue). In~\cite{lobo-son-01}, reasoning tasks
are considered which center around actions. Further related 
research is on planning, where certain reachability problems are
\PSPACE-complete (cf.\ \cite{bara-etal-2000}). Similar results were
obtained in \cite{wool-00b} for related agent design
problems. However, in all these works, the problems considered
are ad hoc, and no reasoning language is considered.

Fagin \emph{et al.}'s \cite{fagi-etal-95a} important work on knowledge
in multi-agent systems addresses evolving knowledge, but mainly at an
axiomatic level. Wooldridge's \cite{wool-00} logic for reasoning about
multi-agent systems embeds $\textrm{CTL}^\ast$ and has belief, desire
and intention modalities. The underlying model is very broad, and aims
at agent communication and cooperation. It remains to see how our
particular framework fits into these approaches.

Leite \cite{leite-02} introduces in his Ph.D.\ thesis a language, KABUL,
which is inspired by our $\UL$ language, but goes beyond it, since
this language foresees also possible updates to the update
policy. That is, the function $\Pi$ may change over time, depending on
external events. This is not modeled by our evolution frames, in which
$\Pi$ is the same at every instance of time. However, a generalization
towards a time-dependent update policy---and possibly other
time-dependent components of an evolution frame---seems not difficult
to accomplish. Furthermore, Leite's work does not include a formal
language for expressing properties of evolving knowledge bases like
ours, and also does not address complexity issues of the
framework.

\subsection{Further work} 

In this paper, we have presented a general framework for modeling
knowledge bases built over (extended) logic programs, for which we
have then defined a formal language, \EKBL, for stating and evaluating
properties of a nonmonotonic knowledge base which evolves over time.
As we have shown, this framework, which results from an abstraction of
previous work on update languages for nonmonotonic logic programs
\cite{eite-etal-01,eite-etal-02c}, can be used to abstractly model
several approaches for updating logic programs in the
literature. Knowledge about properties of the framework may thus be
helpful to infer properties of these and other update approaches, and
in particular about their computational properties. 
In this line, we
have studied semantic properties of the framework, and we have
identified several classes of evolution frames for which reasoning
about evolving knowledge bases in the language \EKBL\ is decidable. In
the course of this, we have established that reasoning about
propositional evolving knowledge bases maintained by $\UL$ update
policies under the answer set semantics
\cite{eite-etal-01,eite-etal-02c} is \PSPACE-complete.

While we have tackled several issues in this paper, other issues remain
for further work. One issue is to identify further meaningful semantic
constraints on evolution frames or their components, and investigate
the semantic and computational properties of the resultant evolution
frames. For example, iterativity of the compilation $\compo_{\EF}$,
i.e., where the events are incorporated one at a time, or properties of the
belief operator $\bel$, would be interesting to explore. 

Another interesting topic, and actually related to this, is finding
fragments of lower comlexity and, in particular, of polynomial-time
complexity. Furthermore, the investigation of special event classes,
\egc \emph{event patterns}, which exhibit regularities in sequences of
events, is an interesting issue.

\bibliographystyle{abbrv}


\newcommand{\SortNoOp}[1]{}

\end{document}